\documentclass{article}

\usepackage{microtype}

\usepackage{graphicx}

\usepackage{subcaption}
\usepackage{color}
\usepackage{diagbox}
\usepackage{multirow}
\usepackage{float}
\usepackage{kotex}
\usepackage[percent]{overpic}
\usepackage{svg}

\usepackage{booktabs} 

\usepackage{hyperref}

\usepackage[accepted]{icml2025}

\usepackage{amsmath}
\usepackage{amssymb}
\usepackage{mathtools}
\usepackage{amsthm}

\usepackage[capitalize,noabbrev]{cleveref}

\theoremstyle{plain}
\newtheorem{theorem}{Theorem}[section]

\theoremstyle{definition}
\newtheorem{definition}[theorem]{Definition}

\theoremstyle{remark}

\usepackage{amsmath,mathtools}
\usepackage{makecell}
\makeatletter 
\newcommand{\LCASES}[1]{$\m@th\displaystyle{#1}$\hfil}
\newcommand{\CCASES}[1]{\hfil$\m@th\displaystyle{#1}$\hfil}
\newcommand{\RCASES}[1]{\hfil$\m@th\displaystyle{#1}$}
\makeatother

\newcases{ecases}{\quad}{\CCASES{##}}{\LCASES{##}}{\lbrace}{.}
\newcases{ecases*}{\quad}{\CCASES{##}}{{##}\hfil}{\lbrace}{.}

\usepackage[textsize=tiny]{todonotes}

\icmltitlerunning{Wolfpack Adversarial Attack for Robust Multi-Agent Reinforcement Learning}

\begin{document}

\twocolumn[
\icmltitle{Wolfpack Adversarial Attack for Robust Multi-Agent Reinforcement Learning}



\icmlsetsymbol{equal}{*}

\begin{icmlauthorlist}
\icmlauthor{Sunwoo Lee}{yyy}
\icmlauthor{Jaebak Hwang}{yyy}
\icmlauthor{Yonghyeon Jo}{yyy}
\icmlauthor{Seungyul Han}{yyy} \hspace{-0.12in} $^{*}$
\end{icmlauthorlist}

\icmlaffiliation{yyy}{Graduate School of Artificial Intelligence, UNIST, Ulsan, South Korea}

\icmlcorrespondingauthor{Seungyul Han}{syhan@unist.ac.kr}

\icmlkeywords{Machine Learning, ICML}

\vskip 0.3in
]



\printAffiliationsAndNotice{}  

\begin{abstract}
Traditional robust methods in multi-agent reinforcement learning (MARL) often struggle
against coordinated adversarial attacks in cooperative scenarios. To address this limitation,
we propose the Wolfpack Adversarial Attack
framework, inspired by wolf hunting strategies,
which targets an initial agent and its assisting agents to disrupt cooperation. Additionally,
we introduce the Wolfpack-Adversarial Learning for MARL (WALL) framework, which trains
robust MARL policies to defend against the
proposed Wolfpack attack by fostering systemwide collaboration. Experimental results underscore the devastating impact of the Wolfpack attack and the significant robustness improvements
achieved by WALL. Our code is available at
https://github.com/sunwoolee0504/WALL.

\end{abstract}
\section{Introduction}
\label{sec:intro}

Multi-agent Reinforcement Learning (MARL) has gained attention for solving complex problems requiring agent cooperation \cite{oroojlooy2023review} and competition, such as drone control \cite{yun2022cooperative}, autonomous navigation \cite{chen2023deep}, robotics \cite{orr2023multi}, and energy management \cite{jendoubi2023multi}. To handle partially observable environments, the Centralized Training and Decentralized Execution (CTDE) framework \cite{oliehoek2008optimal} trains a global value function centrally while agents execute policies based on local observations. Notable credit-assignment methods in CTDE include Value Decomposition Networks (VDN) \cite{sunehag2017value}, QMIX \cite{rashid2020monotonic}, which satisfies the Individual-Global-Max (IGM) condition ensuring that optimal joint actions align with positive gradients in global and individual value functions, and QPLEX \cite{wang2020qplex}, which encodes IGM into its architecture. However, CTDE methods face challenges from exploration inefficiencies \cite{mahajan2019maven, jo2024fox} and mismatches between training and deployment environments, leading to unexpected agent behaviors and degraded performance \cite{moos2022robust, guo2022towards}. Thus, enhancing the robustness of CTDE remains a critical research focus.

To improve learning robustness, single-agent RL methods have explored strategies based on game theory \cite{yu2021robust}, such as max-min approaches and adversarial learning \cite{goodfellow2014explaining, huang2017adversarial, pattanaik2017robust, pinto2017robust}. In multi-agent systems, simultaneous agent interactions introduce additional uncertainties \cite{zhang2021multi}. To address this, methods like perturbing local observations \cite{lin2020robustness}, training with adversarial policies for Nash equilibrium \cite{li2023byzantine}, adversarial value decomposition \cite{phan2021resilient}, and attacking inter-agent communication \cite{xue2021mis} have been proposed. However, these approaches often target a single agent per attack, overlooking interdependencies in cooperative MARL, making them vulnerable to scenarios where multiple agents are attacked simultaneously.

To overcome the vulnerabilities posed by coordinated adversarial attacks in MARL, we propose the Wolfpack adversarial attack framework, inspired by wolf hunting strategies. This approach disrupts inter-agent cooperation by targeting a single agent and subsequently attacking the group of agents assisting the initially targeted agent, resulting in more devastating impacts. Experimental results reveal that traditional robust MARL methods are highly susceptible to such coordinated attacks, underscoring the need for new defense mechanisms.
In response, we also introduce the Wolfpack-Adversarial Learning for MARL (WALL) framework, a robust policy training approach specifically designed to counter the Wolfpack Adversarial Attack. By fostering system-wide collaboration and avoiding reliance on specific agent subsets, WALL enables agents to defend effectively against coordinated attacks. Experimental evaluations demonstrate that WALL significantly improves robustness compared to existing methods while maintaining high performance under a wide range of adversarial attack scenarios. The key contributions of this paper in constructing the Wolfpack Adversarial Attack are summarized as follows:
\vspace{-1em}

\begin{itemize}
\item A novel MARL attack strategy, \textbf{Wolfpack Adversarial Attack}, is introduced, targeting multiple agents simultaneously to foster stronger and more resilient agent cooperation during policy training.
\item \textbf{The follow-up agent group selection} method is proposed to target agents with significant behavioral adjustments to an initial attack, enabling subsequent sequential attacks and amplifying their overall impact.
\item \textbf{A planner-based attacking step selector} predicts future $Q$-value reductions caused by the attack, enabling the selection of critical time steps to maximize impact and improve learning robustness.
\end{itemize}

\section{Related Works}
\textbf{Robust MARL Strategies:}
Recent research has focused on robust MARL to address unexpected changes in multi-agent environments. Max-min optimization \cite{chinchuluun2008pareto, han2021max} has been applied to traditional MARL algorithms for robust learning \cite{li2019robust, wang2022data}. Robust Nash equilibrium has been redefined to better suit multi-agent systems \cite{zhang2020robust, li2023byzantine}. Regularization-based approaches have also been explored to improve MARL robustness \cite{lin2020robustness, li2023mir2, wang2023regularization, bukharin2024robust}, alongside distributional reinforcement learning methods to manage uncertainties \cite{li2020multi, xu2021mmd, du2024robust, geng2024noise}.

\textbf{Adversarial Attacks for Resilient RL:}
To strengthen RL, numerous studies have explored adversarial learning to train policies under worst-case scenarios \cite{pattanaik2017robust, tessler2019action, pinto2017robust, chae2022robust}. These attacks introduce perturbations to various MDP components, including state \cite{zhang2020robust_1, zhang2021robust, everett2021certifiable, li2023ats, qiaoben2024understanding}, action \cite{tan2020robustifying, lee2021query, liu2024robust}, and reward \cite{wang2020reinforcement, zhang2020adaptive, rakhsha2021reward, xu2022efficient, cai2023reward, bouhaddi2023multi, xu2024reward, bouhaddi2024rewards}. Adversarial attacks have recently been extended to multi-agent setups, introducing uncertainties to state or observation \cite{, han2022solution, he2023robust, zhang2023safe, zhou2023robust}, actions \cite{yuan2023robust}, and rewards \cite{kardecs2011discounted}. Further research has applied adversarial attacks to value decomposition frameworks \cite{phan2021resilient}, selected critical agents for targeted attacks \cite{yuan2023robust, zhou2024adversarial}, and analyzed their effects on inter-agent communication \cite{xue2021mis, tu2021adversarial, sun2023certifiably, yuan2024robust}.

\textbf{Model-based Frameworks for Robust RL:}
Model-based methods have been extensively studied to enhance RL robustness \cite{berkenkamp2017safe, panaganti2021sample, curi2021combining, clavier2023towards, shi2024distributionally, ramesh2024distributionally}, including adversarial extensions \cite{wang2020falsification, kobayashi2024lira}. Transition models have been leveraged to improve robustness \cite{mankowitz2019robust, ye2024towards, herremans2024robust}, and offline setups have been explored for robust training \cite{rigter2022rambo, bhardwaj2024adversarial}. In multi-agent systems, model-based approaches address challenges like constructing worst-case sets \cite{shi2024sample} and managing transition kernel uncertainty \cite{he2022robust}.

\begin{figure*}[ht!]
    \centering
    \includegraphics[width=0.9\textwidth]{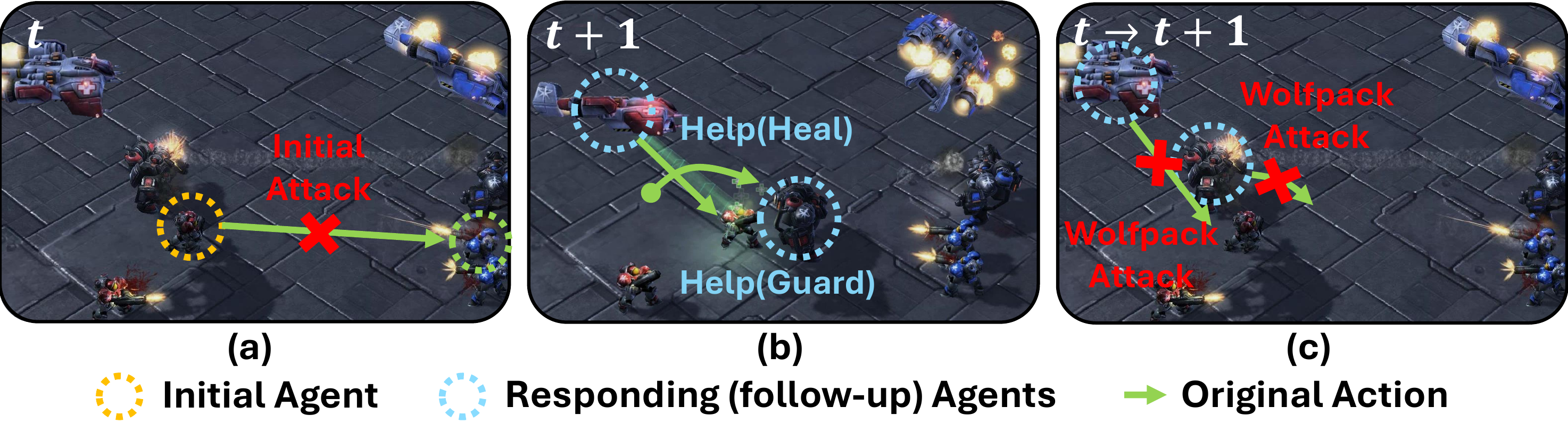}
    \caption{Visualization of Wolfpack attack strategy during combat in the StarCraft II environment: (a) The initial agent is attacked, disrupting its original action (b) Responding (follow-up) agents to help the initially attacked agent and (c) Wolfpack adversarial attack that disrupts help actions of follow-up agents.}
    \vspace{-0.1in}
    \label{fig:concept}
\end{figure*}

\section{Background}
\label{sec:background}

\subsection{Dec-POMDP and Value-based CTDE Setup}
A fully cooperative multi-agent environment is modeled as a decentralized partially observable Markov decision process (Dec-POMDP) \cite{oliehoek2016concise}, defined by the tuple $\mathcal{M}=\langle \mathcal{N}, \mathcal{S}, \mathcal{A},P,\Omega, O, R, \gamma\rangle$. $\mathcal{N}={1,\ldots,n}$ is the set of agents, $\mathcal{S}$ the global state space, $\mathcal{A}=\mathcal{A}^1\times\cdots\times\mathcal{A}^n$ the joint action space, $P$ the state transition probability, $\Omega$ the observation space, $R$ the reward function, and $\gamma\in[0,1)$ the discount factor. At time $t$, each agent $i$ observes $o_t^i=O(s_t,i)\in \Omega$ and takes action $a_t^i\in\mathcal{A}^i$ based on its individual policy $\pi^i(\cdot|\tau_t^i)$, where $\tau_t^i$ is the agent's trajectory up to $t$. The joint action $\mathbf{a}_t=\langle a_t^1,\ldots,a_t^n \rangle$ sampled from the joint policy $\mathbf{\pi}:=\langle\pi^1,\cdots,\pi^n\rangle$ leads to the next state $s_{t+1}\sim P(\cdot|s_t,\mathbf{a}_t)$ and reward $r_t:=R(s_t,\mathbf{a}_t)$. MARL aims to find the optimal joint policy that maximizes $\sum_{t=0}^\infty \gamma^t r_t$. As noted, this paper adopts the centralized training with decentralized execution (CTDE) setup, where the joint value $Q^{tot}(s_t,\mathbf{a}_t)$ is learned using temporal-difference (TD) learning. Through credit assignment, individual value functions $Q^i(\tau_t^i,a_t^i)$ are learned, guiding individual policies $\pi^i$ to select actions that maximize $Q^i$, i.e., $\pi^i:=\arg\max_{a_t^i\in\mathcal{A}^i} Q^i(\tau_t^i,\cdot)$.

\subsection{Robust MARL with Adversarial Attack Policy}
Among various methods for robust learning in MARL,  \citet{yuan2023robust} defined an adversarial attack policy $\pi_{\mathrm{adv}}$ and implemented robust MARL by training multi-agent policies to defend against attacks executed by $\pi_{\mathrm{adv}}:\mathcal{S}\times\mathcal{A}\times\mathbb{N} \rightarrow \mathcal{A}$. A cooperative MARL environment with an adversarial policy $\pi_{\mathrm{adv}}$ can be described as a Limited Policy Adversary Dec-POMDP (LPA-Dec-POMDP) $\tilde{\mathcal{M}}$, defined as follows:

\begin{definition} [Limited Policy Adversary Dec-POMDP]
Given a Dec-POMDP $\mathcal{M}$ and a fixed adversarial policy $\pi_{\mathrm{adv}}$, we define a Limited Policy Adversary Dec-POMDP (LPA-Dec-POMDP) $\tilde{\mathcal{M}}=\langle \mathcal{N}, \mathcal{S}, \mathcal{A},P,K,\Omega, O, R, \gamma\rangle$, where $K$ is the maximum number of attacks,
$\pi_{\mathrm{adv}}(\cdot|s_t,\mathbf{a}_t,k_t)$ executes joint action $\tilde{\mathbf{a}}_t$ to disrupt the original action $\mathbf{a}_t$ chosen by $\pi$, and $k_t\leq K$ indicates the number of remaining attacks.
\label{def:LPA}
\end{definition}
Here, if $\pi_{\mathrm{adv}}$ selects an attack action $\tilde{\mathbf{a}}_t$ different from the original action $\mathbf{a}_t$, the remaining number of attacks $k_t$ decreases by 1. Once $k_t$ reaches 0, no further attacks can be performed. In this framework, \citet{yuan2023robust} also demonstrated that the LPA-Dec-POMDP can be represented as another Dec-POMDP induced by $\pi_{\mathrm{adv}}$, with the convergence of the optimal policy $\pi^*$ under $\tilde{\mathcal{M}}$ guaranteed. In particular, it considers an adversarial attack targeting a chosen agent $i$ by minimizing its individual value function $Q^i$, as proposed by \cite{pattanaik2017robust}, i.e., $\tilde{a}_t^i=\arg\min_{a\in\mathcal{A}^i} Q^i(\tau_t^i,a)$. Additionally, to enhance the attack's effectiveness, an evolutionary generation of attacker (EGA) approach is proposed, which combines multiple adversarial policies.

\section{Methodology}

\subsection{Motivation of Wolfpack Attack Strategy}
\label{subsec:motiv}

Existing adversarial attackers typically target only a single agent per attack, without coordination or relationships between successive attacks. In a cooperative MARL setup, such simplistic attacks enable non-targeted agents to learn effective policies to counteract the attack. However, we observe that policies trained under these conditions are vulnerable to coordinated attacks. As illustrated in Fig. \ref{fig:concept}(a), a single agent is attacked at time $t$. In Fig. \ref{fig:concept}(b), at the next step $t+1$, responding agents adjust their actions, such as healing or moving to guard, to protect the initially attacked agent. In contrast, Fig. \ref{fig:concept}(c) demonstrates a coordinated attack strategy that targets the agents responding to the initial attack. Such coordinated attacks render the learned policy ineffective, preventing it from countering the attacks entirely. This highlights that coordinated attacks are far more detrimental than existing attack methods, and current robust policies fail to defend effectively against them.   

As depicted in Fig. \ref{fig:concept}(c), targeting agents that respond to an initial attack aligns with the Wolfpack attack strategy, a tactic widely employed in traditional military operations, as discussed in Section \ref{sec:intro}. To adapt this concept to a cooperative multi-agent setup, we define a Wolfpack adversarial attack as a coordinated strategy where one agent is attacked initially, followed by targeting the group of follow-up agents that respond to defend against the initial attack, as shown in Fig. \ref{fig:concept}(c). Leveraging this approach, we aim to develop robust policies capable of effectively countering Wolfpack adversarial attack, thereby significantly enhancing the overall resilience of the learning process.

\subsection{Wolfpack Adversarial Attack}

In this section, we formally propose the Wolfpack adversarial attack, as introduced in the previous sections. The Wolfpack attack consists of two components: initial attacks, where a single agent is targeted at a specific time step $t_{\mathrm{init}}$, and follow-up group attacks, where the group of agents responding to the initial attack is selected and targeted over the subsequent steps $t_{\mathrm{init}}+1, \cdots, t_{\mathrm{init}}+t_{\mathrm{WP}}$. Over the course of an episode, a maximum of $K_{\mathrm{WP}}$ Wolfpack attacks can be executed. Consequently, the total number of attack steps is given by $K = K_{\mathrm{WP}} \times (t_{\mathrm{WP}} + 1)$. The Wolfpack adversarial attacker $\pi_{\mathrm{adv}}^{\mathrm{WP}}$ can then be defined as follows:

\begin{definition} [Wolfpack Adversarial Attacker]
A Wolfpack adversarial attacker $\pi_{\mathrm{adv}}^{\mathrm{WP}}:\mathcal{S}\times\mathcal{A}\times\mathbb{N} \rightarrow \mathcal{A}$ is defined as $\tilde{\mathbf{a}}_t = \pi_{\mathrm{adv}}^{\mathrm{WP}}(s_t, \mathbf{a}_t,k_t)$, where $\tilde{a}_t^i=\arg\min_{a_t^i\in\mathcal{A}^i} Q^{tot}(s_t,a_t^i,\mathbf{a}_t^{-i})$ for all $i\in \mathcal{N}_{t,\mathrm{attack}}$, and $\tilde{a}_t^i = a_t^i$ otherwise. Here, $\mathbf{a}_t^{-i}$ represents the joint actions of all agents excluding the $i$-th agent, and $\mathcal{N}_{t,\mathrm{attack}}$ denotes the set of agents targeted for adversarial attack, defined as
\vspace{-1.5em}

{\small\begin{equation*}
\mathcal{N}_{t,\mathrm{attack}}=
\begin{ecases*}
  \emptyset  & if $k_t=0$,\\
  \{i\} & else if $t=t_\mathrm{init}$, $i\sim \mathrm{Unif}(\mathcal{N})$,\\
  \mathcal{N}_{\mathrm{follow-up}}  & else if $t=t_{\mathrm{init}}+1,  \cdots, t_{\mathrm{init}}+t_{\mathrm{WP}}$,\\
  \emptyset           & otherwise,
\end{ecases*}
\end{equation*}}
where $\mathrm{Unif}(\cdot)$ is the Uniform distribution, $\mathcal{N}_{\mathrm{follow-up}}:=\{i_1,\cdots,i_{m}\}\subset \mathcal{N}$ is the group of agents selected for follow-up attack, and $m$ is the number of follow-up agents.\label{def:wolfpack}\end{definition}

Here, note that $k_t$ decreases by $1$ for every attack step such that $\tilde{\mathbf{a}}_t \neq \mathbf{a}_t$, as in the ordinary adversarial attack policy, and the total value function $Q^{tot}$ is used for the attack instead of $Q^i$. The proposed Wolfpack adversarial attacker $\pi_{\mathrm{adv}}^{\mathrm{WP}}$ is a special case of the adversarial policy defined in Definition \ref{def:LPA}. Consequently, the proposed attacker forms an LPA-Dec-POMDP $\tilde{\mathcal{M}}$ induced by $\pi_{\mathrm{adv}}^{\mathrm{WP}}$, and as demonstrated in \citet{yuan2023robust}, the convergence of MARL within the LPA-Dec-POMDP can be guaranteed. The proposed Wolfpack attack involves two key issues: how to design the group of follow-up agents $\mathcal{N}_{\mathrm{follow-up}}$ and when to select $t_{\mathrm{init}}$. The following sections address these aspects in detail.

\subsection{Follow-up Agent Group Selection Method}
\label{subsec:follow}

\begin{figure}[t!]
    \begin{center}
    \centerline{\includegraphics[width=0.49\textwidth]{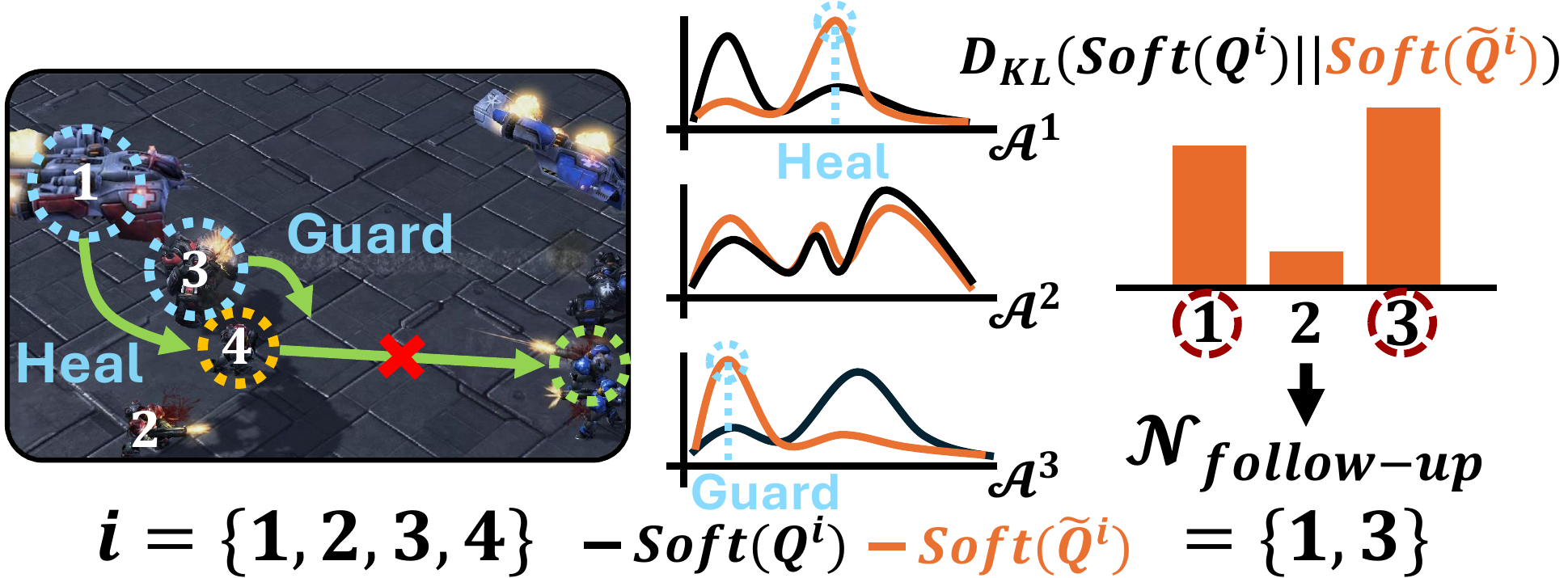}}
    \caption{Visualization of follow-up agent group selection method: Agent $4$ is initially attacked, and the $m$ agents exhibiting the largest changes in $Q^i$ are selected from $\{1, 2, 3\}$ $(m=2)$.}
    \vspace{-0.3in}
    \label{fig:follow}
    \end{center}
\end{figure}

In the Wolfpack adversarial attacker, we aim to identify the follow-up agent group $\mathcal{N}_{\mathrm{follow-up}}$ that actively responds to the initial attack $\pi_{\mathrm{adv}}^{\mathrm{WP}}(s_{t_{\mathrm{init}}}, \mathbf{a}_{t_{\mathrm{init}}}, k_{t_{\mathrm{init}}})$ and target them in subsequent steps. To do this, we define the difference between the $Q$-functions from the original action and the initial attack at time $t$ as:
\begin{align*}
\Delta Q_t^{tot} = Q^{tot}(s_t, \mathbf{a}_t) - Q^{tot}(s_t, \tilde{\mathbf{a}}_t),
\end{align*}
where $\Delta Q_t^{tot} \geq 0$ for all $t$ such that $\mathcal{N}_{t,\mathrm{attack}} \neq \emptyset$, because $\tilde{\mathbf{a}}_t$ minimizes $Q^{tot}$ for the agent indices selected by $\pi_{\mathrm{adv}}^{\mathrm{WP}}$. Assuming the $i$-th agent is the target of the initial attack, updating $Q^{tot}$ based on $\Delta Q_{t_\mathrm{init}}^{tot}$ adjusts each agent's individual value function $Q^j$ to increase $Q^{tot}$ for all $j \neq i \in \mathcal{N}$, in accordance with the credit assignment principle in CTDE algorithms \cite{sunehag2017value, rashid2020monotonic}, as shown below:
\begin{equation}
\label{eq:1}
\tilde{Q}^i(\tau_{t_\mathrm{init}}^i,\cdot) = Q^i(\tau_{t_\mathrm{init}}^i,\cdot) - \alpha_{\mathrm{lr}}\left.{\partial \Delta Q_{t_\mathrm{init}}^{tot} \over \partial Q^{i}(\tau_{t_\mathrm{init}}^i,\mathbf{a})}\right|_{\mathbf{a}=\tilde{\mathbf{a}}_{t_\mathrm{init}}},
\end{equation}
where $\alpha_{\mathrm{lr}}$ is the learning rate. As agents select actions based on $Q^j$, changes in $Q^j$ indicate adjustments in their policies in response to the initial attack. Agents with the largest changes in $Q^j$ are identified as follow-up agents, while the $i$-th agent is excluded as it is already under attack and cannot respond immediately. 

To identify the follow-up agent group, the updated $\tilde{Q}^j$ and original $Q^j$ are transformed into distributions using the Softmax function $\mathrm{Soft}()$. This transformation softens the deterministic policy $\pi^j$, which directly selects an action to maximize $Q^j$, making distributional differences easier to compute. The follow-up agent group is determined by selecting the $m$ agents that maximize the Kullback-Leibler (KL) divergence $D_{\mathrm{KL}}$ between these distributions:
\begin{equation}
\label{eq:2}
\begin{aligned}
    \mathcal{N}_{\mathrm{follow-up}}&=\underset{\mathcal{N}' \subset\mathcal{N},|\mathcal{N}'|=m,j\in\mathcal{N}',j\neq i}{\arg\max} \sum_j D_{\mathrm{KL}}\big(\\
    &\mathrm{Soft}(Q^j(\tau_{t_\mathrm{init}}^j,\cdot)) || \mathrm{Soft}(\tilde{Q}^j(\tau_{t_\mathrm{init}}^j,\cdot))\big).
\end{aligned}    
\end{equation}
Using the proposed method, the follow-up agent group is identified as the agents whose policy distributions experience the most significant changes following the initial attack. Fig. \ref{fig:follow} illustrates this process. After the initial attack, $Q$-differences are computed for the remaining agents ${1,2,3}$, and those with the largest changes in individual value functions are selected as the follow-up agent group. These agents are targeted over the next $t_{\mathrm{WP}}$ time steps to prevent them from effectively responding. In Section \ref{sec:exp}, we analyze how the proposed method enhances attack criticalness by comparing it to naive selection methods based solely on observation distances.

\begin{figure}[t!]
    \begin{center}
    \centerline{\includegraphics[width=0.9\columnwidth]{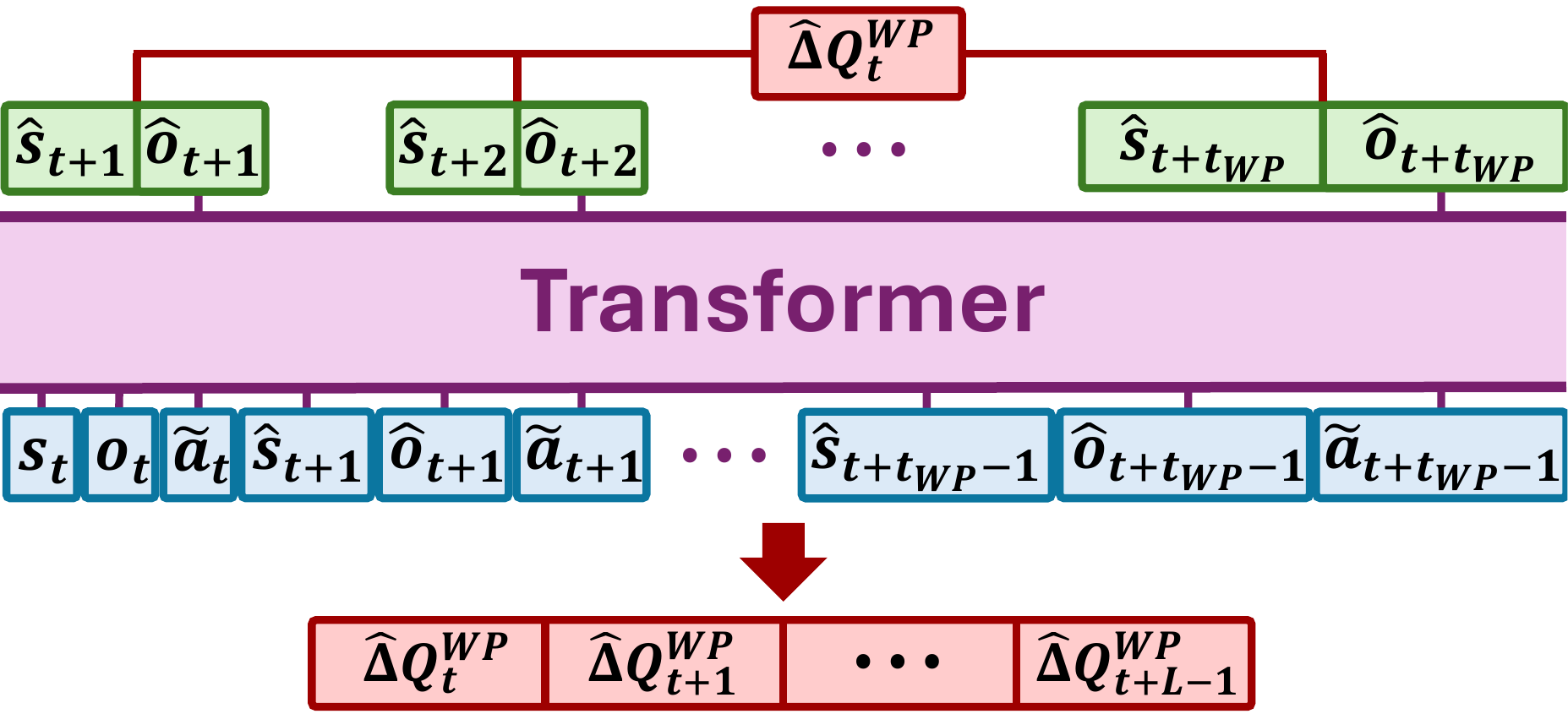}}
    \caption{Planning with Transformer}
    \vspace{-0.3in}
    \label{fig:trans}
    \end{center}
\end{figure}

\subsection{Planner-based Critical Attacking Step Selection}
\label{subsec:attackstep}

In the proposed Wolfpack adversarial attacker $\pi_{\mathrm{adv}}^{\mathrm{WP}}$, the follow-up agent group is defined, leaving the task of determining the timing of initial attacks $t_{\mathrm{init}}$, executed $K_{\mathrm{WP}}$ times within an episode. While Random Step Selection involves choosing time steps randomly, existing methods show that selecting steps to minimize the rewards of the execution policy $\pi$ leads to more effective attacks and facilitates robust learning \cite{yuan2023robust}. However, in coordinated attacks like Wolfpack, targeting steps that cause the greatest reduction in the Q-function value $\Delta Q_t^{\mathrm{WP}}$ ensures a more devastating and lasting impact on the agents' ability to recover and respond. Thus, we propose selecting initial attack times based on the total reduction in $\Delta Q_t^{\mathrm{WP}}$, defined as:
\begin{equation*}
\Delta Q_t^\mathrm{WP} = \sum_{l=t}^{t+t_{\mathrm{WP}}} \Delta Q_l^{tot},
\end{equation*}
where the Wolfpack attack is performed from $t$ (initial attack) to $t+1, \cdots, t+t_{\mathrm{WP}}$ (follow-up attacks). Initial attack time steps $t_{\mathrm{init}}$ are chosen to maximize $\Delta Q_t^\mathrm{WP}$, which captures the total $Q$-value reduction caused by the attack over $t_{\mathrm{WP}}+1$ steps, enhancing the criticalness of the attack. However, computing $\Delta Q_t^\mathrm{WP}$ for every time step is computationally expensive as it requires generating attacked samples through interactions with the environment. To mitigate this, a stored buffer is utilized to plan trajectories of future states and observations for the attack.

\begin{figure}[t!]
    \centering
    \centerline{\includegraphics[width=\columnwidth]{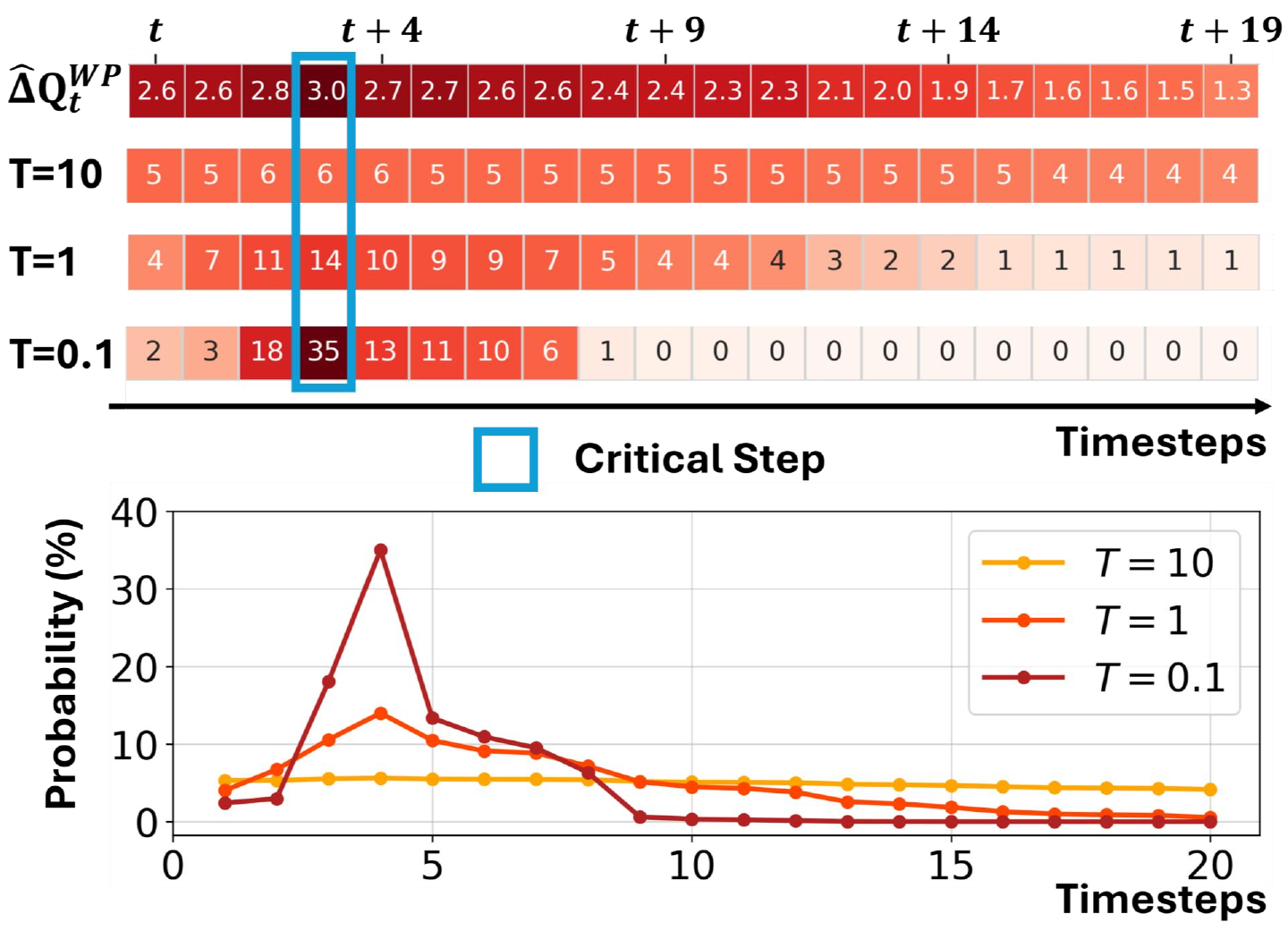}}
    \vspace{-0.1in}
    \caption{Attacking step probabilites}
    \vspace{-0.1in}
    \label{fig:step}
\end{figure}
For planning, we employ a Transformer \cite{vaswani2017attention}, commonly used in sequential learning, which leverages an attention mechanism for efficient learning. As shown in Fig. \ref{fig:trans}, the Transformer learns environment dynamics $P$ using replay buffer trajectories to predict future states and observations $(\hat{s}_{t+1}, \hat{\mathbf{o}}_{t+1}, \cdots, \hat{s}_{t+t_{\mathrm{WP}}}, \hat{\mathbf{o}}_{t+t_{\mathrm{WP}}})$, where $\mathbf{o}_t$ represents the joint observation used to compute $Q^{tot}$. Actions $\tilde{\mathbf{a}}_l = \pi_{\mathrm{adv}}^{\mathrm{WP}}(\hat{s}_l, \mathbf{a}_l, k_l)$ for $\mathbf{a}_l \sim \pi$ are generated by $\pi_{\mathrm{adv}}^{\mathrm{WP}}$ for $l = t, \cdots, t+t_{\mathrm{WP}}$, with $\hat{s}_t = s_t$. Using the planner, we estimate the $Q$-value reduction $\hat{\Delta} Q_t^\mathrm{WP}$ caused by the Wolfpack attack. For $L$ time steps $l = t, \cdots, t+L-1$, we compute future $Q$-differences $\hat{\Delta} Q_l^\mathrm{WP}$ and select $t_{\mathrm{init}}$ based on the initial attack probability $P_{t,\mathrm{attack}}$:
\begin{equation}
P_{t,\mathrm{attack}} = \left\{\mathrm{Soft}\left(\hat{\Delta} Q_t^{\mathrm{WP}}/T,\cdots,\hat{\Delta} Q_{t+L-1}^{\mathrm{WP}}/T\right)\right\}_1,
\label{eq:initialprob}
\end{equation}
where ${\mathbf{x}}_l$ indicates the $l$-th element of $\mathbf{x}$, and $T > 0$ is the temperature. In this paper, we set $L=20$ as it provides an appropriate attack period. After selecting $K_{\mathrm{WP}}$ initial attacks, no further attacks are performed. Fig. \ref{fig:step} shows how step probabilities are distributed for different $T$ values ($T=0.1, 1, 10$). At each time $t$, the planner predicts $\hat{\Delta}Q_t^{\mathrm{WP}}$ for $t$ to $t+L-1$, forming soft initial attack probabilities. A larger $T$ results in more uniform probabilities, while a smaller $T$ increases the likelihood of targeting critical steps where $\hat{\Delta}Q_t^{\mathrm{WP}}$ is highest. These critical steps are selected with the highest probabilities for initial attacks. In Section \ref{sec:exp}, we analyze the effectiveness of this method in delivering more critical attacks compared to Random Step Selection and examine the impact of $T$ on performance in practical environments. Since the proposed method involves planning at every evaluation, we also train a separate model to predict $\hat{\Delta}Q_t^{\mathrm{WP}}$, significantly reducing computational complexity. Details of this approach and the Transformer training loss functions are provided in Appendix \ref{appsubsec:trans}.

\begin{figure}[t!]
    \centering
    \includegraphics[width=0.9\columnwidth]{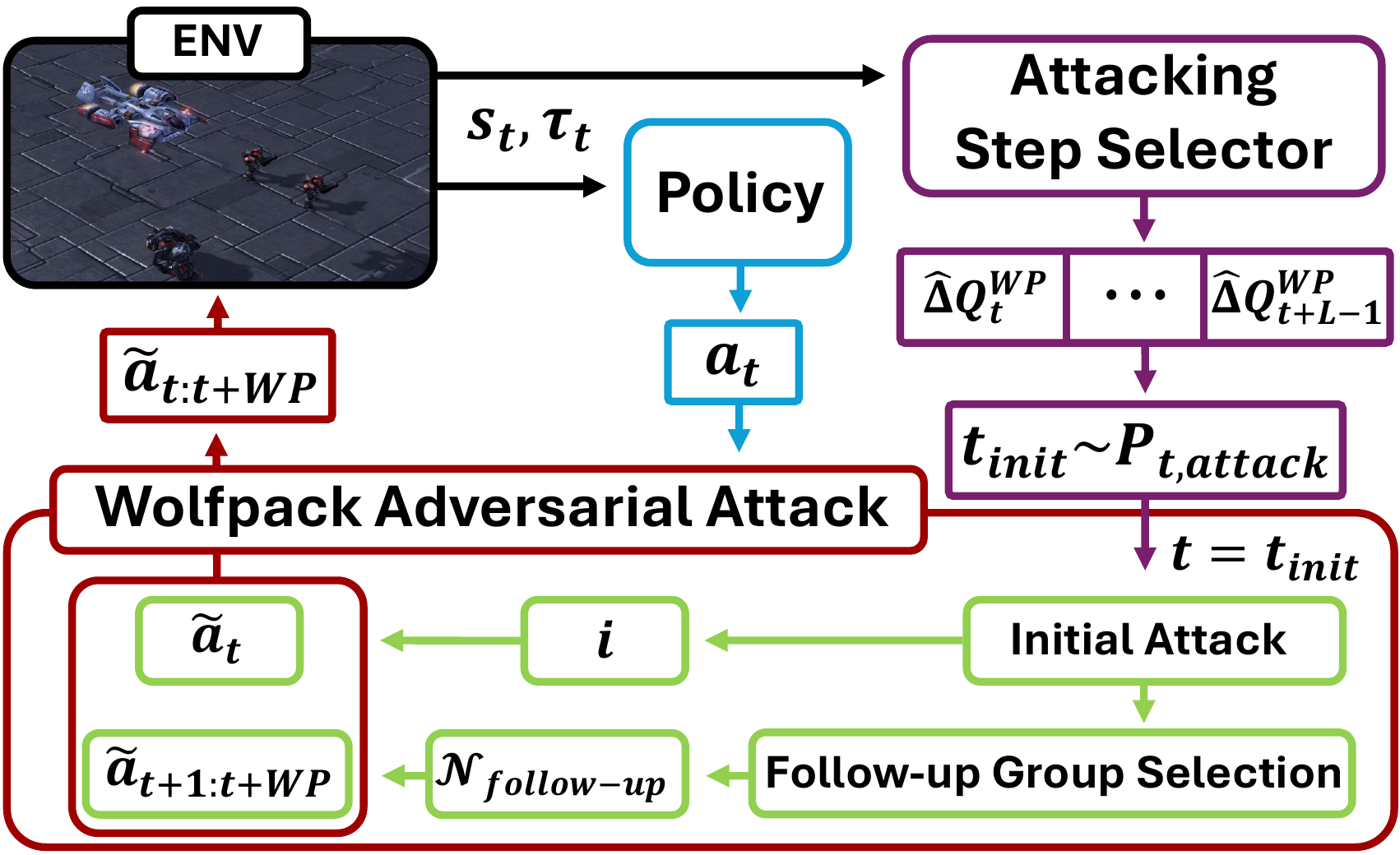}
    \vspace{-0.05in}
    \caption{Illustration of the proposed WALL framework}
    \vspace{-0.15in}
    \label{fig:framework}
\end{figure}

\subsection{WALL: A Robust MARL Algorithm}

Similar to other robust MARL methods, we propose the Wolfpack-Adversarial Learning for MARL (WALL) framework, a robust policy designed to counter the Wolfpack attack by performing MARL on the LPA-Dec-POMDP $\tilde{\mathcal{M}}$ with the Wolfpack attacker $\pi_{\mathrm{adv}}^{\mathrm{WP}}$. While the proposed Wolfpack framework is broadly applicable to most CTDE algorithms, we primarily applied it to well-known value-based CTDE methods, including QMIX \cite{rashid2020monotonic}, VDN \cite{sunehag2017value}, and QPLEX \cite{wang2020qplex}. Detailed implementations, including loss functions for the planner Transformer and the value functions, are provided in Appendix \ref{appsubsec:marl}. The proposed WALL framework is illustrated in Fig. \ref{fig:framework} and summarized in Algorithm \ref{alg:wolfpack}.

\vspace{-.5em}

\begin{algorithm}[htb]
   \caption{WALL framework}
   \label{alg:wolfpack}
\begin{algorithmic}[1]
   \STATE {\bfseries Initialize:} Value function $Q^{tot}$, Planning Transformer
   \FOR{each training iteration}
       \FOR{each environment step $t$}
           \STATE Sample the action $\mathbf{a}_t$: $a_t^i\sim \epsilon$-greedy($Q^i$)
           \STATE Compute \( P_{t,\mathrm{attack}} \) using Planner and sample \(t_{\mathrm{init}}\)
           \IF{$t = t_{\mathrm{init}}$}
               \STATE Perform the initial attack: $\tilde{\mathbf{a}}_t \sim \pi_{\mathrm{adv}}^{\mathrm{WP}}$
           \ELSIF{$t_{\mathrm{init}}+1 \leq t \leq t_{\mathrm{init}}+t_{\mathrm{WP}}$}
               \STATE Select the follow-up agent group $\mathcal{N}_{\mathrm{follow-up}}$
               \STATE Perform the follow-up attack: $\tilde{\mathbf{a}}_t \sim \pi_{\mathrm{adv}}^{\mathrm{WP}}$
           \ELSE
               \STATE Execute the original action $\mathbf{a}_t$
           \ENDIF
       \ENDFOR
       \STATE Update the $Q^{tot}$ using a CTDE algorithm
       \STATE Update the Planning Transformer
   \ENDFOR
\end{algorithmic}
\end{algorithm}
\vspace{-1em}

\begin{figure*}[ht!]
    \centering
    \begin{subfigure}{0.397\columnwidth}
        \centering
        \includegraphics[width=\linewidth]{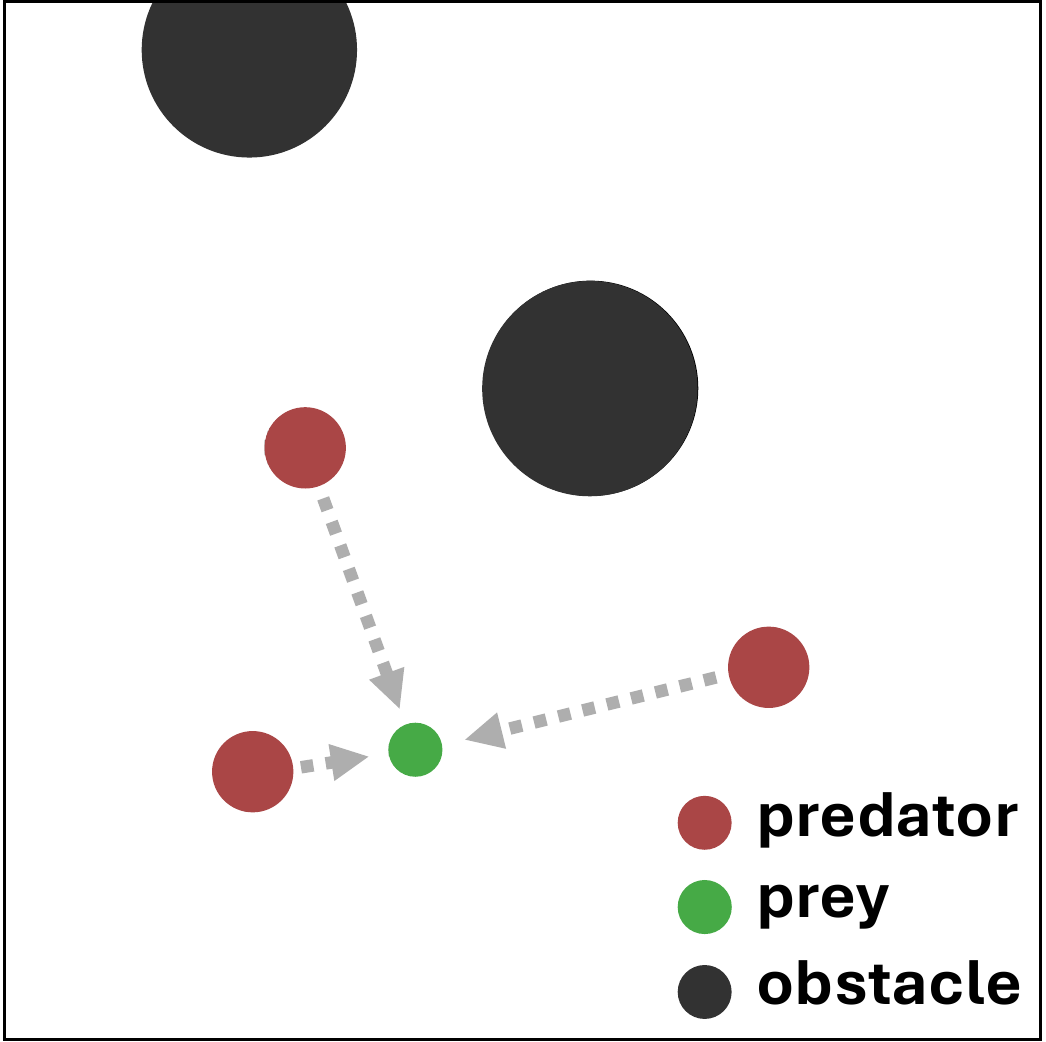}
        \caption{}  
    \end{subfigure}
    \begin{subfigure}{0.397\columnwidth}
        \centering
        \includegraphics[width=\linewidth]{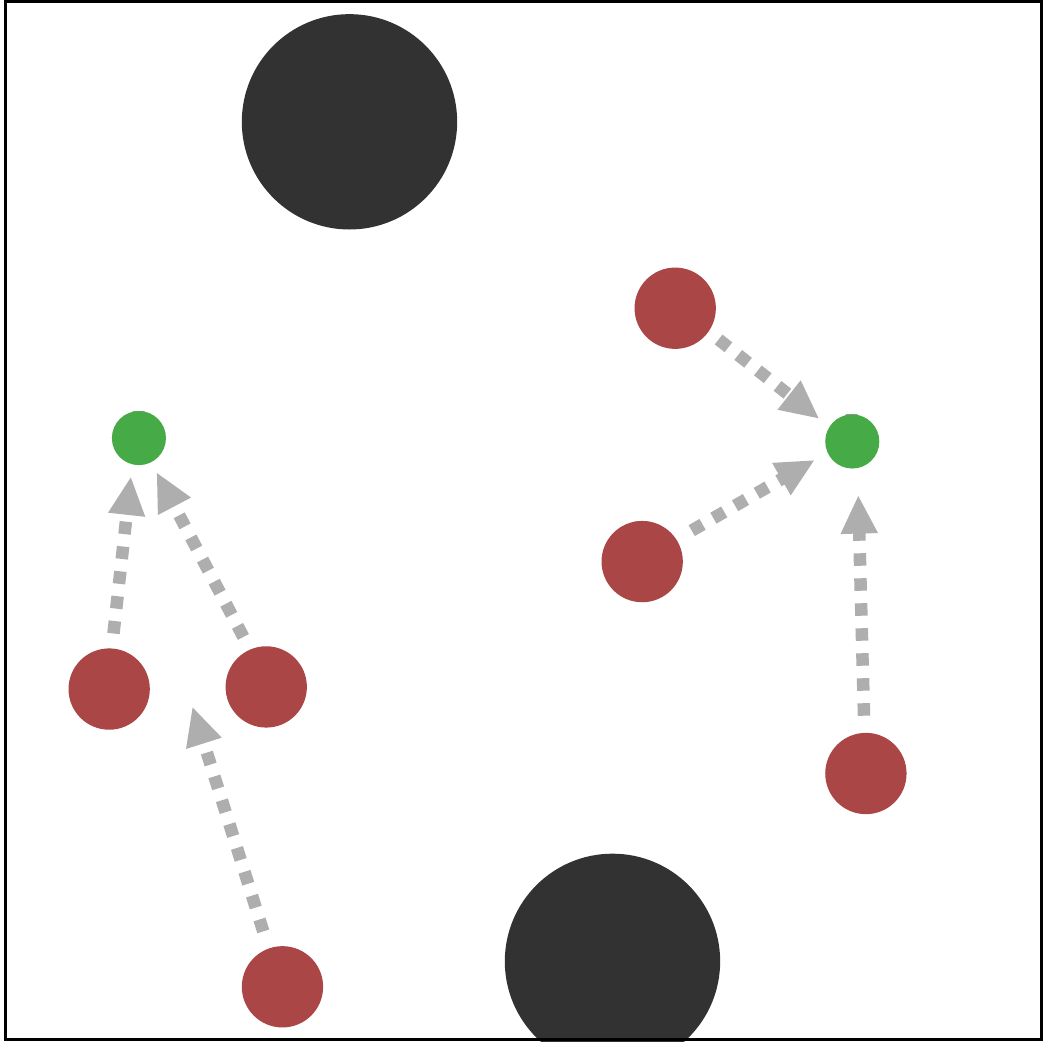}
        \caption{}  
    \end{subfigure}
    \begin{subfigure}{0.62\columnwidth}
        \centering
        \includegraphics[width=\linewidth]{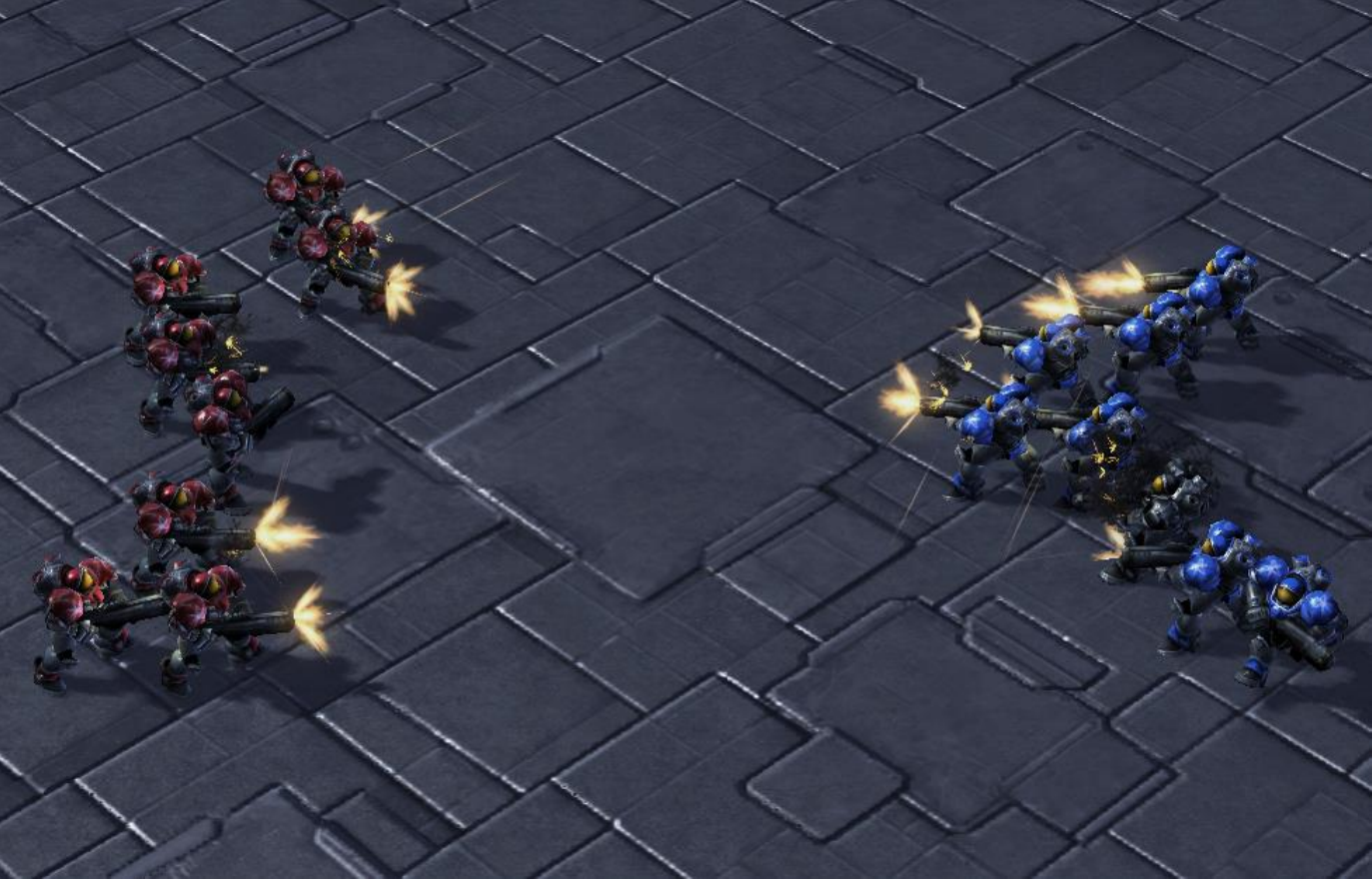}
        \caption{}  
    \end{subfigure}
    \begin{subfigure}{0.62\columnwidth}
        \centering
        \includegraphics[width=\linewidth]{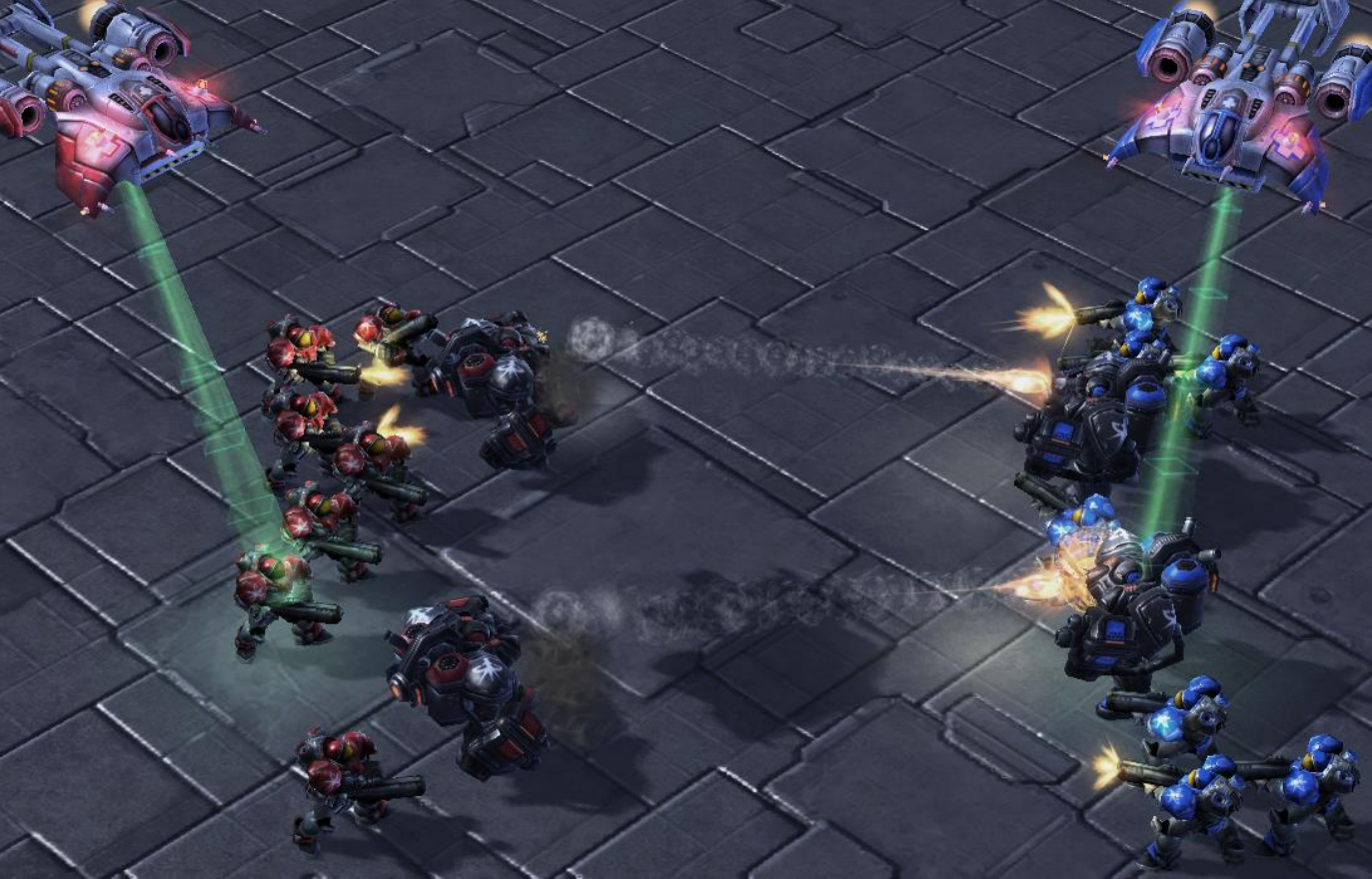}
        \caption{}  
    \end{subfigure}
    \vspace{-0.1in}
    \caption{MARL benchmarks used in our experiments: (a) \texttt{PP\_3/1} and (b) \texttt{PP\_6/2} in MPE, and (c) \texttt{8m} and (d) \texttt{MMM} scenarios in SMAC.}
    \label{fig:vis_benchmarks}
\end{figure*}

\section{Experiments}
\label{sec:exp}

In this section, we evaluate the proposed methods on two standard benchmarks in MARL research: the Multi-Agent Particle Environment (MPE)~\cite{lowe2017multi} and the StarCraft II Multi-Agent Challenge (SMAC)~\cite{samvelyan2019starcraft}, as illustrated in Fig.~\ref{fig:vis_benchmarks}. Specifically, we compare: (1) the impact of the proposed Wolfpack adversarial attack against other adversarial attacks, and (2) the robustness of the WALL framework in defending against such attacks compared to other robust MARL methods. Also, an ablation study analyzes the effect of the proposed components and hyperparameters on robustness. All results are reported as the mean and standard deviation (shaded areas for graphs and $\pm$ values for tables) across 5 random seeds. Our code is available at \href{https://github.com/sunwoolee0504/WALL}{https://github.com/sunwoolee0504/WALL}.

\vspace{0.5cm}

\subsection{Environmental Setup}

The MPE environment provides a multi-agent setting where agents interact through simple physical dynamics. We conduct experiments on three predator-prey (PP) scenarios with varying agent-to-target ratios: \texttt{PP\_3/1}, \texttt{PP\_6/2}, and \texttt{PP\_9/3}. In these tasks, multiple predator agents must coordinate to capture one or more prey agents moving adversarially.
The SMAC environment serves as a challenging benchmark requiring effective agent cooperation to defeat opponents. We evaluate the proposed method across six scenarios: \texttt{2s3z}, \texttt{3m}, \texttt{3s\_vs\_3z}, \texttt{8m}, \texttt{MMM}, and \texttt{1c3s5z}. We perform parameter searches for the number of follow-up agents $m$, total Wolfpack attacks $K_{\mathrm{WP}}$, and attack duration $t_{\mathrm{WP}}$, using optimal settings for comparisons. To ensure realistic constraints, we set $m$ to $m < \left\lfloor \frac{n-1}{2} \right\rfloor$, where $n$ is the maximum number of allied units. We provide details on environment setups and experimental configurations, including hyperparameter settings, in Appendices \ref{appsec:env} and \ref{appsec:expdetail}. All MARL methods are evaluated on the QMIX baseline, with comparison for other CTDE baselines in Appendix \ref{appsubsec:compotherctde}.

\begin{table}[t!]
\centering
\vspace{1em}
\resizebox{0.49\textwidth}{!}{
\begin{tabular}{|c|c|c|c|c|c|}
\hline
\multicolumn{2}{|c|}{\diagbox[innerwidth=5cm,dir=NW]{Method}{Scenario}} 
& \begin{tabular}[c]{@{}c@{}}PP\_3/1 \\ \end{tabular}
& \begin{tabular}[c]{@{}c@{}}PP\_6/2 \\ \end{tabular}
& \begin{tabular}[c]{@{}c@{}}PP\_9/3 \\ \end{tabular} 
& Mean \\
\hline
\multirow{4}{*}{Natural}
& Vanilla QMIX & $165.4 \pm 1.3$ & $538.8\pm5.5$ & $661.9\pm5.4$ & $455.4 \pm 2.1$ \\
& RANDOM       & $178.3 \pm 1.0$ & $663.8\pm4.0$ & $666.9\pm3.5$ & $503.0 \pm 1.3$ \\
& ROMANCE      & $175.0 \pm 0.7$ & $648.6\pm3.9$ & $721.4\pm7.5$ & $515.0 \pm 1.4$ \\
& WALL (ours)  & $\mathbf{202.9 \pm 1.7}$ & $\mathbf{675.0\pm3.8}$ & $\mathbf{802.5\pm5.7}$ & $\mathbf{560.1 \pm 1.7}$ \\
\hline\multirow{4}{*}{\makecell[c]{Radom \\ Attack}} 
& Vanilla QMIX & $158.9\pm2.1$ & $522.6\pm2.4$ & $656.4\pm3.0$ & $445.9 \pm 2.0$ \\
& RANDOM       & $170.1\pm1.3$ & $638.7\pm2.0$ & $657.9\pm2.2$ & $488.9 \pm 1.9$ \\
& ROMANCE      & $173.4\pm1.4$ & $624.2\pm3.5$ & $701.7\pm4.0$ & $499.7 \pm 2.4$ \\
& WALL (ours)  & $\mathbf{202.9\pm1.7}$ & $\mathbf{654.0\pm2.1}$ & $\mathbf{802.5\pm2.1}$ & $\mathbf{553.1 \pm 1.7}$ \\
\hline
\multirow{4}{*}{EGA}
& Vanilla QMIX & $153.7\pm2.0$ & $499.8\pm0.6$ & $585.1\pm1.7$ & $412.8 \pm 1.4$ \\
& RANDOM       & $157.6\pm1.9$ & $604.6\pm3.2$ & $589.7\pm1.7$ & $450.6 \pm 1.4$ \\
& ROMANCE      & $167.1\pm2.4$ & $605.1\pm2.7$ & $594.4\pm1.3$ & $455.5 \pm 1.5$ \\
& WALL (ours)  & $\mathbf{166.5\pm1.7}$ & $\mathbf{685.5\pm5.0}$ & $\mathbf{682.7\pm0.8}$ & $\mathbf{511.5 \pm 1.4}$ \\
\hline
\multirow{4}{*}{\makecell[c]{Wolfpack\\ Adversarial \\ Attack (ours)}} 
& Vanilla QMIX & $135.7\pm1.5$ & $415.6\pm8.4$ & $551.7\pm7.0$ & $367.6 \pm 5.0$ \\
& RANDOM       & $157.5\pm1.4$ & $571.0\pm9.0$ & $624.3\pm7.7$ & $450.9 \pm 8.4$ \\
& ROMANCE      & $159.3\pm1.5$ & $554.6\pm9.9$ & $570.3\pm7.5$ & $428.0 \pm 9.4$ \\
& WALL (ours)  & $\mathbf{171.5\pm1.5}$ & $\mathbf{599.0\pm9.4}$ & $\mathbf{698.1\pm9.1}$ & $\mathbf{489.5 \pm 0.6}$ \\
\hline
\end{tabular}
}
\caption{Average cumulative rewards of robust MARL policies under various attack settings in the MPE environments.}
\vspace{-0.5in}
\label{table:mpe_perf}
\end{table}

\begin{table*}[t!]
\centering
\resizebox{0.94\textwidth}{!}{
\begin{tabular}{|c|c|c|c|c|c|c|c|c|}
\hline 
\multicolumn{2}{|c|}{\diagbox[innerwidth=5cm,dir=NW]{Method}{Scenario}}
& \begin{tabular}[c]{@{}c@{}}2s3z \end{tabular} 
& \begin{tabular}[c]{@{}c@{}}3m\end{tabular} 
& \begin{tabular}[c]{@{}c@{}}3s\_vs\_3z \end{tabular} 
& \begin{tabular}[c]{@{}c@{}}8m \end{tabular} 
& \begin{tabular}[c]{@{}c@{}}MMM \end{tabular} 
& \begin{tabular}[c]{@{}c@{}}1c3s5z  \end{tabular}  
& Mean \\ 
\hline
\multirow{7}{*}{Natural} 
& Vanilla QMIX & $98.0\pm1.5$ & $99.2\pm1.0$ & $99.2\pm1.6$ & $97.6\pm2.1$ & $99.2\pm0.5$ & $99.1\pm1.1$ & $98.7\pm0.5$ \\
& RANDOM & $\mathbf{99.7\pm0.5}$ & $99.1\pm1.2$ & $99.0\pm0.8$ & $99.2\pm1.0$ & $\mathbf{99.6\pm0.6}$ & $99.3\pm1.0$ & $99.3\pm0.2$ \\
& RARL & $97.8\pm2.0$ & $93.8\pm3.2$ & $93.1\pm17.4$ & $95.5\pm3.7$ & $90.6\pm20.8$ & $84.0\pm33.7$ & $92.5\pm5.3$ \\
& RAP & $98.8\pm1.3$ & $95.8\pm4.4$ & $99.5\pm1.0$ & $94.7\pm6.7$ & $95.5\pm12.1$ & $84.2\pm16.9$ & $94.7\pm2.6$ \\
& ERNIE & $98.2\pm1.3$ & $99.2\pm1.2$ & $99.8\pm0.4$ & $\mathbf{99.8\pm0.5}$ & $98.5\pm1.7$ & $99.2\pm1.0$ & $99.1\pm0.5$ \\
& ROMANCE & $96.4\pm2.9$ & $93.6\pm13.7$ & $99.7\pm0.5$ & $99.6\pm0.6$ & $96.4\pm6.6$ & $96.5\pm4.3$ & $97.0\pm2.1$ \\
& WALL (ours) & $99.4\pm0.6$ & $\mathbf{99.7\pm0.8}$ & $\mathbf{99.8\pm0.7}$ & $99.3\pm0.6$ & $99.0\pm2.1$ & $\mathbf{99.5\pm0.6}$ & $\mathbf{99.4\pm0.5}$ \\
\hline
\multirow{7}{*}{Random Attack} 
& Vanilla QMIX & $80.4\pm3.2$ & $69.6\pm10.4$ & $91.4\pm4.6$ & $69.0\pm7.1$ & $66.4\pm20.5$ & $94.8\pm3.4$ & $78.6\pm2.2$ \\
& RANDOM & $90.8\pm3.8$ & $76.4\pm17.3$ & $97.4\pm0.6$ & $80.2\pm4.9$ & $95.4\pm5.3$ & $96.0\pm2.9$ & $89.4\pm2.5$ \\
& RARL & $86.8\pm4.0$ & $58.7\pm15.4$ & $89.2\pm20.4$ & $70.2\pm5.8$ & $84.2\pm5.4$ & $79.1\pm22.2$ & $78.0\pm7.2$ \\
& RAP & $91.0\pm4.7$ & $69.2\pm11.1$ & $97.8\pm1.6$ & $85.0\pm11.4$ & $86.7\pm30.3$ & $86.6\pm12.5$ & $83.0\pm3.7$ \\
& ERNIE & $83.2\pm6.9$ & $65.2\pm4.9$ & $90.2\pm9.2$ & $76.2\pm11.8$ & $86.0\pm18.2$ & $95.6\pm3.2$ & $82.7\pm5.2$ \\
& ROMANCE & $90.2\pm2.3$ & $71.6\pm10.8$ & $99.6\pm0.6$ & $84.8\pm5.0$ & $86.8\pm16.3$ & $94.0\pm1.9$ & $87.8\pm2.4$ \\
& WALL (ours) & $\mathbf{94.6\pm4.5}$ & $\mathbf{87.4\pm1.8}$ & $\mathbf{99.8\pm0.5}$ & $\mathbf{95.8\pm3.4}$ & $\mathbf{99.4\pm1.1}$ & $\mathbf{98.6\pm1.6}$ & $\mathbf{95.9\pm0.5}$ \\
\hline
\multirow{7}{*}{EGA} 
& Vanilla QMIX & $54.0\pm7.6$ & $66.5\pm15.5$ & $72.4\pm15.1$ & $71.2\pm20.0$ & $70.6\pm14$ & $83.0\pm2.6$ & $69.6\pm3.8$ \\
& RANDOM & $65.3\pm3.3$ & $70.6\pm38.6$ & $68.8\pm23.5$ & $87.5\pm5.5$ & $84.4\pm3.1$ & $84.5\pm2.9$ & $76.9\pm5.4$ \\
& RARL & $62.6\pm9.4$ & $74.4\pm12.4$ & $88.4\pm17.7$ & $78.4\pm9.1$ & $83.4\pm11.9$ & $80.1\pm11.4$ & $77.9\pm10.0$ \\
& RAP & $70.4\pm13.0$ & $84.4\pm7.3$ & $83.8\pm15.8$ & $86.2\pm3.8$ & $83.9\pm16.4$ & $80.2\pm5.4$ & $81.5\pm4.3$ \\
& ERNIE & $52.4\pm9.6$ & $60.4\pm20.1$ & $83.2\pm9.7$ & $81.6\pm8.5$ & $85.0\pm4.6$ & $93.6\pm2.2$ & $76.0\pm3.0$ \\
& ROMANCE & $79.8\pm2.8$ & $85.8\pm4.6$ & $91.0\pm5.1$ & $90.9\pm4.0$ & $87.8\pm11.7$ & $89.6\pm2.9$ & $87.5\pm1.6$ \\
& WALL (ours) & $\mathbf{88.6\pm5.4}$ & $\mathbf{87.0\pm5.4}$ & $\mathbf{98.7\pm0.8}$ & $\mathbf{95.8\pm2.9}$ & $\mathbf{94.2\pm3.8}$ & $\mathbf{97.0\pm1.3}$ & $\mathbf{93.6\pm1.5}$ \\
\hline
\multirow{7}{*}{\makecell[c]{Wolfpack\\ Adversarial \\ Attack (ours)}} 
& Vanilla QMIX & $39.8\pm7.5$ & $31.0\pm11.8$ & $84.4\pm4.7$ & $11.4\pm13.3$ & $10.4\pm14.3$ & $59.2\pm4.2$ & $39.4\pm4.5$ \\
& RANDOM & $60.4\pm27.4$ & $57.4\pm30.5$ & $91.0\pm3.1$ & $40.4\pm14.8$ & $63.6\pm28.7$ & $68.4\pm19.6$ & $63.5\pm3.1$ \\
& RARL & $52.4\pm15.8$ & $31.1\pm20.3$ & $90.0\pm17.4$ & $14.2\pm9.7$ & $51.1\pm36.8$ & $75.9\pm13.6$ & $52.5\pm8.0$ \\
& RAP & $60.0\pm10.3$ & $37.5\pm10.4$ & $95.4\pm3.9$ & $35.6\pm14.4$ & $47.0\pm36.9$ & $75.7\pm26.1$ & $58.5\pm5.7$ \\
& ERNIE & $43.2\pm13.0$ & $35.4\pm6.2$ & $94.8\pm4.4$ & $26.4\pm10.4$ & $26.2\pm17.3$ & $77.0\pm9.3$ & $50.5\pm6.6$ \\
& ROMANCE & $62.4\pm5.1$ & $34.8\pm14.3$ & $98.6\pm0.6$ & $28.6\pm14.2$ & $48.8\pm17.9$ & $81.2\pm4.5$ & $59.1\pm2.0$ \\
& WALL (ours) & $\mathbf{92.2\pm3.7}$ & $\mathbf{90.8\pm4.9}$ & $\mathbf{99.8\pm0.5}$ & $\mathbf{83.6\pm5.0}$ & $\mathbf{95.0\pm4.5}$ & $\mathbf{98.8\pm1.6}$ & $\mathbf{93.4\pm1.1}$ \\
\hline
\end{tabular}}
\caption{Average test win rates of robust MARL policies under various attack settings in the SMAC environments.}
\vspace{-0.1in}
\label{table:perf}
\end{table*}

{\bf Adversarial Attacker Baselines:} 
To compare the severity of different attacks, we consider the following 4 scenarios: \textbf{Natural}, representing the case where no attacks are performed; \textbf{Random Attack}, where time steps, agents, and actions are randomly selected to execute attacks; \textbf{Evolutionary Generation of Attackers (EGA)} \cite{yuan2023robust}, which combines multiple single-agent-targeted attackers generated from various seeds as described in Section \ref{sec:background}; and the proposed \textbf{Wolfpack Adversarial Attack}. For a fair comparison, adversarial attackers are trained on independent seeds to execute unseen attacks.

{\bf Robust MARL Baselines:} To compare the severity of attack baselines and the robustness of policies trained under adversarial attack scenarios, we evaluate QMIX-trained policies under the following attack conditions: \textbf{Vanilla QMIX}, assuming no adversarial attacks; \textbf{RANDOM}, using Random Attack; \textbf{RARL} \cite{pinto2017robust}, where adversarial attackers tailor attacks to the learned policy; \textbf{RAP} \cite{vinitsky2020robust}, an extension of RARL that uniformly samples attackers to prevent overfitting and introduce diversity; \textbf{ROMANCE} \cite{yuan2023robust}, an RAP extension countering diverse EGA attacks; \textbf{ERNIE} \cite{bukharin2024robust}, enhancing robustness via adversarial regularization in observations and actions; and the proposed \textbf{WALL}. All robust MARL methods follow author-provided methodologies and parameters. Further details on the MARL baselines are available in Appendix \ref{appsec:marlbase}. All policies are trained for 3M timesteps, starting from a pretrained Vanilla QMIX model trained for 1M timesteps.

\begin{figure}[h!]
    \centering
    \includegraphics[width=0.49\textwidth]{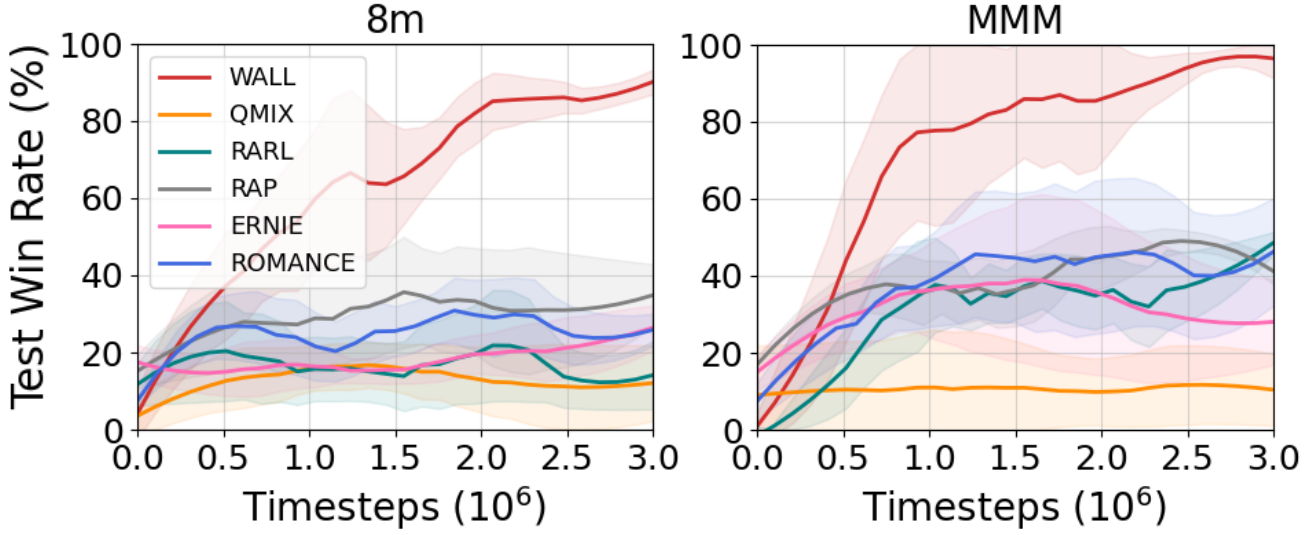} 
    \vspace{-0.2in}
    \caption{Learning curves of MARL methods for Wolfpack attack}
    \label{fig:learning_curves}
    \vspace{-0.25in}
\end{figure}

\subsection{Performance Comparison in MPE and SMAC}

Table~\ref{table:mpe_perf} presents the average cumulative rewards over the last 100 episodes under various attack settings in the MPE environments. The results show that the proposed Wolfpack adversarial attack is significantly more disruptive than existing methods such as EGA and Random Attack. For example, for Vanilla QMIX, the average cumulative reward across the three predator-prey scenarios drops by $455.4 - 367.6 = 87.8$ under the proposed Wolfpack attack, compared to $455.4 - 412.8 = 42.6$ under EGA and $455.4 - 445.9 = 9.5$ under Random Attack. These results demonstrate that the Wolfpack attack imposes a much more severe degradation in policy performance. In contrast, the proposed WALL framework consistently achieves the best performance across all attack types. Notably, in the MPE scenarios, WALL outperforms all baselines not only under adversarial attacks but also in the natural setting, demonstrating superior policy quality even without external threats.

\begin{figure*}[ht!]
    \begin{center}
    \centerline{\includegraphics[width=0.95\textwidth]{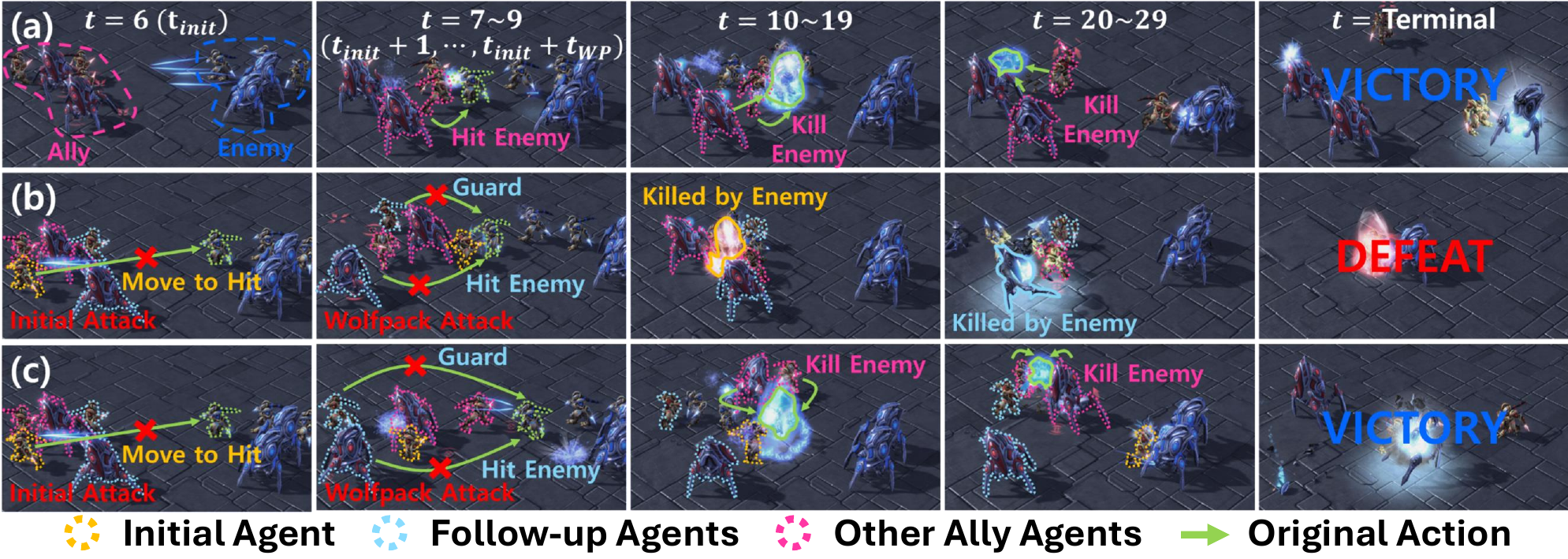}}
    \vspace{-0.1in}
    \caption{Attack comparison on 2s3z task in the SMAC: (a) QMIX/Natural, (b) QMIX/Wolfpack attack, and (c) WALL/Wolfpack attack}
    \label{fig:visualization}
    \vspace{-0.3in}
    \end{center}
\end{figure*}

For SMAC, Table~\ref{table:perf} presents the average win rates over the last 100 episodes of MARL policies under different attack baselines. The results show that the proposed Wolfpack adversarial attack is significantly more powerful than existing methods such as EGA and Random Attack. For example, EGA reduces the performance of Vanilla QMIX by $98.7 - 69.6 = 29.1\%$ and RANDOM by $99.3 - 76.9 = 22.4\%$ compared to the natural scenario. In contrast, the Wolfpack attack reduces Vanilla QMIX performance by $98.7 - 39.4 = 59.3\%$ and RANDOM by $99.3 - 63.5 = 35.8\%$, demonstrating its greater impact. 
In addition, the proposed WALL framework, which is trained to defend against the Wolfpack attack, outperforms other robust MARL methods under all attack types, showcasing its superior robustness. Notably, although RANDOM is trained specifically against Random Attack and ROMANCE against EGA, WALL achieves better performance against both attack types. These results highlight the effectiveness of WALL in enabling robust learning under diverse adversarial scenarios.
Fig. \ref{fig:learning_curves} further illustrates this in the \texttt{8m} and \texttt{MMM} environments, where performance differences with existing methods are most pronounced, showing the average win rate of each policy over training steps under unseen Wolfpack adversarial attacks. The results reveal that WALL not only achieves higher robustness but also adapts more quickly to attacks. Similar trends are observed for other CTDE algorithms, such as VDN and QPLEX, as detailed in Appendix \ref{appsubsec:compotherctde}, confirming the robustness of the proposed method.

To support a more practical evaluation, we assess computational complexity and general robustness under common perturbations in Appendix~\ref{appsubsec:computation_cost} and Appendix~\ref{appsubsec:general_robustness}, respectively. WALL incurs about 30\% higher training cost than ROMANCE but achieves substantially better performance due to its critical step selection. For robustness, we consider perturbations including Gaussian noise in observations and test-time parameter shifts such as reduced allied HP, under which WALL still outperforms existing baselines. These results demonstrate the practical effectiveness of WALL under both computational and environmental challenges.

\subsection{Visualization of Wolfpack Adversarial Attack}

To analyze the superior performance of the Wolfpack attack, we provide a visualization of its execution in the SMAC environment. Fig. \ref{fig:visualization} illustrates a scenario where the proposed step selector identifies $t=6$ as a critical initial step to initiate the attack. Prior to $t=6$, all setups are assumed to follow the same trajectory. Fig. \ref{fig:visualization}(a) shows Vanilla QMIX in a Natural scenario without attack, where our agents successfully defeat all enemy agents, achieving victory. Fig. \ref{fig:visualization}(b) demonstrates Vanilla QMIX under the Wolfpack adversarial attack, with follow-up agents targeted during $t=7$ to $t=9$. This leaves other agents unable to effectively defend against the adversarial attack, resulting in defeat as all agents are killed by the enemy. Fig. \ref{fig:visualization}(c) highlights a policy trained with the WALL framework. Despite the same follow-up agents are targeted during $t=7$ to $t=9$, WALL trains non-attacked agents to back up and protect the attacked agents, enabling ally agents to eliminate enemy agents and secure victory. This visualization demonstrates how the Wolfpack attack disrupts agent coordination and how the WALL framework robustly defends against such attacks. Visualizations of other SMAC tasks and detailed follow-up agent selection are provided in Appendix \ref{appsec:vis}.

\subsection{Ablation Study}

To evaluate the impact of each component and hyperparameter in the proposed Wolfpack adversarial attack, we conduct an ablation study focusing on the following aspects: component evaluation, step selection temperature $T$, and the number of follow-up agents $m$. The ablation study is conducted in the \texttt{8m} and \texttt{MMM} environments, where the performance differences are most pronounced. Additionally, more ablation studies on other hyperparameters, such as the total number of Wolfpack attacks $K_{\mathrm{WP}}$ and the attack duration $t_{\mathrm{WP}}$, are provided in Appendix \ref{appsec:addabl}.

{\bf Component Evaluation:} 
To evaluate the impact of each proposed component on attack severity and policy robustness, we consider five setups:    `Default', which uses all proposed components as designed; `Init. agent (min)', where the initial target agent $i$ is selected to minimize $Q^{tot}$, i.e., $i = \arg\min_j {\min_{a^j} Q^{tot}(s_{t_{\mathrm{init}}}, a_{t_{\mathrm{init}}}^j, \mathbf{a}_{t{\mathrm{init}}}^{-j})},$ instead of random selection; `Follow-up (L2)', which selects $m$ agents closest to the initial agent based on L2 distance instead of the proposed follow-up selection method; `Step (Random)', which uses random step selection instead of the proposed step selection method, while keeping the same total number of attacks; and `Agents \& Step (Random)', which randomly selects both $m$ follow-up agents and attack steps.

\begin{figure}[t!]
    \centering
    \begin{subfigure}{0.9\columnwidth} 
        \centering
        \includegraphics[width=1.0\textwidth]{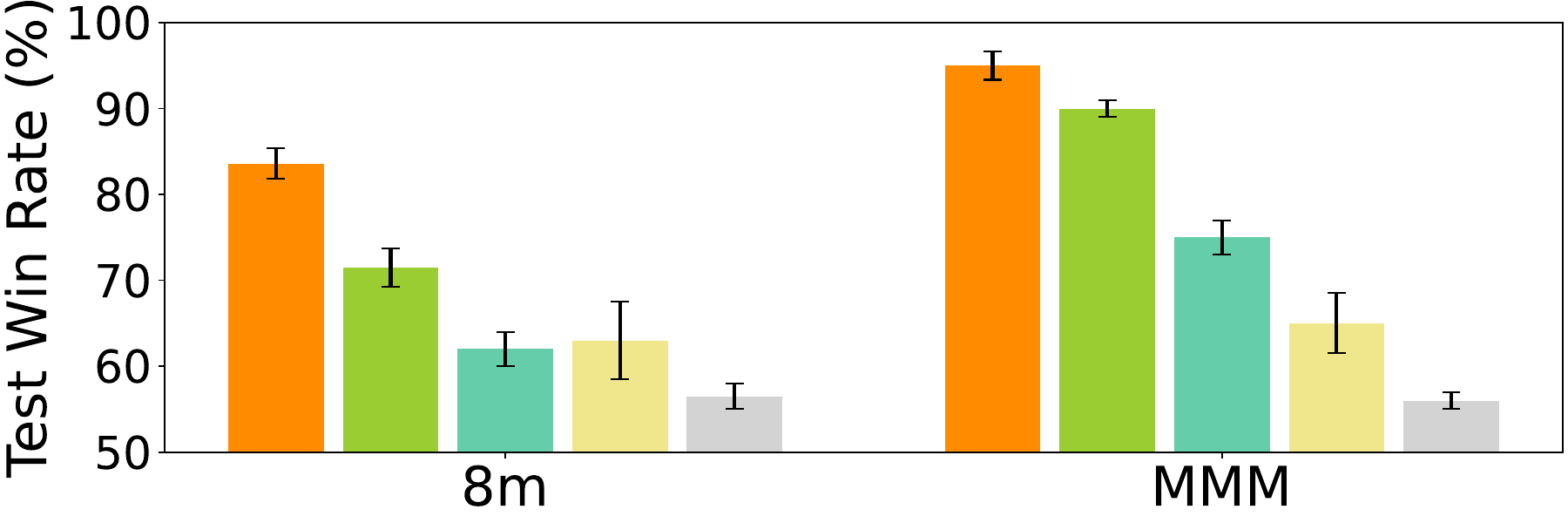} 
        \vspace{-0.25in}
        \caption{Robustness of WALL variations}  
    \end{subfigure}
    \vspace{0.1in}
    \begin{subfigure}{0.9\columnwidth} 
        \centering
        \includegraphics[width=1.0\textwidth]{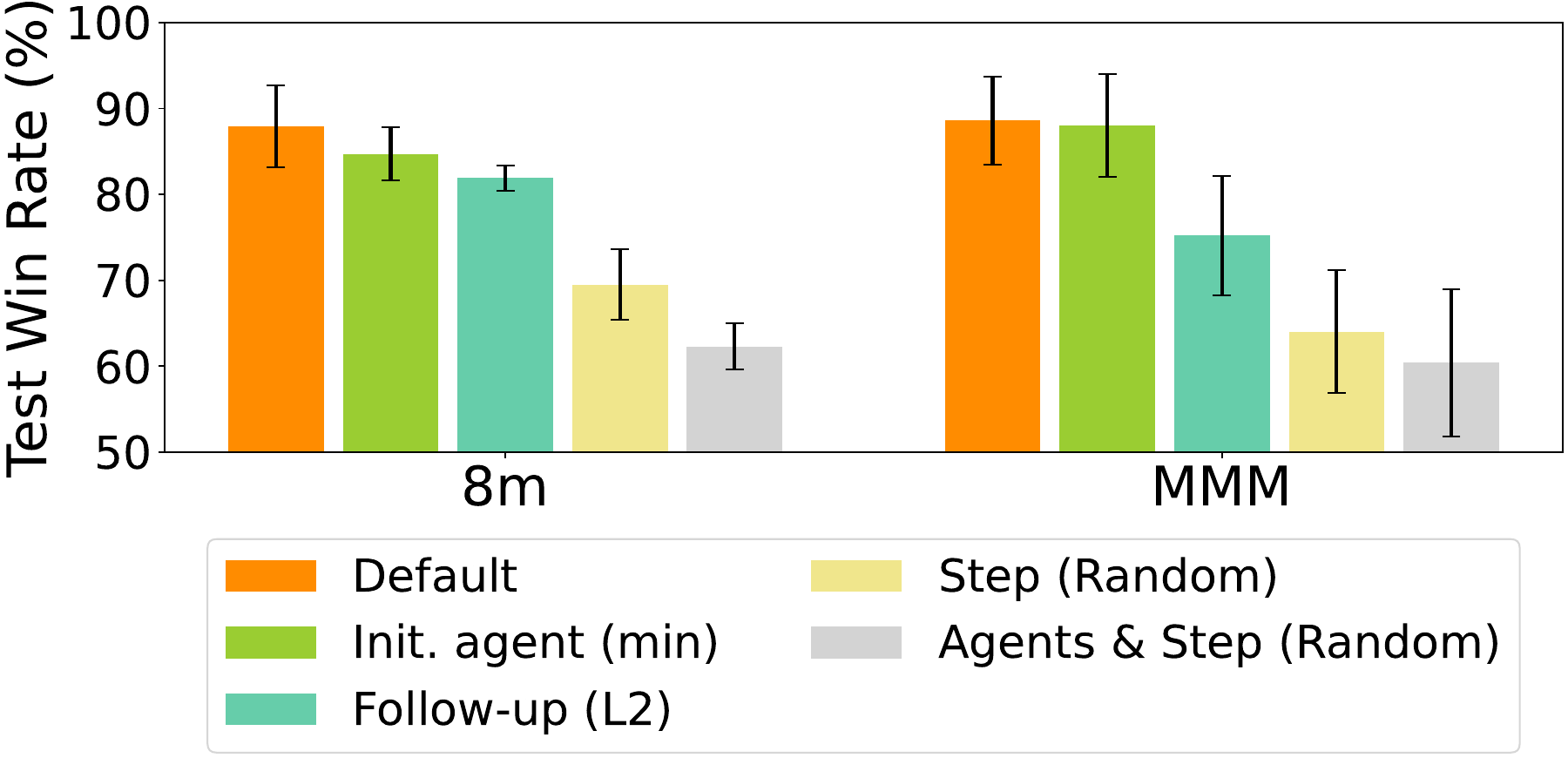} 
        \caption{Degradation of Vanilla QMIX under each attack setup}  
    \end{subfigure}
    \vspace{-0.2in}
    \caption{Component evaluation of our WALL/Wolfpack attack}
    \vspace{-0.2in}
    \label{fig:component}
\end{figure}

For each setup, we train the Wolfpack adversarial attack and the corresponding robust policy of WALL. Fig. \ref{fig:component}(a) shows the robustness of WALL trained under each setup when exposed to the default Wolfpack attack, while Fig. \ref{fig:component}(b) illustrates how each attack setup degrades the performance of `Vanilla QMIX' compared to its `Natural' performance. Randomly selecting the initial agent yields more robust policies than selecting the agent minimizing $Q^{tot}$ (`Init. agent (min)'), as random selection introduces diversity in attack scenarios. While selecting the $Q^{tot}$-minimizing agent may slightly enhance attack severity in cases like \texttt{MMM}, the added diversity from random selection generally improves robustness. Comparing `Default' and `Follow-up (L2)' shows that the proposed follow-up selection method enables more severe attacks and trains more robust policies than simply targeting agents closest to the initial agent. Similarly, `Default' outperforms `Step (Random)' in both attack severity and robustness, demonstrating that the proposed planner effectively identifies critical steps to minimize $Q^{tot}$, producing stronger policies. Finally, `Default' achieves significantly better robustness and more critical attacks compared to `Agents \& Step (Random)', highlighting the combined effectiveness of the proposed components.

\begin{figure}[t!]
    \centering
    \begin{subfigure}{0.45\columnwidth}
        \centering
        \includegraphics[width=\textwidth]{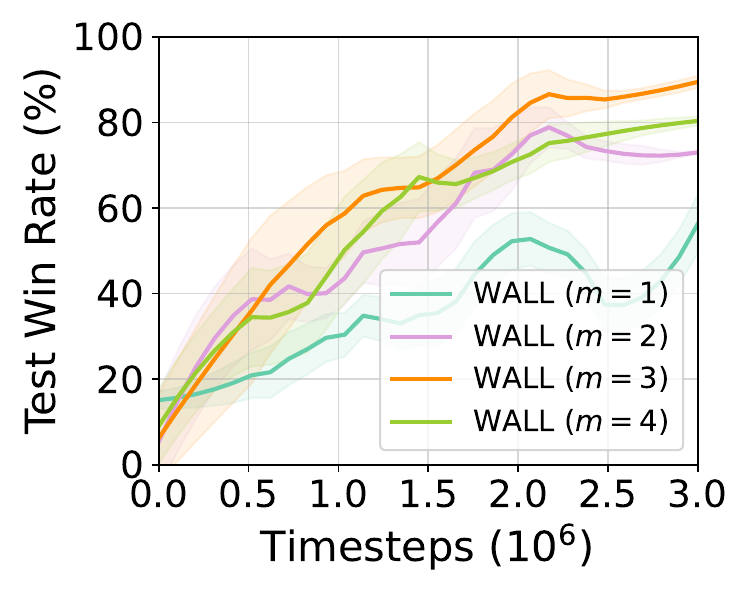}\vspace{-.5em}
        \caption{8m}
    \end{subfigure}
    \begin{subfigure}{0.45\columnwidth}
        \centering
        \includegraphics[width=\textwidth]{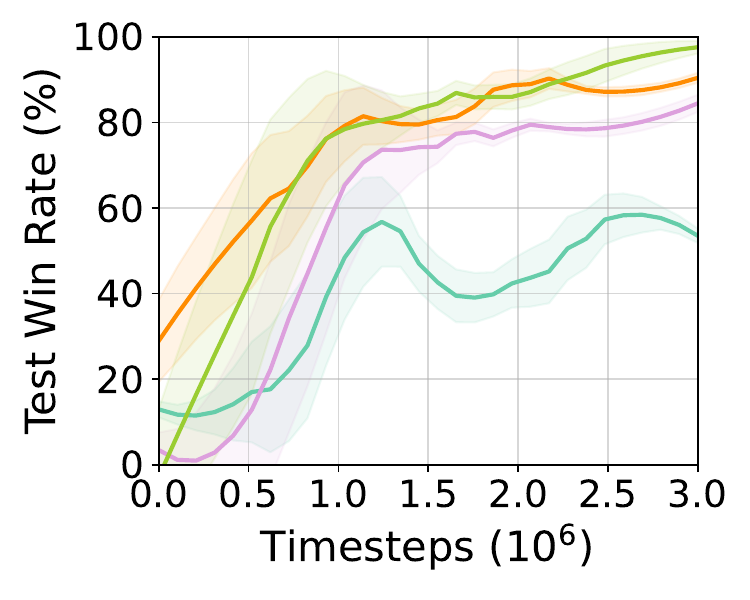}\vspace{-.5em}
        \caption{MMM}
    \end{subfigure}
    
    \vspace{-1em}
    \caption{Number of follow-up agents $m$}
    \label{fig:numfollow}
    \vspace{-1em}
\end{figure}

{\bf Number of Follow-up Agents $m$:} To analyze the impact of the hyperparameter $m$, which determines the number of follow-up agents, on robustness, Fig. \ref{fig:numfollow} shows how WALL trained with different values of $m$ defend against the default Wolfpack attack. To prevent excessive attack that could cause learning to fail, we assume $m < \left\lfloor \frac{n-1}{2} \right\rfloor$. The results indicate that in the \texttt{8m} environment, when $m$ is small, only a few agents defending against the initial attack are targeted, leading to reduced robustness. Conversely, when $m=4$, too many agents are attacked, causing learning to deteriorate. Therefore, $m=3$ yields the most robust performance and is considered the default hyperparameter. Similarly, in the \texttt{MMM} environment, $m=4$ results in the most robust performance and is set as the default. Notably, when $m=1$, the attack becomes a single-agent attack. As discussed in Section \ref{subsec:motiv}, performing coordinated multi-agent attack ($m>1$) enables much more robust learning, demonstrating the effectiveness and superiority of the proposed Wolfpack attack framework.

\begin{figure}[t!]
    \centering
    \begin{subfigure}{0.45\columnwidth}
        \centering
        \includegraphics[width=\textwidth]{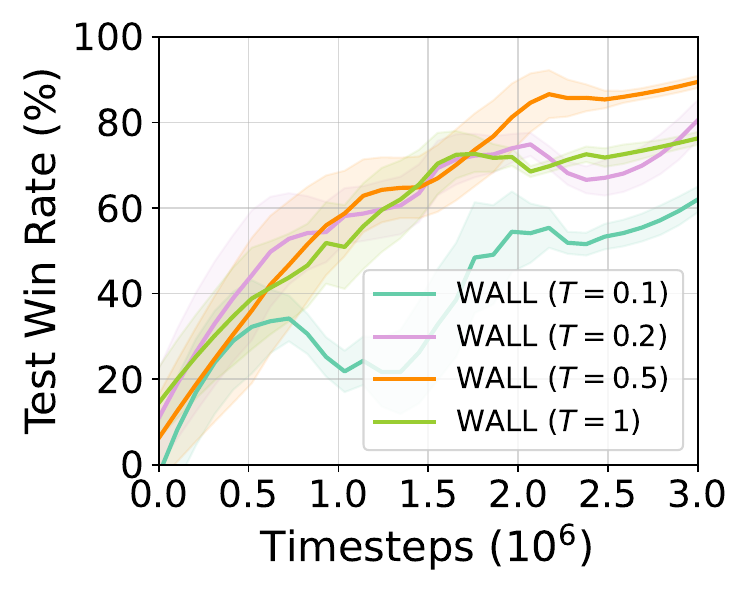}\vspace{-.5em}
        \caption{8m}
    \end{subfigure}
    \begin{subfigure}{0.45\columnwidth}
        \centering
        \includegraphics[width=\textwidth]{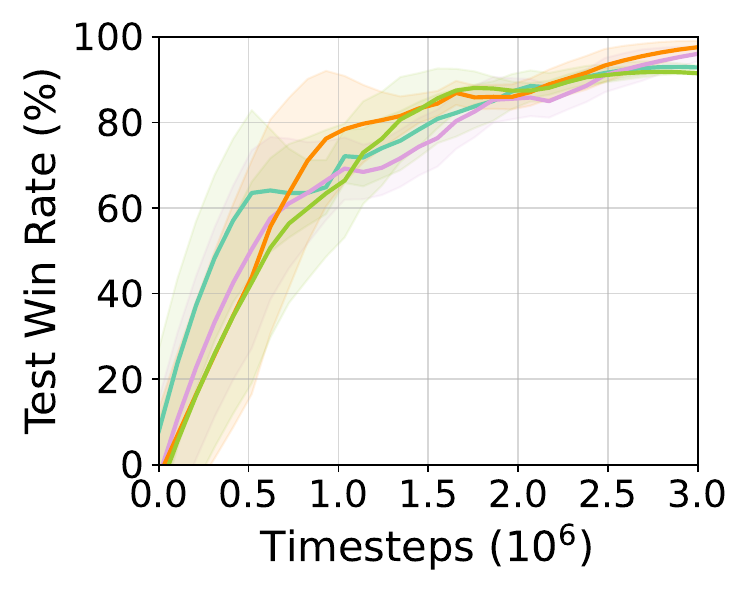}\vspace{-.5em}
        \caption{MMM}
    \end{subfigure}
    \vspace{-1em}
    \caption{Step selection temperature $T$}
    \vspace{-1em}
    \label{fig:temp}
\end{figure}

{\bf Step Selection Temperature $T$:} $T$ is a hyperparameter in Eq. \eqref{eq:initialprob} that controls the temperature of the initial attack probability. A larger $T$ results in more random attacks across steps, while a smaller $T$ focuses attacks on critical steps with higher probability. Fig. \ref{fig:temp} illustrates the performance of WALL policies trained with varying values of $T$ against the default Wolfpack attack. In both the \texttt{8m} and \texttt{MMM} environments, a very small $T$ causes the attack to target only specific steps, leading to policies that are less robust against diverse attacks. Conversely, a very large $T$ leads to overly uniform attacks, failing to target critical steps effectively, which also results in less robust policies. Based on these findings, we determined that $T = 0.5$ strikes a balance between targeting critical steps and maintaining robustness, and we set this as the default parameter.

\section{Limitations}
While the proposed WALL significantly improves robustness in MARL, it has a few limitations. The first is the additional computational overhead introduced by training the Transformer for identifying critical steps. However, as shown in our analysis, this overhead is justified given that other baselines fail to achieve comparable performance even with extended training. Another limitation is the need for hyperparameter tuning to construct the Wolfpack attack. Nevertheless, the method is not highly sensitive to these hyperparameters, and the provided ablation study offers practical guidelines for selecting appropriate configurations.

\section{Conclusions}

In this paper, we propose the Wolfpack adversarial attack, a coordinated strategy inspired by the Wolfpack tactic used in military operations, which significantly outperforms existing adversarial attacks. Additionally, we develop WALL, a robust MARL method designed to counter the proposed attack, demonstrating superior performance across various SMAC environments. Overall, our WALL framework enhances the robustness of MARL algorithms.


\section*{Acknowledgment}
This work was supported by Institute of Information \& Communications Technology Planning \& Evaluation (IITP) grant funded by the Korea government (MSIT) (No.RS-2022-II220469, Development of Core Technologies for Task-oriented Reinforcement Learning for Commercialization of Autonomous Drones), Innovative Human Resource Development for Local Intellectualization program through the Institute of Information \& Communications Technology Planning \& Evaluation (IITP) grant funded by the Korea government (MSIT) (IITP-2025-RS-2022-00156361), and Institute of Information \& Communications Technology Planning \& Evaluation (IITP) grant funded by the Korea government (MSIT) (No.RS-2020-II201336, Artificial Intelligence graduate school support (UNIST)).

\section*{Impact Statement}
This paper presents work whose goal is to advance the field of Machine Learning. There are many potential societal consequences of our work, none which we feel must be specifically highlighted here.

\bibliography{example_paper}

\begin{thebibliography}{85}
\providecommand{\natexlab}[1]{#1}
\providecommand{\url}[1]{\texttt{#1}}
\expandafter\ifx\csname urlstyle\endcsname\relax
  \providecommand{\doi}[1]{doi: #1}\else
  \providecommand{\doi}{doi: \begingroup \urlstyle{rm}\Url}\fi

\bibitem[Berkenkamp et~al.(2017)Berkenkamp, Turchetta, Schoellig, and Krause]{berkenkamp2017safe}
Berkenkamp, F., Turchetta, M., Schoellig, A., and Krause, A.
\newblock Safe model-based reinforcement learning with stability guarantees.
\newblock \emph{Advances in neural information processing systems}, 30, 2017.

\bibitem[Bhardwaj et~al.(2024)Bhardwaj, Xie, Boots, Jiang, and Cheng]{bhardwaj2024adversarial}
Bhardwaj, M., Xie, T., Boots, B., Jiang, N., and Cheng, C.-A.
\newblock Adversarial model for offline reinforcement learning.
\newblock \emph{Advances in Neural Information Processing Systems}, 36, 2024.

\bibitem[Bouhaddi \& Adi(2023)Bouhaddi and Adi]{bouhaddi2023multi}
Bouhaddi, M. and Adi, K.
\newblock Multi-environment training against reward poisoning attacks on deep reinforcement learning.
\newblock In \emph{SECRYPT}, pp.\  870--875, 2023.

\bibitem[Bouhaddi \& Adi(2024)Bouhaddi and Adi]{bouhaddi2024rewards}
Bouhaddi, M. and Adi, K.
\newblock When rewards deceive: Counteracting reward poisoning on online deep reinforcement learning.
\newblock In \emph{2024 IEEE International Conference on Cyber Security and Resilience (CSR)}, pp.\  38--44. IEEE, 2024.

\bibitem[Bukharin et~al.(2024)Bukharin, Li, Yu, Zhang, Chen, Zuo, Zhang, Zhang, and Zhao]{bukharin2024robust}
Bukharin, A., Li, Y., Yu, Y., Zhang, Q., Chen, Z., Zuo, S., Zhang, C., Zhang, S., and Zhao, T.
\newblock Robust multi-agent reinforcement learning via adversarial regularization: Theoretical foundation and stable algorithms.
\newblock \emph{Advances in Neural Information Processing Systems}, 36, 2024.

\bibitem[Cai et~al.(2023)Cai, Zhu, and Hu]{cai2023reward}
Cai, K., Zhu, X., and Hu, Z.
\newblock Reward poisoning attacks in deep reinforcement learning based on exploration strategies.
\newblock \emph{Neurocomputing}, 553:\penalty0 126578, 2023.

\bibitem[Chae et~al.(2022)Chae, Han, Jung, Cho, Choi, and Sung]{chae2022robust}
Chae, J., Han, S., Jung, W., Cho, M., Choi, S., and Sung, Y.
\newblock Robust imitation learning against variations in environment dynamics.
\newblock In \emph{International Conference on Machine Learning}, pp.\  2828--2852. PMLR, 2022.

\bibitem[Chen et~al.(2023)Chen, Ma, and Oyekan]{chen2023deep}
Chen, J., Ma, R., and Oyekan, J.
\newblock A deep multi-agent reinforcement learning framework for autonomous aerial navigation to grasping points on loads.
\newblock \emph{Robotics and Autonomous Systems}, 167:\penalty0 104489, 2023.

\bibitem[Chen et~al.(2021)Chen, Lu, Rajeswaran, Lee, Grover, Laskin, Abbeel, Srinivas, and Mordatch]{chen2021decision}
Chen, L., Lu, K., Rajeswaran, A., Lee, K., Grover, A., Laskin, M., Abbeel, P., Srinivas, A., and Mordatch, I.
\newblock Decision transformer: Reinforcement learning via sequence modeling.
\newblock \emph{Advances in neural information processing systems}, 34:\penalty0 15084--15097, 2021.

\bibitem[Chinchuluun et~al.(2008)Chinchuluun, Migdalas, Pardalos, and Pitsoulis]{chinchuluun2008pareto}
Chinchuluun, A., Migdalas, A., Pardalos, P.~M., and Pitsoulis, L.
\newblock \emph{Pareto optimality, game theory and equilibria}, volume~17.
\newblock Springer New York, 2008.

\bibitem[Clavier et~al.(2023)Clavier, Pennec, and Geist]{clavier2023towards}
Clavier, P., Pennec, E.~L., and Geist, M.
\newblock Towards minimax optimality of model-based robust reinforcement learning.
\newblock \emph{arXiv preprint arXiv:2302.05372}, 2023.

\bibitem[Curi et~al.(2021)Curi, Bogunovic, and Krause]{curi2021combining}
Curi, S., Bogunovic, I., and Krause, A.
\newblock Combining pessimism with optimism for robust and efficient model-based deep reinforcement learning.
\newblock In \emph{International Conference on Machine Learning}, pp.\  2254--2264. PMLR, 2021.

\bibitem[Du et~al.(2024)Du, Chen, Wang, Xing, Yang, Philip, Chang, and He]{du2024robust}
Du, X., Chen, H., Wang, C., Xing, Y., Yang, J., Philip, S.~Y., Chang, Y., and He, L.
\newblock Robust multi-agent reinforcement learning via bayesian distributional value estimation.
\newblock \emph{Pattern Recognition}, 145:\penalty0 109917, 2024.

\bibitem[Everett et~al.(2021)Everett, L{\"u}tjens, and How]{everett2021certifiable}
Everett, M., L{\"u}tjens, B., and How, J.~P.
\newblock Certifiable robustness to adversarial state uncertainty in deep reinforcement learning.
\newblock \emph{IEEE Transactions on Neural Networks and Learning Systems}, 33\penalty0 (9):\penalty0 4184--4198, 2021.

\bibitem[Geng et~al.(2024)Geng, Xiao, Li, Wei, Wang, and Zhao]{geng2024noise}
Geng, W., Xiao, B., Li, R., Wei, N., Wang, D., and Zhao, Z.
\newblock Noise distribution decomposition based multi-agent distributional reinforcement learning.
\newblock \emph{IEEE Transactions on Mobile Computing}, 2024.

\bibitem[Goodfellow et~al.(2014)Goodfellow, Shlens, and Szegedy]{goodfellow2014explaining}
Goodfellow, I.~J., Shlens, J., and Szegedy, C.
\newblock Explaining and harnessing adversarial examples.
\newblock \emph{arXiv preprint arXiv:1412.6572}, 2014.

\bibitem[Guo et~al.(2022)Guo, Chen, Hao, Yin, Yu, and Li]{guo2022towards}
Guo, J., Chen, Y., Hao, Y., Yin, Z., Yu, Y., and Li, S.
\newblock Towards comprehensive testing on the robustness of cooperative multi-agent reinforcement learning.
\newblock In \emph{Proceedings of the IEEE/CVF conference on computer vision and pattern recognition}, pp.\  115--122, 2022.

\bibitem[Han \& Sung(2021)Han and Sung]{han2021max}
Han, S. and Sung, Y.
\newblock A max-min entropy framework for reinforcement learning.
\newblock \emph{Advances in Neural Information Processing Systems}, 34:\penalty0 25732--25745, 2021.

\bibitem[Han et~al.(2022)Han, Su, He, Han, Yang, Zou, and Miao]{han2022solution}
Han, S., Su, S., He, S., Han, S., Yang, H., Zou, S., and Miao, F.
\newblock What is the solution for state-adversarial multi-agent reinforcement learning?
\newblock \emph{arXiv preprint arXiv:2212.02705}, 2022.

\bibitem[He et~al.(2022)He, Wang, Han, Zou, and Miao]{he2022robust}
He, S., Wang, Y., Han, S., Zou, S., and Miao, F.
\newblock A robust and constrained multi-agent reinforcement learning framework for electric vehicle amod systems.
\newblock \emph{Dynamics}, 8\penalty0 (10), 2022.

\bibitem[He et~al.(2023)He, Han, Su, Han, Zou, and Miao]{he2023robust}
He, S., Han, S., Su, S., Han, S., Zou, S., and Miao, F.
\newblock Robust multi-agent reinforcement learning with state uncertainty.
\newblock \emph{arXiv preprint arXiv:2307.16212}, 2023.

\bibitem[Herremans et~al.(2024)Herremans, Anwar, and Mercelis]{herremans2024robust}
Herremans, S., Anwar, A., and Mercelis, S.
\newblock Robust model-based reinforcement learning with an adversarial auxiliary model.
\newblock \emph{arXiv preprint arXiv:2406.09976}, 2024.

\bibitem[Huang et~al.(2017)Huang, Papernot, Goodfellow, Duan, and Abbeel]{huang2017adversarial}
Huang, S., Papernot, N., Goodfellow, I., Duan, Y., and Abbeel, P.
\newblock Adversarial attacks on neural network policies.
\newblock \emph{arXiv preprint arXiv:1702.02284}, 2017.

\bibitem[Jendoubi \& Bouffard(2023)Jendoubi and Bouffard]{jendoubi2023multi}
Jendoubi, I. and Bouffard, F.
\newblock Multi-agent hierarchical reinforcement learning for energy management.
\newblock \emph{Applied Energy}, 332:\penalty0 120500, 2023.

\bibitem[Jo et~al.(2024)Jo, Lee, Yeom, and Han]{jo2024fox}
Jo, Y., Lee, S., Yeom, J., and Han, S.
\newblock Fox: Formation-aware exploration in multi-agent reinforcement learning.
\newblock In \emph{Proceedings of the AAAI Conference on Artificial Intelligence}, volume~38, pp.\  12985--12994, 2024.

\bibitem[Karde{\c{s}} et~al.(2011)Karde{\c{s}}, Ord{\'o}{\~n}ez, and Hall]{kardecs2011discounted}
Karde{\c{s}}, E., Ord{\'o}{\~n}ez, F., and Hall, R.~W.
\newblock Discounted robust stochastic games and an application to queueing control.
\newblock \emph{Operations research}, 59\penalty0 (2):\penalty0 365--382, 2011.

\bibitem[Kobayashi(2024)]{kobayashi2024lira}
Kobayashi, T.
\newblock Lira: Light-robust adversary for model-based reinforcement learning in real world.
\newblock \emph{arXiv preprint arXiv:2409.19617}, 2024.

\bibitem[Lee et~al.(2021)Lee, Esfandiari, Tan, and Sarkar]{lee2021query}
Lee, X.~Y., Esfandiari, Y., Tan, K.~L., and Sarkar, S.
\newblock Query-based targeted action-space adversarial policies on deep reinforcement learning agents.
\newblock In \emph{Proceedings of the ACM/IEEE 12th international conference on cyber-physical systems}, pp.\  87--97, 2021.

\bibitem[Li et~al.(2020)Li, Wang, Tian, Jia, and Zheng]{li2020multi}
Li, R., Wang, R., Tian, T., Jia, F., and Zheng, Z.
\newblock Multi-agent reinforcement learning based on value distribution.
\newblock In \emph{Journal of Physics: Conference Series}, volume 1651, pp.\  012017. IOP Publishing, 2020.

\bibitem[Li et~al.(2019)Li, Wu, Cui, Dong, Fang, and Russell]{li2019robust}
Li, S., Wu, Y., Cui, X., Dong, H., Fang, F., and Russell, S.
\newblock Robust multi-agent reinforcement learning via minimax deep deterministic policy gradient.
\newblock In \emph{Proceedings of the AAAI conference on artificial intelligence}, volume~33, pp.\  4213--4220, 2019.

\bibitem[Li et~al.(2023{\natexlab{a}})Li, Guo, Xiu, Xu, Yu, Wang, Liu, Yang, and Liu]{li2023byzantine}
Li, S., Guo, J., Xiu, J., Xu, R., Yu, X., Wang, J., Liu, A., Yang, Y., and Liu, X.
\newblock Byzantine robust cooperative multi-agent reinforcement learning as a bayesian game.
\newblock \emph{arXiv preprint arXiv:2305.12872}, 2023{\natexlab{a}}.

\bibitem[Li et~al.(2023{\natexlab{b}})Li, Xu, Guo, Feng, Wang, Liu, Yang, Liu, and Lv]{li2023mir2}
Li, S., Xu, R., Guo, J., Feng, P., Wang, J., Liu, A., Yang, Y., Liu, X., and Lv, W.
\newblock Mir2: Towards provably robust multi-agent reinforcement learning by mutual information regularization.
\newblock \emph{arXiv preprint arXiv:2310.09833}, 2023{\natexlab{b}}.

\bibitem[Li et~al.(2023{\natexlab{c}})Li, Li, Feng, Wang, and Pan]{li2023ats}
Li, X., Li, Y., Feng, Z., Wang, Z., and Pan, Q.
\newblock Ats-o2a: A state-based adversarial attack strategy on deep reinforcement learning.
\newblock \emph{Computers \& Security}, 129:\penalty0 103259, 2023{\natexlab{c}}.

\bibitem[Lin et~al.(2020)Lin, Dzeparoska, Zhang, Leon-Garcia, and Papernot]{lin2020robustness}
Lin, J., Dzeparoska, K., Zhang, S.~Q., Leon-Garcia, A., and Papernot, N.
\newblock On the robustness of cooperative multi-agent reinforcement learning.
\newblock In \emph{2020 IEEE Security and Privacy Workshops (SPW)}, pp.\  62--68. IEEE, 2020.

\bibitem[Liu et~al.(2024)Liu, Kuang, and Wang]{liu2024robust}
Liu, Q., Kuang, Y., and Wang, J.
\newblock Robust deep reinforcement learning with adaptive adversarial perturbations in action space.
\newblock \emph{arXiv preprint arXiv:2405.11982}, 2024.

\bibitem[Lowe et~al.(2017)Lowe, Wu, Tamar, Harb, Pieter~Abbeel, and Mordatch]{lowe2017multi}
Lowe, R., Wu, Y.~I., Tamar, A., Harb, J., Pieter~Abbeel, O., and Mordatch, I.
\newblock Multi-agent actor-critic for mixed cooperative-competitive environments.
\newblock \emph{Advances in neural information processing systems}, 30, 2017.

\bibitem[Mahajan et~al.(2019)Mahajan, Rashid, Samvelyan, and Whiteson]{mahajan2019maven}
Mahajan, A., Rashid, T., Samvelyan, M., and Whiteson, S.
\newblock Maven: Multi-agent variational exploration.
\newblock \emph{Advances in neural information processing systems}, 32, 2019.

\bibitem[Mankowitz et~al.(2019)Mankowitz, Levine, Jeong, Shi, Kay, Abdolmaleki, Springenberg, Mann, Hester, and Riedmiller]{mankowitz2019robust}
Mankowitz, D.~J., Levine, N., Jeong, R., Shi, Y., Kay, J., Abdolmaleki, A., Springenberg, J.~T., Mann, T., Hester, T., and Riedmiller, M.
\newblock Robust reinforcement learning for continuous control with model misspecification.
\newblock \emph{arXiv preprint arXiv:1906.07516}, 2019.

\bibitem[Moos et~al.(2022)Moos, Hansel, Abdulsamad, Stark, Clever, and Peters]{moos2022robust}
Moos, J., Hansel, K., Abdulsamad, H., Stark, S., Clever, D., and Peters, J.
\newblock Robust reinforcement learning: A review of foundations and recent advances.
\newblock \emph{Machine Learning and Knowledge Extraction}, 4\penalty0 (1):\penalty0 276--315, 2022.

\bibitem[Oliehoek et~al.(2008)Oliehoek, Spaan, and Vlassis]{oliehoek2008optimal}
Oliehoek, F.~A., Spaan, M.~T., and Vlassis, N.
\newblock Optimal and approximate q-value functions for decentralized pomdps.
\newblock \emph{Journal of Artificial Intelligence Research}, 32:\penalty0 289--353, 2008.

\bibitem[Oliehoek et~al.(2016)Oliehoek, Amato, et~al.]{oliehoek2016concise}
Oliehoek, F.~A., Amato, C., et~al.
\newblock \emph{A concise introduction to decentralized POMDPs}, volume~1.
\newblock Springer, 2016.

\bibitem[Oroojlooy \& Hajinezhad(2023)Oroojlooy and Hajinezhad]{oroojlooy2023review}
Oroojlooy, A. and Hajinezhad, D.
\newblock A review of cooperative multi-agent deep reinforcement learning.
\newblock \emph{Applied Intelligence}, 53\penalty0 (11):\penalty0 13677--13722, 2023.

\bibitem[Orr \& Dutta(2023)Orr and Dutta]{orr2023multi}
Orr, J. and Dutta, A.
\newblock Multi-agent deep reinforcement learning for multi-robot applications: A survey.
\newblock \emph{Sensors}, 23\penalty0 (7):\penalty0 3625, 2023.

\bibitem[Panaganti \& Kalathil(2021)Panaganti and Kalathil]{panaganti2021sample}
Panaganti, K. and Kalathil, D.
\newblock Sample complexity of model-based robust reinforcement learning.
\newblock In \emph{2021 60th IEEE Conference on Decision and Control (CDC)}, pp.\  2240--2245. IEEE, 2021.

\bibitem[Pattanaik et~al.(2017)Pattanaik, Tang, Liu, Bommannan, and Chowdhary]{pattanaik2017robust}
Pattanaik, A., Tang, Z., Liu, S., Bommannan, G., and Chowdhary, G.
\newblock Robust deep reinforcement learning with adversarial attacks.
\newblock \emph{arXiv preprint arXiv:1712.03632}, 2017.

\bibitem[Phan et~al.(2021)Phan, Belzner, Gabor, Sedlmeier, Ritz, and Linnhoff-Popien]{phan2021resilient}
Phan, T., Belzner, L., Gabor, T., Sedlmeier, A., Ritz, F., and Linnhoff-Popien, C.
\newblock Resilient multi-agent reinforcement learning with adversarial value decomposition.
\newblock In \emph{Proceedings of the AAAI Conference on Artificial Intelligence}, volume~35, pp.\  11308--11316, 2021.

\bibitem[Pinto et~al.(2017)Pinto, Davidson, Sukthankar, and Gupta]{pinto2017robust}
Pinto, L., Davidson, J., Sukthankar, R., and Gupta, A.
\newblock Robust adversarial reinforcement learning.
\newblock In \emph{International conference on machine learning}, pp.\  2817--2826. PMLR, 2017.

\bibitem[Qiaoben et~al.(2024)Qiaoben, Ying, Zhou, Su, Zhu, and Zhang]{qiaoben2024understanding}
Qiaoben, Y., Ying, C., Zhou, X., Su, H., Zhu, J., and Zhang, B.
\newblock Understanding adversarial attacks on observations in deep reinforcement learning.
\newblock \emph{Science China Information Sciences}, 67\penalty0 (5):\penalty0 1--15, 2024.

\bibitem[Rakhsha et~al.(2021)Rakhsha, Zhang, Zhu, and Singla]{rakhsha2021reward}
Rakhsha, A., Zhang, X., Zhu, X., and Singla, A.
\newblock Reward poisoning in reinforcement learning: Attacks against unknown learners in unknown environments.
\newblock \emph{arXiv preprint arXiv:2102.08492}, 2021.

\bibitem[Ramesh et~al.(2024)Ramesh, Sessa, Hu, Krause, and Bogunovic]{ramesh2024distributionally}
Ramesh, S.~S., Sessa, P.~G., Hu, Y., Krause, A., and Bogunovic, I.
\newblock Distributionally robust model-based reinforcement learning with large state spaces.
\newblock In \emph{International Conference on Artificial Intelligence and Statistics}, pp.\  100--108. PMLR, 2024.

\bibitem[Rashid et~al.(2020)Rashid, Samvelyan, De~Witt, Farquhar, Foerster, and Whiteson]{rashid2020monotonic}
Rashid, T., Samvelyan, M., De~Witt, C.~S., Farquhar, G., Foerster, J., and Whiteson, S.
\newblock Monotonic value function factorisation for deep multi-agent reinforcement learning.
\newblock \emph{Journal of Machine Learning Research}, 21\penalty0 (178):\penalty0 1--51, 2020.

\bibitem[Rigter et~al.(2022)Rigter, Lacerda, and Hawes]{rigter2022rambo}
Rigter, M., Lacerda, B., and Hawes, N.
\newblock Rambo-rl: Robust adversarial model-based offline reinforcement learning.
\newblock \emph{Advances in neural information processing systems}, 35:\penalty0 16082--16097, 2022.

\bibitem[Samvelyan et~al.(2019)Samvelyan, Rashid, De~Witt, Farquhar, Nardelli, Rudner, Hung, Torr, Foerster, and Whiteson]{samvelyan2019starcraft}
Samvelyan, M., Rashid, T., De~Witt, C.~S., Farquhar, G., Nardelli, N., Rudner, T.~G., Hung, C.-M., Torr, P.~H., Foerster, J., and Whiteson, S.
\newblock The starcraft multi-agent challenge.
\newblock \emph{arXiv preprint arXiv:1902.04043}, 2019.

\bibitem[Shi \& Chi(2024)Shi and Chi]{shi2024distributionally}
Shi, L. and Chi, Y.
\newblock Distributionally robust model-based offline reinforcement learning with near-optimal sample complexity.
\newblock \emph{Journal of Machine Learning Research}, 25\penalty0 (200):\penalty0 1--91, 2024.

\bibitem[Shi et~al.(2024)Shi, Mazumdar, Chi, and Wierman]{shi2024sample}
Shi, L., Mazumdar, E., Chi, Y., and Wierman, A.
\newblock Sample-efficient robust multi-agent reinforcement learning in the face of environmental uncertainty.
\newblock \emph{arXiv preprint arXiv:2404.18909}, 2024.

\bibitem[Sun et~al.(2023)Sun, Zheng, Hassanzadeh, Liang, Feizi, Ganesh, and Huang]{sun2023certifiably}
Sun, Y., Zheng, R., Hassanzadeh, P., Liang, Y., Feizi, S., Ganesh, S., and Huang, F.
\newblock Certifiably robust policy learning against adversarial multi-agent communication.
\newblock In \emph{The Eleventh International Conference on Learning Representations}, 2023.

\bibitem[Sunehag et~al.(2017)Sunehag, Lever, Gruslys, Czarnecki, Zambaldi, Jaderberg, Lanctot, Sonnerat, Leibo, Tuyls, et~al.]{sunehag2017value}
Sunehag, P., Lever, G., Gruslys, A., Czarnecki, W.~M., Zambaldi, V., Jaderberg, M., Lanctot, M., Sonnerat, N., Leibo, J.~Z., Tuyls, K., et~al.
\newblock Value-decomposition networks for cooperative multi-agent learning.
\newblock \emph{arXiv preprint arXiv:1706.05296}, 2017.

\bibitem[Tan et~al.(2020)Tan, Esfandiari, Lee, Sarkar, et~al.]{tan2020robustifying}
Tan, K.~L., Esfandiari, Y., Lee, X.~Y., Sarkar, S., et~al.
\newblock Robustifying reinforcement learning agents via action space adversarial training.
\newblock In \emph{2020 American control conference (ACC)}, pp.\  3959--3964. IEEE, 2020.

\bibitem[Terry et~al.(2021)Terry, Black, Grammel, Jayakumar, Hari, Sullivan, Santos, Dieffendahl, Horsch, Perez-Vicente, et~al.]{terry2021pettingzoo}
Terry, J., Black, B., Grammel, N., Jayakumar, M., Hari, A., Sullivan, R., Santos, L.~S., Dieffendahl, C., Horsch, C., Perez-Vicente, R., et~al.
\newblock Pettingzoo: Gym for multi-agent reinforcement learning.
\newblock \emph{Advances in Neural Information Processing Systems}, 34:\penalty0 15032--15043, 2021.

\bibitem[Tessler et~al.(2019)Tessler, Efroni, and Mannor]{tessler2019action}
Tessler, C., Efroni, Y., and Mannor, S.
\newblock Action robust reinforcement learning and applications in continuous control.
\newblock In \emph{International Conference on Machine Learning}, pp.\  6215--6224. PMLR, 2019.

\bibitem[Tu et~al.(2021)Tu, Wang, Wang, Manivasagam, Ren, and Urtasun]{tu2021adversarial}
Tu, J., Wang, T., Wang, J., Manivasagam, S., Ren, M., and Urtasun, R.
\newblock Adversarial attacks on multi-agent communication.
\newblock In \emph{Proceedings of the IEEE/CVF International Conference on Computer Vision}, pp.\  7768--7777, 2021.

\bibitem[Vaswani(2017)]{vaswani2017attention}
Vaswani, A.
\newblock Attention is all you need.
\newblock \emph{Advances in Neural Information Processing Systems}, 2017.

\bibitem[Vinitsky et~al.(2020)Vinitsky, Du, Parvate, Jang, Abbeel, and Bayen]{vinitsky2020robust}
Vinitsky, E., Du, Y., Parvate, K., Jang, K., Abbeel, P., and Bayen, A.
\newblock Robust reinforcement learning using adversarial populations.
\newblock \emph{arXiv preprint arXiv:2008.01825}, 2020.

\bibitem[Wang et~al.(2020{\natexlab{a}})Wang, Liu, and Li]{wang2020reinforcement}
Wang, J., Liu, Y., and Li, B.
\newblock Reinforcement learning with perturbed rewards.
\newblock In \emph{Proceedings of the AAAI conference on artificial intelligence}, volume~34, pp.\  6202--6209, 2020{\natexlab{a}}.

\bibitem[Wang et~al.(2020{\natexlab{b}})Wang, Ren, Liu, Yu, and Zhang]{wang2020qplex}
Wang, J., Ren, Z., Liu, T., Yu, Y., and Zhang, C.
\newblock Qplex: Duplex dueling multi-agent q-learning.
\newblock \emph{arXiv preprint arXiv:2008.01062}, 2020{\natexlab{b}}.

\bibitem[Wang et~al.(2023)Wang, Chen, Huang, Zhang, Zhao, and Qu]{wang2023regularization}
Wang, S., Chen, W., Huang, L., Zhang, F., Zhao, Z., and Qu, H.
\newblock Regularization-adapted anderson acceleration for multi-agent reinforcement learning.
\newblock \emph{Knowledge-Based Systems}, 275:\penalty0 110709, 2023.

\bibitem[Wang et~al.(2020{\natexlab{c}})Wang, Nair, and Althoff]{wang2020falsification}
Wang, X., Nair, S., and Althoff, M.
\newblock Falsification-based robust adversarial reinforcement learning.
\newblock In \emph{2020 19th IEEE International Conference on Machine Learning and Applications (ICMLA)}, pp.\  205--212. IEEE, 2020{\natexlab{c}}.

\bibitem[Wang et~al.(2022)Wang, Wang, Zhou, Velasquez, and Zou]{wang2022data}
Wang, Y., Wang, Y., Zhou, Y., Velasquez, A., and Zou, S.
\newblock Data-driven robust multi-agent reinforcement learning.
\newblock In \emph{2022 IEEE 32nd International Workshop on Machine Learning for Signal Processing (MLSP)}, pp.\  1--6. IEEE, 2022.

\bibitem[Xu et~al.(2022)Xu, Zeng, and Singh]{xu2022efficient}
Xu, Y., Zeng, Q., and Singh, G.
\newblock Efficient reward poisoning attacks on online deep reinforcement learning.
\newblock \emph{arXiv preprint arXiv:2205.14842}, 2022.

\bibitem[Xu et~al.(2024)Xu, Gumaste, and Singh]{xu2024reward}
Xu, Y., Gumaste, R., and Singh, G.
\newblock Reward poisoning attack against offline reinforcement learning.
\newblock \emph{arXiv preprint arXiv:2402.09695}, 2024.

\bibitem[Xu et~al.(2021)Xu, Li, Bai, and Fan]{xu2021mmd}
Xu, Z., Li, D., Bai, Y., and Fan, G.
\newblock Mmd-mix: Value function factorisation with maximum mean discrepancy for cooperative multi-agent reinforcement learning.
\newblock In \emph{2021 International Joint Conference on Neural Networks (IJCNN)}, pp.\  1--7. IEEE, 2021.

\bibitem[Xue et~al.(2021)Xue, Qiu, An, Rabinovich, Obraztsova, and Yeo]{xue2021mis}
Xue, W., Qiu, W., An, B., Rabinovich, Z., Obraztsova, S., and Yeo, C.~K.
\newblock Mis-spoke or mis-lead: Achieving robustness in multi-agent communicative reinforcement learning.
\newblock \emph{arXiv preprint arXiv:2108.03803}, 2021.

\bibitem[Ye et~al.(2024)Ye, He, Gu, and Zhang]{ye2024towards}
Ye, C., He, J., Gu, Q., and Zhang, T.
\newblock Towards robust model-based reinforcement learning against adversarial corruption.
\newblock \emph{arXiv preprint arXiv:2402.08991}, 2024.

\bibitem[Yu et~al.(2021)Yu, Gehring, Sch{\"a}fer, and Anandkumar]{yu2021robust}
Yu, J., Gehring, C., Sch{\"a}fer, F., and Anandkumar, A.
\newblock Robust reinforcement learning: A constrained game-theoretic approach.
\newblock In \emph{Learning for Dynamics and Control}, pp.\  1242--1254. PMLR, 2021.

\bibitem[Yuan et~al.(2023)Yuan, Zhang, Xue, Yin, Chen, Guan, Li, Qian, and Yu]{yuan2023robust}
Yuan, L., Zhang, Z., Xue, K., Yin, H., Chen, F., Guan, C., Li, L., Qian, C., and Yu, Y.
\newblock Robust multi-agent coordination via evolutionary generation of auxiliary adversarial attackers.
\newblock In \emph{Proceedings of the AAAI Conference on Artificial Intelligence}, volume~37, pp.\  11753--11762, 2023.

\bibitem[Yuan et~al.(2024)Yuan, Jiang, Li, Chen, Zhang, and Yu]{yuan2024robust}
Yuan, L., Jiang, T., Li, L., Chen, F., Zhang, Z., and Yu, Y.
\newblock Robust cooperative multi-agent reinforcement learning via multi-view message certification.
\newblock \emph{Science China Information Sciences}, 67\penalty0 (4):\penalty0 142102, 2024.

\bibitem[Yun et~al.(2022)Yun, Park, Kim, Shin, Jung, Mohaisen, and Kim]{yun2022cooperative}
Yun, W.~J., Park, S., Kim, J., Shin, M., Jung, S., Mohaisen, D.~A., and Kim, J.-H.
\newblock Cooperative multiagent deep reinforcement learning for reliable surveillance via autonomous multi-uav control.
\newblock \emph{IEEE Transactions on Industrial Informatics}, 18\penalty0 (10):\penalty0 7086--7096, 2022.

\bibitem[Zhang et~al.(2020{\natexlab{a}})Zhang, Chen, Xiao, Li, Liu, Boning, and Hsieh]{zhang2020robust_1}
Zhang, H., Chen, H., Xiao, C., Li, B., Liu, M., Boning, D., and Hsieh, C.-J.
\newblock Robust deep reinforcement learning against adversarial perturbations on state observations.
\newblock \emph{Advances in Neural Information Processing Systems}, 33:\penalty0 21024--21037, 2020{\natexlab{a}}.

\bibitem[Zhang et~al.(2021{\natexlab{a}})Zhang, Chen, Boning, and Hsieh]{zhang2021robust}
Zhang, H., Chen, H., Boning, D., and Hsieh, C.-J.
\newblock Robust reinforcement learning on state observations with learned optimal adversary.
\newblock \emph{arXiv preprint arXiv:2101.08452}, 2021{\natexlab{a}}.

\bibitem[Zhang et~al.(2020{\natexlab{b}})Zhang, Sun, Tao, Genc, Mallya, and Basar]{zhang2020robust}
Zhang, K., Sun, T., Tao, Y., Genc, S., Mallya, S., and Basar, T.
\newblock Robust multi-agent reinforcement learning with model uncertainty.
\newblock \emph{Advances in neural information processing systems}, 33:\penalty0 10571--10583, 2020{\natexlab{b}}.

\bibitem[Zhang et~al.(2021{\natexlab{b}})Zhang, Yang, and Ba{\c{s}}ar]{zhang2021multi}
Zhang, K., Yang, Z., and Ba{\c{s}}ar, T.
\newblock Multi-agent reinforcement learning: A selective overview of theories and algorithms.
\newblock \emph{Handbook of reinforcement learning and control}, pp.\  321--384, 2021{\natexlab{b}}.

\bibitem[Zhang et~al.(2020{\natexlab{c}})Zhang, Ma, Singla, and Zhu]{zhang2020adaptive}
Zhang, X., Ma, Y., Singla, A., and Zhu, X.
\newblock Adaptive reward-poisoning attacks against reinforcement learning.
\newblock In \emph{International Conference on Machine Learning}, pp.\  11225--11234. PMLR, 2020{\natexlab{c}}.

\bibitem[Zhang et~al.(2023)Zhang, Sun, Huang, and Miao]{zhang2023safe}
Zhang, Z., Sun, Y., Huang, F., and Miao, F.
\newblock Safe and robust multi-agent reinforcement learning for connected autonomous vehicles under state perturbations.
\newblock \emph{arXiv preprint arXiv:2309.11057}, 2023.

\bibitem[Zhou et~al.(2023)Zhou, Liu, and Zhou]{zhou2023robust}
Zhou, Z., Liu, G., and Zhou, M.
\newblock A robust mean-field actor-critic reinforcement learning against adversarial perturbations on agent states.
\newblock \emph{IEEE Transactions on Neural Networks and Learning Systems}, 2023.

\bibitem[Zhou et~al.(2024)Zhou, Liu, Guo, and Zhou]{zhou2024adversarial}
Zhou, Z., Liu, G., Guo, W., and Zhou, M.
\newblock Adversarial attacks on multiagent deep reinforcement learning models in continuous action space.
\newblock \emph{IEEE Transactions on Systems, Man, and Cybernetics: Systems}, 2024.

\end{thebibliography}
\bibliographystyle{icml2025}

\newpage
\appendix
\onecolumn

\counterwithin{table}{section}
\counterwithin{figure}{section} 
\renewcommand{\theequation}{\thesection.\arabic{equation}}

\setcounter{equation}{0}

\section{Environmental Setup}
\label{appsec:env}

We conduct experiments in the MPE \cite{lowe2017multi} and SMAC \cite{samvelyan2019starcraft} environments. This section provides detailed descriptions of their setup and features.

\subsection{Multi-Agent Particle Environments (MPE)}
\begin{figure}[h!]
    \centering
    \begin{subfigure}{0.21\columnwidth}
        \centering
        \includegraphics[width=\linewidth]{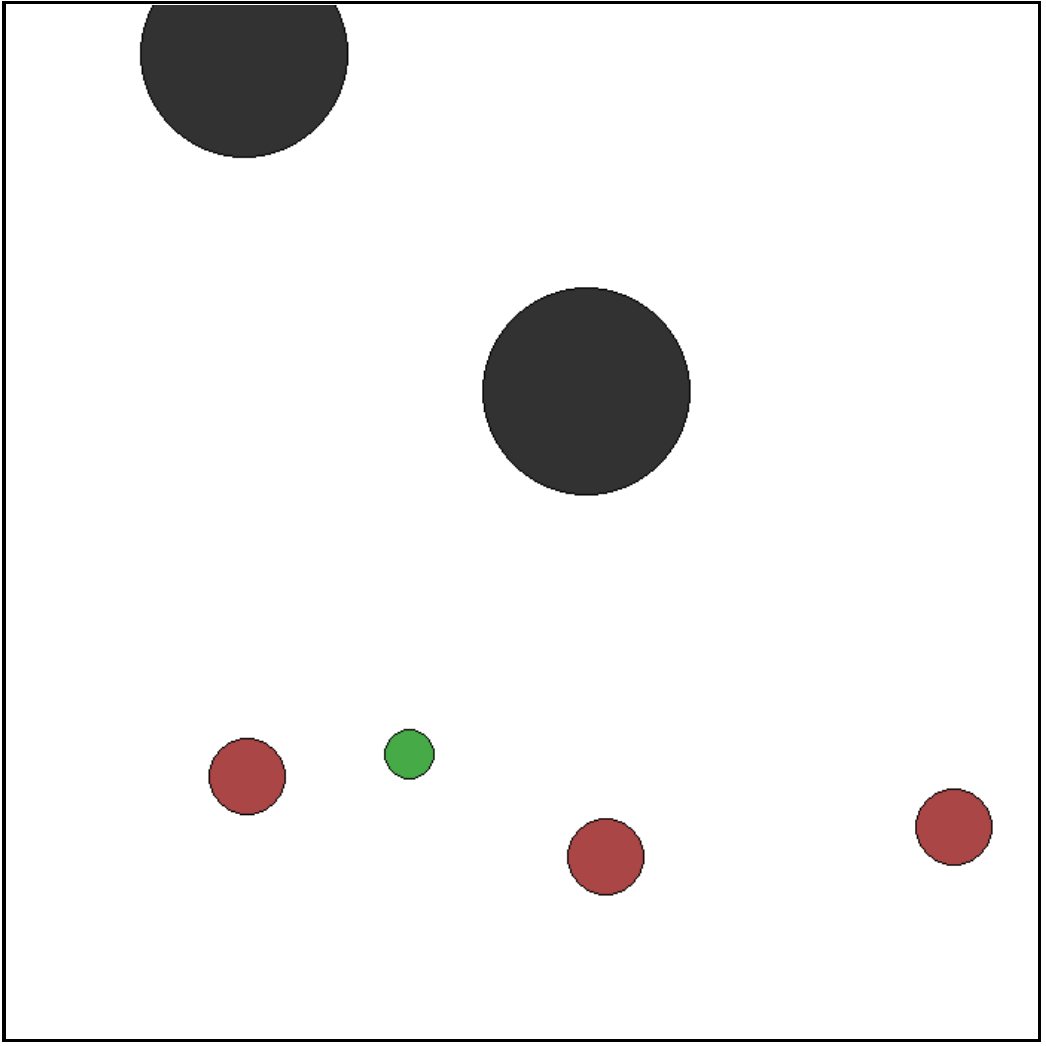}
        \caption{\texttt{PP\_3/1}}  
    \end{subfigure}
    \hspace{0.02\textwidth}
    \begin{subfigure}{0.21\columnwidth}
        \centering
        \includegraphics[width=\linewidth]{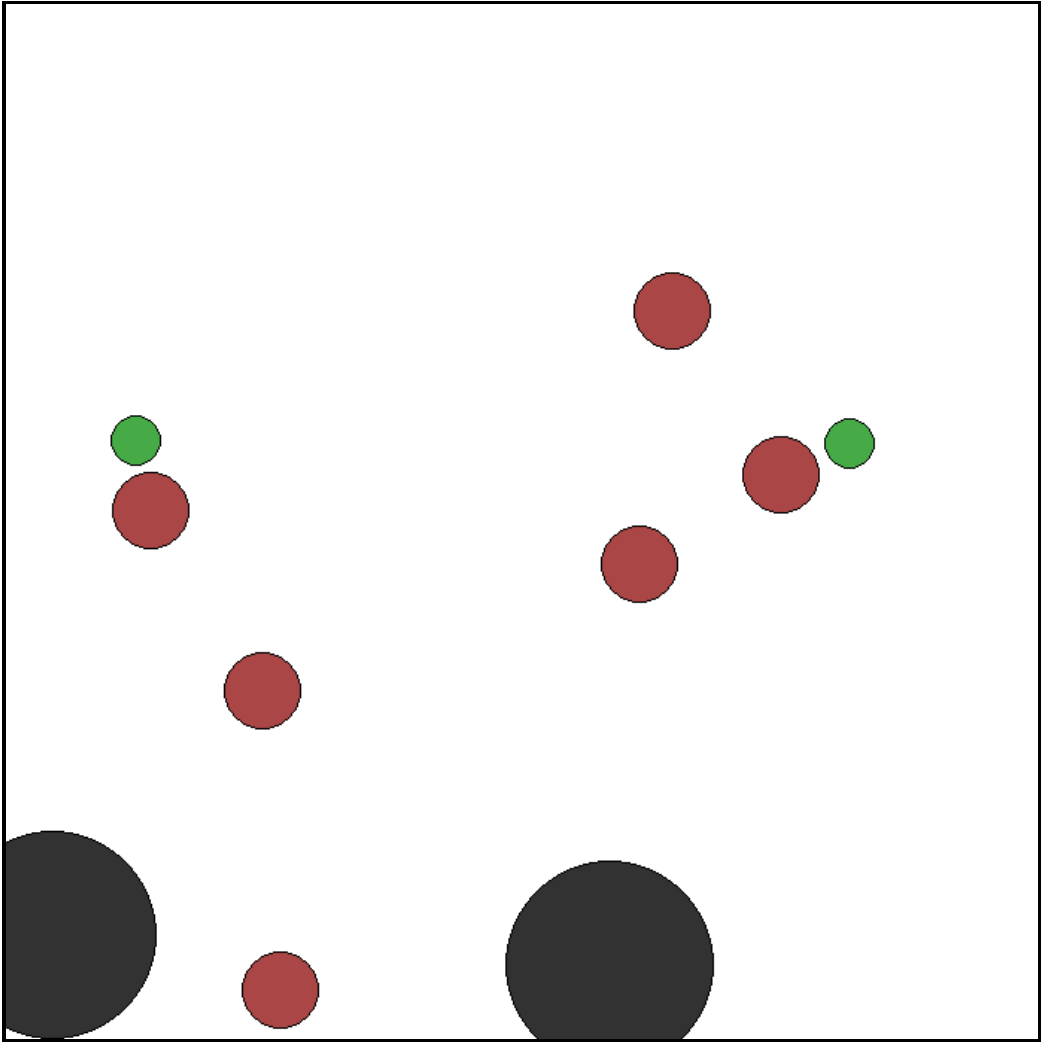}
        \caption{\texttt{PP\_6/2}}  
    \end{subfigure}
    \hspace{0.02\textwidth}
    \begin{subfigure}{0.21\columnwidth}
        \centering
        \includegraphics[width=\linewidth]{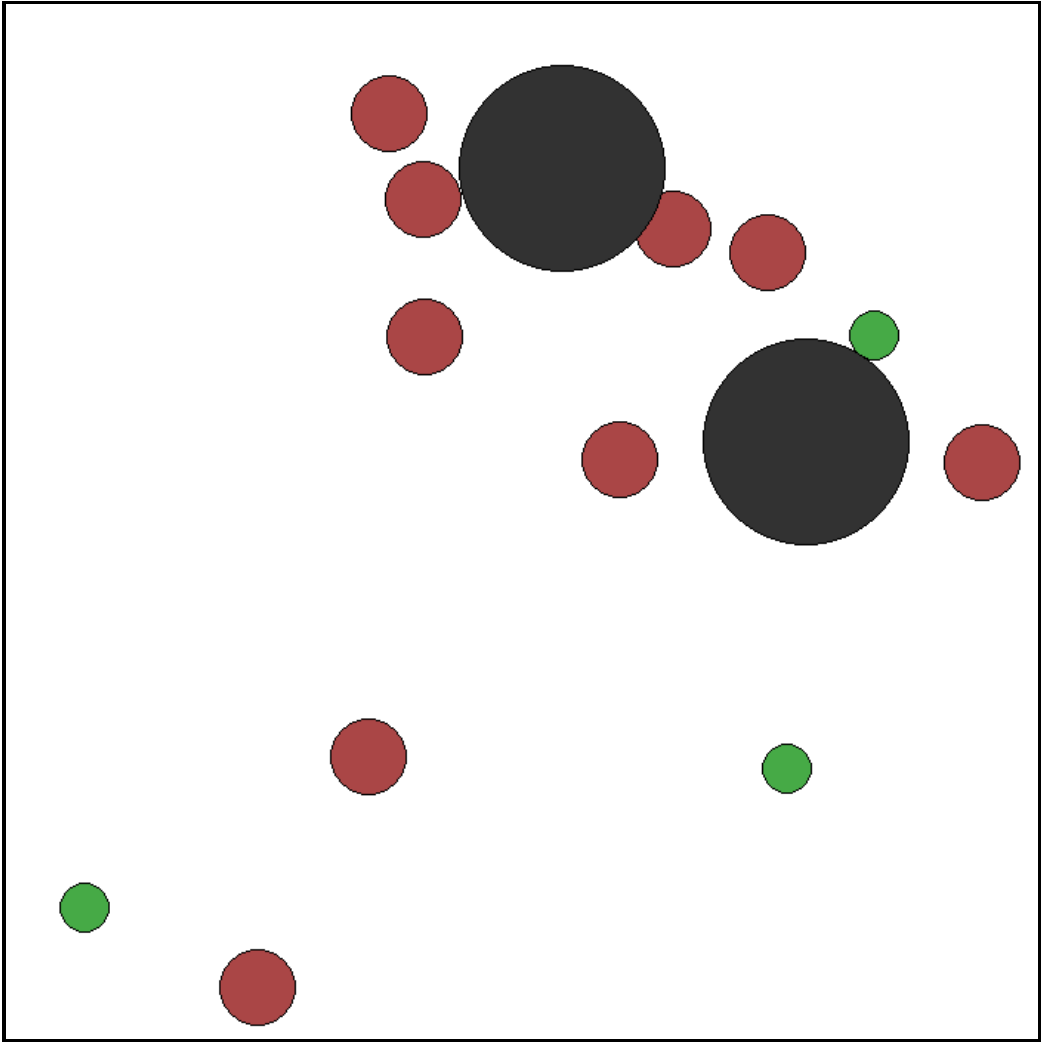}
        \caption{\texttt{PP\_9/3}}  
    \end{subfigure}
    \vspace{-0.1in}
    \caption{Visualization of PP scenarios in the MPE environment}
    \label{fig:mpe_scenarios}
\end{figure}

The Multi-Agent Particle Environment (MPE) \cite{lowe2017multi} is a widely used benchmark suite consisting of multi-agent scenarios. Agents are modeled as particles capable of movement and interaction, governed by simple physical dynamics. MPE includes both cooperative and competitive tasks, with each scenario sharing a continuous state space and typically partial observability. A standardized implementation of MPE is available through the PettingZoo library \cite{terry2021pettingzoo}.

\textbf{Scenarios} \\
The MPE benchmark includes a variety of multi-agent scenarios. Among them, we focus on the predator-prey environment, which is well-suited for analyzing the impact of attacks on the cooperative structures among agents. We consider multiple variants denoted as $\texttt{PP}\_X/Y$, where $X$ represents the number of predator agents and $Y$ the number of prey agents. In all scenarios, prey agents follow a random policy. Detailed configurations for the three selected variants are summarized in \cref{tab:mpe_scenarios} and Fig. \ref{fig:mpe_scenarios}.

\textbf{State and observation spaces} \\
Each agent in the MPE environment receives a partial observation, which includes its own position and velocity, the relative positions and velocities of other predators, and the relative positions of landmarks. The global state is constructed by concatenating the local observations of all agents.

\textbf{Action space} \\
The action space is discrete. Each agent can choose one of the four cardinal directions or do nothing.

\textbf{Reward function} \\
The reward function $R$ assigns a positive value to predator agents upon a successful collision with a prey:
\begin{align*}
R &= \sum_{g \in \text{prey}} \mathbb{I}(\text{collision}(g, \text{predator}))
\end{align*}
where $\mathbb{I}(\cdot)$ is the indicator function, returning 1 if the condition inside is true and 0 otherwise. Here, $\text{collision}(i, j)$ denotes whether agent $i$ and agent $j$ are in physical contact.

\begin{table}[h!]
    \centering
    \begin{tabular}{llllll}
    \toprule
    \textbf{Map} & \textbf{Predators} & \textbf{Prey} & \textbf{State Dimension} & \textbf{Obs Dimension} & \textbf{Num. of Actions} \\
    \midrule
    PP\_3/1  & 3 Agents & 1 Agent & 48 & 16 & 5 \\
    \midrule
    PP\_6/2  & 6 Agents & 2 Agents & 156 & 26 & 5 \\
    \midrule
    PP\_9/3  & 9 Agents & 3 Agents & 324 & 36 & 5 \\
    \bottomrule
    \end{tabular}
    \caption{The number of agents, the dimensions of state and observation spaces, and the number of actions in MPE scenarios}
    \label{tab:mpe_scenarios}
\end{table}

\newpage

\subsection{The StarCraft Multi-Agent Challenge (SMAC)}

\begin{figure}[h!]
    \centering
    \vspace{-0.1in}
    \begin{subfigure}{0.3\columnwidth}
        \centering
        \includegraphics[width=\linewidth]{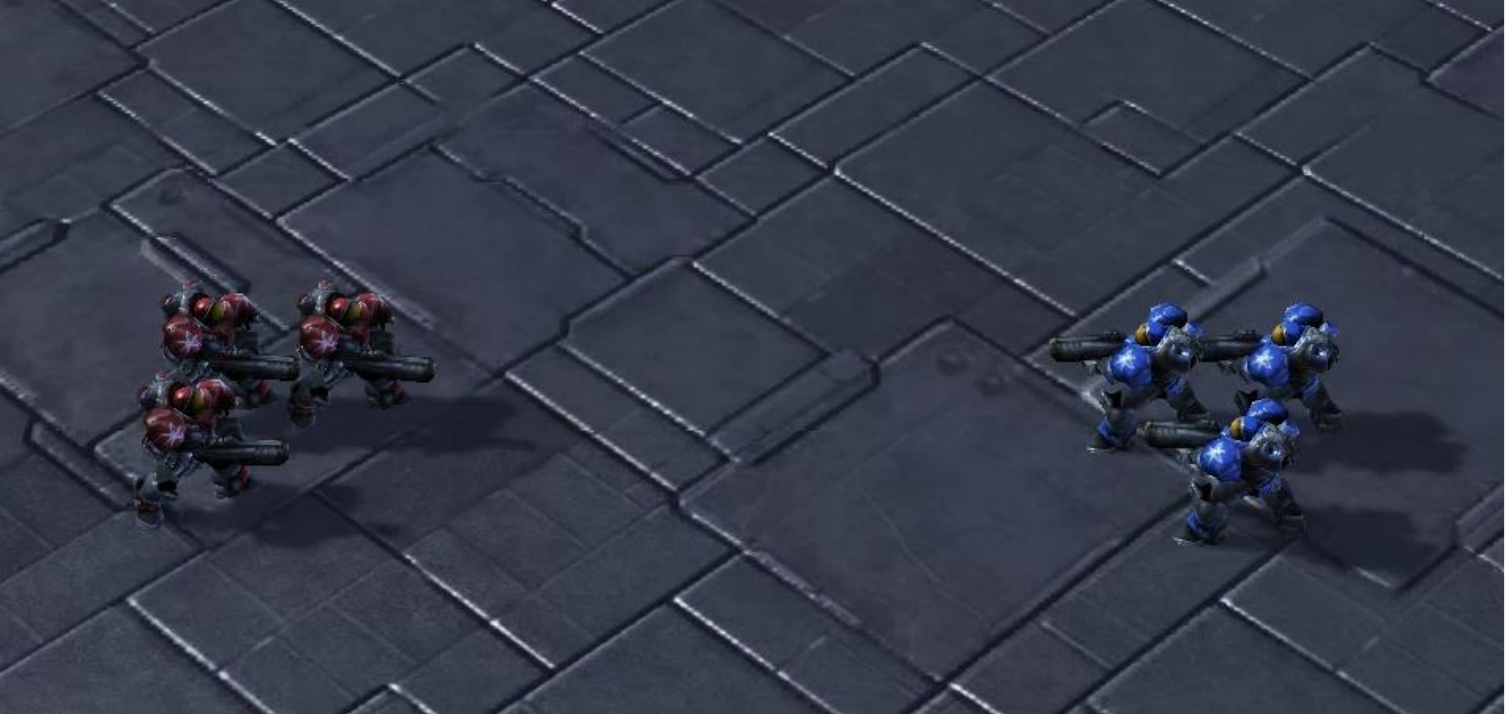}
        \caption{\texttt{3m}}  
    \end{subfigure}
    \begin{subfigure}{0.3\columnwidth}
        \centering
        \includegraphics[width=\linewidth]{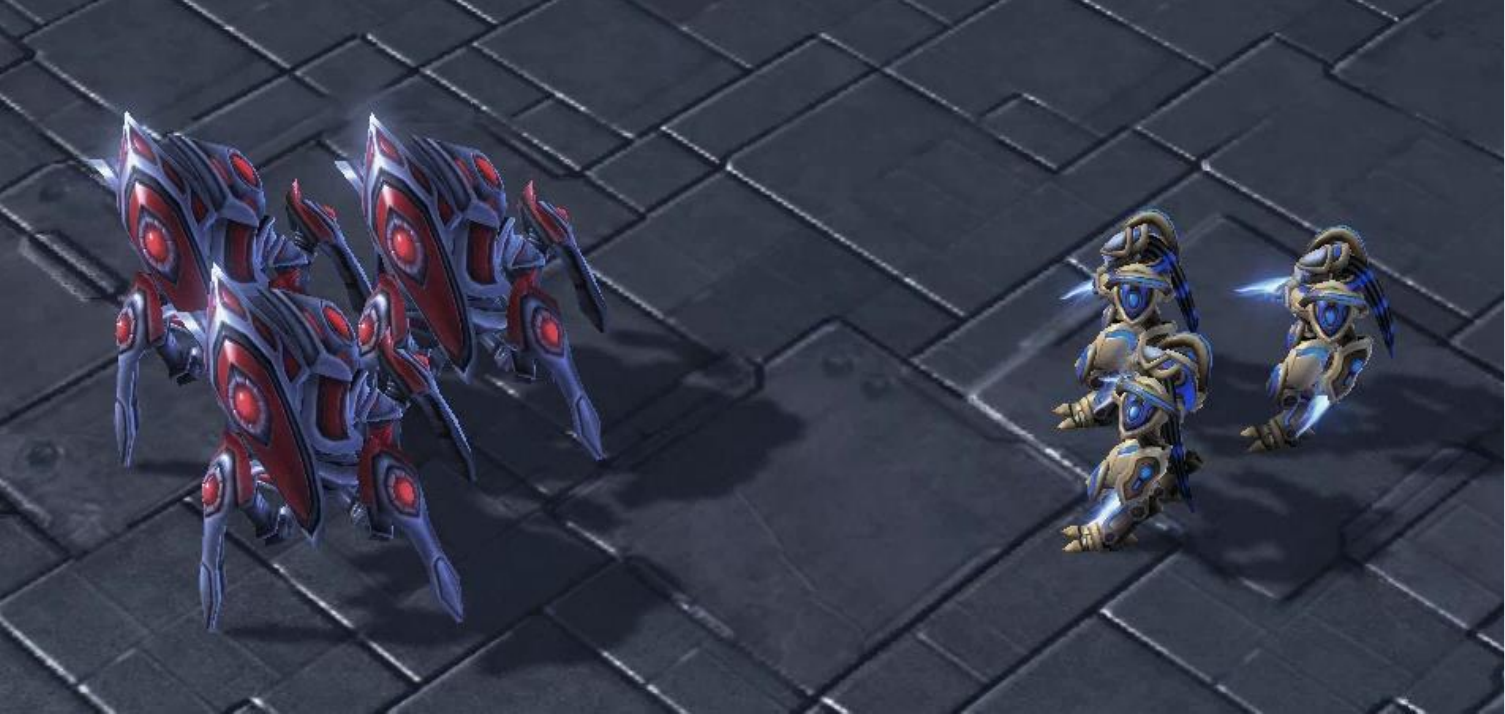}
        \caption{\texttt{3s\_vs\_3z}}  
    \end{subfigure}
    \begin{subfigure}{0.3\columnwidth}
        \centering
        \includegraphics[width=\linewidth]{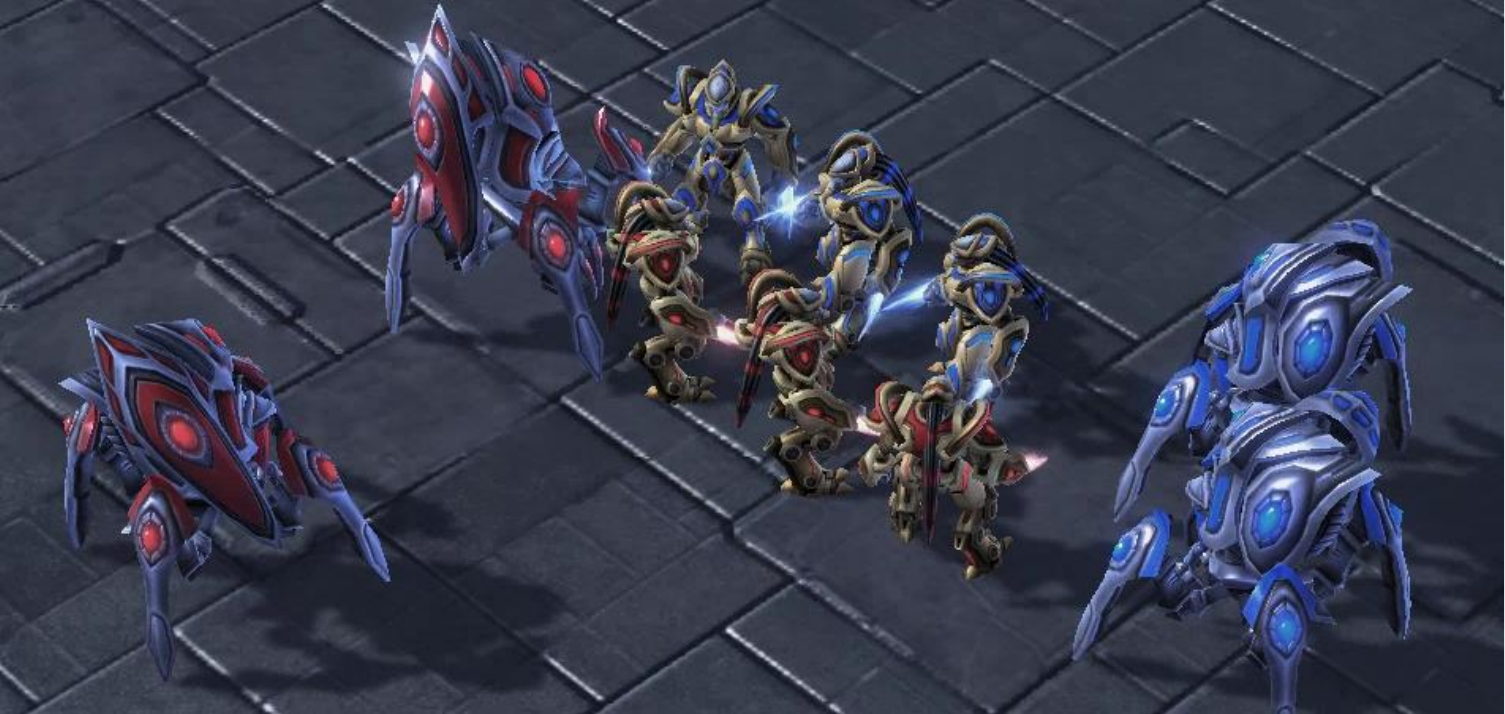}
        \caption{\texttt{2s3z}}  
    \end{subfigure}
    \begin{subfigure}{0.3\columnwidth}
        \centering
        \includegraphics[width=\linewidth]{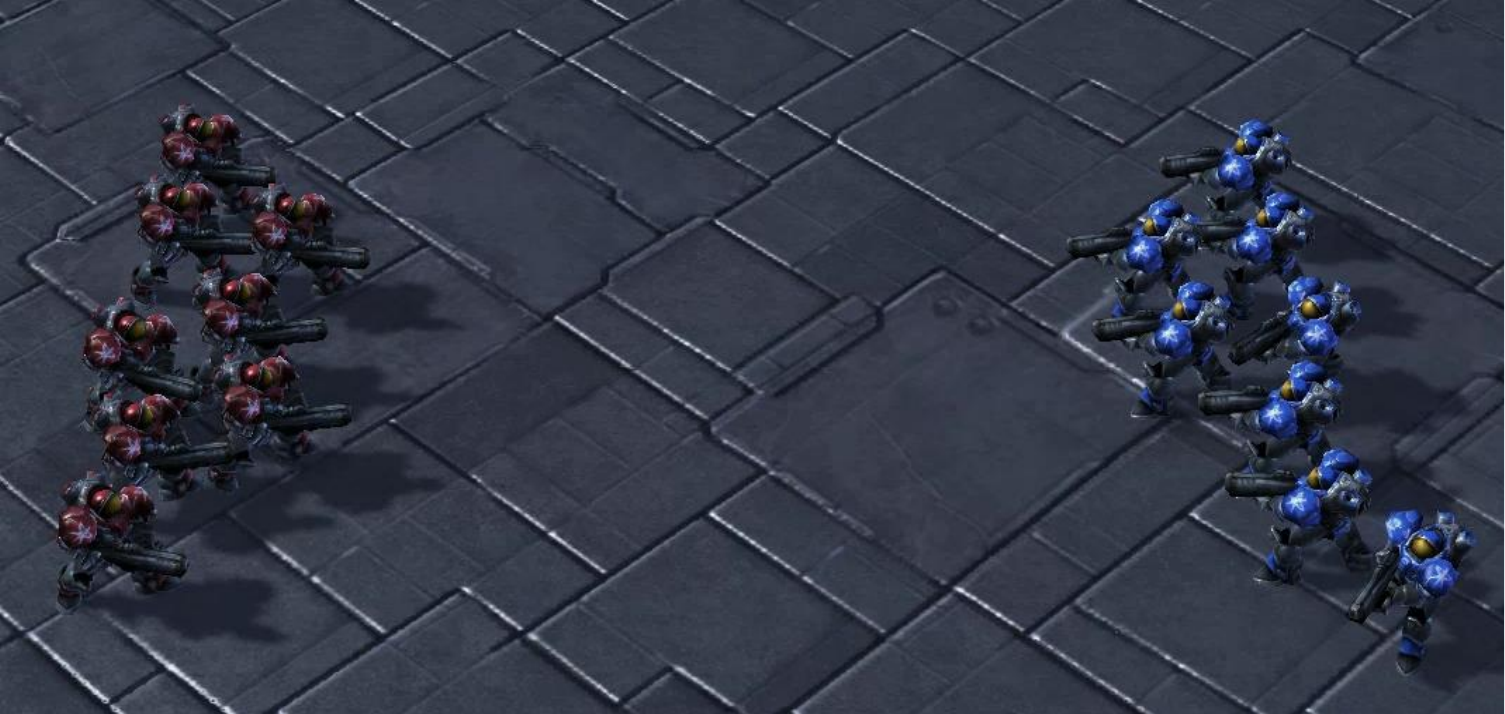}
        \caption{\texttt{8m}}  
    \end{subfigure}
    \begin{subfigure}{0.3\columnwidth}
        \centering
        \includegraphics[width=\linewidth]{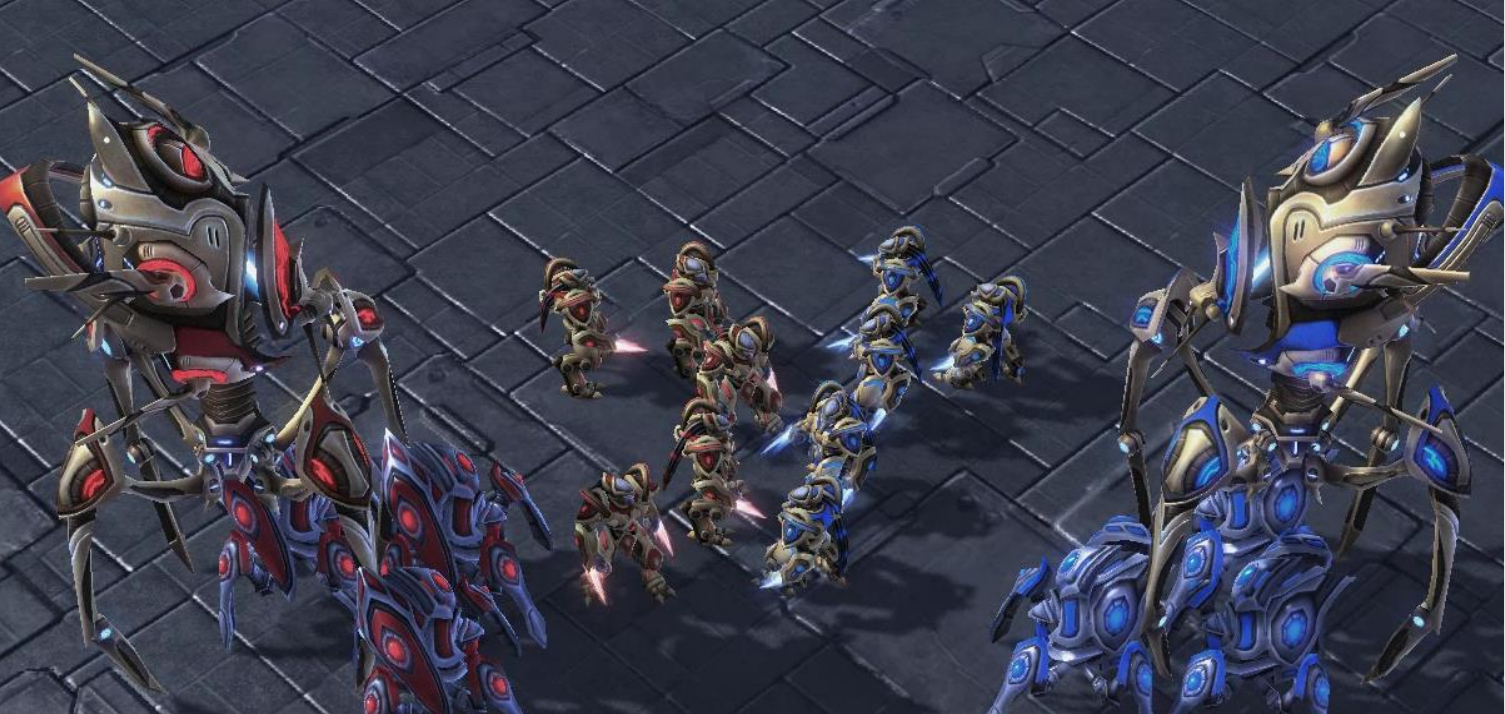}
        \caption{\texttt{1c3s5z}}  
    \end{subfigure}
    \begin{subfigure}{0.3\columnwidth}
        \centering
        \includegraphics[width=\linewidth]{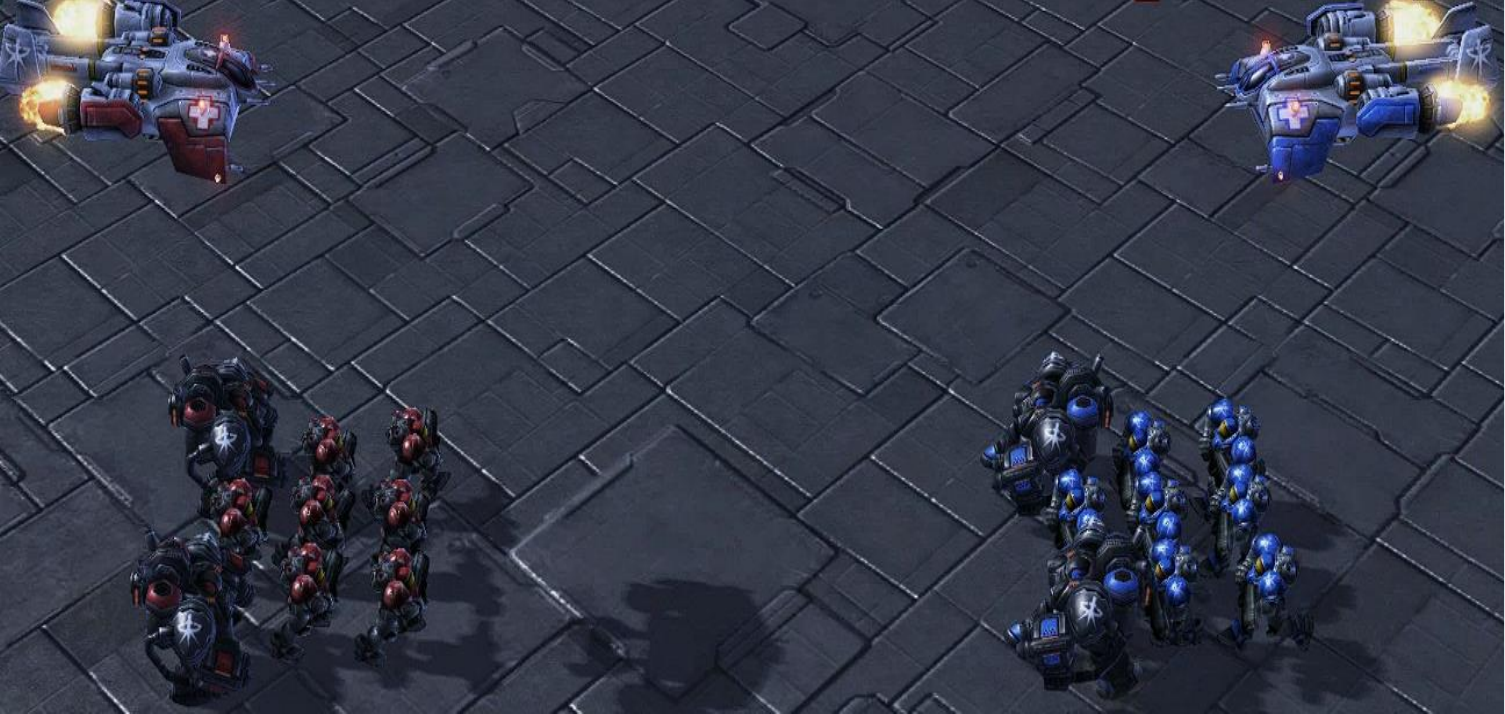}
        \caption{\texttt{MMM}}  
    \end{subfigure}
    \vspace{-0.1in}
    \caption{Visualization of SMAC scenarios}
    \label{fig:scenarios}
\end{figure}

The StarCraft Multi-Agent Challenge (SMAC) serves as a benchmark for cooperative Multi-Agent Reinforcement Learning (MARL), focusing on decentralized micromanagement tasks. Based on the real-time strategy game StarCraft II, SMAC requires each unit to be controlled by independent agents acting solely on local observations. It offers a variety of combat scenarios to evaluate MARL methods effectively.

\textbf{Scenarios} \\
SMAC scenarios involve combat situations between an allied team controlled by learning agents and an enemy team managed by a scripted AI. These scenarios vary in complexity, unit composition, and terrain, challenging agents to use advanced micromanagement techniques such as focus fire, kiting, and terrain exploitation. Scenarios end when all units on one side are eliminated or when the time limit is reached. The objective is to maximize the win rate of the allied agents. Detailed descriptions of the scenarios and unit compositions are provided in \cref{tab:scenarios} and Fig. \ref{fig:scenarios}.

\textbf{State and observation spaces} \\
In the SMAC environment, each agent receives partial observations that contain information about visible allies and enemies within a fixed sight range of 9. These observations do not include any global state and are specifically designed to support decentralized decision-making based only on each agent’s local view of the environment.

The global state, which is used during centralized training, is constructed by aggregating the features of all agents and enemies. It consists of three primary components. The ally state includes each agent’s relative $x$ and $y$ positions, health, energy, shield (if applicable), and unit type. The enemy state is similar but excludes energy, containing relative positions, health, shield, and unit type of enemies. The last action component records the most recent action taken by each agent, represented as a one-hot encoded vector. The full global state is formed by concatenating these components, and its dimensionality depends on the number of agents, enemies, and available actions.

Each agent's observation vector is separately constructed from the following elements. The movement features indicate the four cardinal directions the agent can move in, resulting in a fixed size of 4. The enemy features describe each observed enemy, including available action flag, distance to the agent, relative $x$ and $y$ positions, health, shield (if applicable), and unit type. The ally features encode the same types of information for all visible allies, excluding the observing agent. Finally, the own features contain the observing agent’s own health, shield, and unit type.

The precise dimensions of the observation and state vectors vary depending on the specific map scenario and unit composition, and are summarized in Table~\ref{tab:scenarios}.

\textbf{Action space} \\
Agents can perform discrete actions, including movement in four cardinal directions (North, South, East, West), attacking specific enemy units within a shooting range of 6 units, and specialized actions such as healing for units like Medivacs. Additionally, agents can perform a stop or a no-op action, the latter being restricted to dead units. 

The size of the action space varies depending on the scenario and is defined as $n_{\text{actions}} = 6 + n_{\text{enemies}}$, where $6$ represents movement, stop, or a no-op action. The inclusion of $n_{\text{enemies}}$ accounts for the need to specify which enemy unit is targeted when performing an attack action. The exact size of the action space varies across different maps, as summarized in Table \ref{tab:scenarios}.

\textbf{Reward function} \\
SMAC uses a shaped reward function $R$ to guide learning, including components for damage dealt ($R_{\text{damage}}$), enemy units killed ($R_{\text{enemy\_killed}}$), and scenario victory ($R_{\text{win}}$). The total reward is defined as:
\begin{align*}
R &= \sum_{e \in \text{enemies}} \Delta \text{Health}(e) + \sum_{e \in \text{enemies}} \mathbb{I}(\{\text{Health}(e) = 0\}) \cdot \text{Reward}_{\text{death}} + \mathbb{I}(\{\text{win} = \text{True}\}) \cdot \text{Reward}_{\text{win}} 
\end{align*}
Here, $\text{Health}(e)$ represents the health of an enemy unit $e$, and $\Delta \text{Health}(e)$ is the reduction in its health during a timestep. The indicator function $\mathbb{I}(\cdot)$ returns $1$ if the condition inside is true and $0$ otherwise. The parameters $\text{Reward}_{\text{death}}$ and $\text{Reward}_{\text{win}}$ are scaling factors for rewards when an enemy unit is killed and when the agents win the scenario, set to $10$ and $200$, respectively.

\begin{table}[ht]
    \centering
    \begin{tabular}{llllll}
    \toprule
    \textbf{Map} & \textbf{Ally Units} & \textbf{Enemy Units} & \textbf{State Dimension} & \textbf{Obs Dimension} & \textbf{Num. of Actions} \\
    \midrule
    3m           & 3 Marines                         & 3 Marines   & 48   & 30  & 9  \\
    \midrule
    3s\_vs\_3z    & 3 Stalkers                          & 3 Zealots   & 54   & 36  & 9  \\
    \midrule
    2s3z          & 2 Stalkers,                        & 2 Stalkers,  & 120  & 80  & 11 \\
                          & 3 Zealots                        & 3 Zealots  &       &     &    \\
    \midrule
    8m           & 8 Marines                         & 8 Marines & 168  & 80  & 14 \\
    \midrule
    1c3s5z        & 1 Colossus,                       & 1 Colossus,  & 270  & 162  & 15 \\
                          & 3 Stalkers,                       & 3 Stalkers,  &       &     &    \\
                          & 5 Zealots                        & 5 Zealots   &       &     &    \\
    \midrule
    MMM          & 1 Medivac,                       & 1 Medivac,   & 290  & 160 & 16 \\
                          & 2 Marauders,                      & 2 Marauders, &       &     &    \\
                          & 7 Marines                        & 7 Marines   &       &     &    \\
    \bottomrule
    \end{tabular}
    \caption{The number of agents, the dimensions of state and observation spaces, and the number of actions in SMAC scenarios}
    \label{tab:scenarios}
\end{table}

\newpage
\section{Implementation Details}

In this section, we provide a detailed implementation of the proposed WALL framework. Section \ref{appsubsec:trans} outlines the implementation of the Transformer used to efficiently identify critical initial steps. Section \ref{appsubsec:marl} elaborates on the components constituting the Wolfpack attack and provides an in-depth explanation of the reinforcement learning implementation.

\subsection{Practical Implementation of Planner Transformer}
\begin{figure}[!h]
    \centering
    \begin{subfigure}{0.6\columnwidth} 
        \centering
        \includegraphics[width=1.0\textwidth]{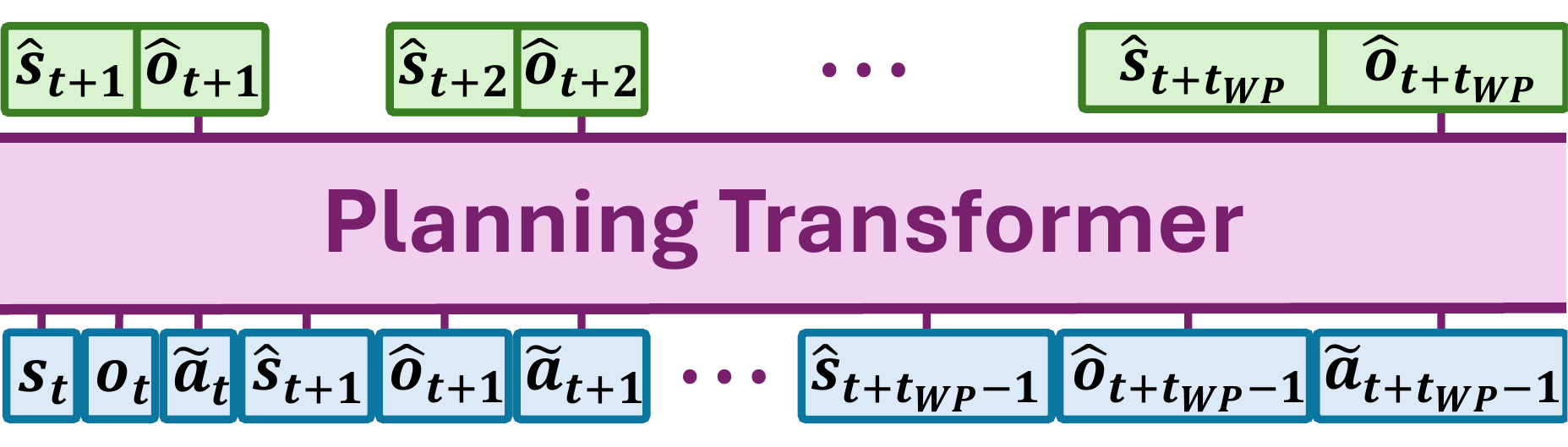} 
        \vspace{-0.25in}
        \caption{Planning Transformer}  
    \end{subfigure}
    \begin{subfigure}{0.6\columnwidth} 
        \centering
        \includegraphics[width=1.0\textwidth]{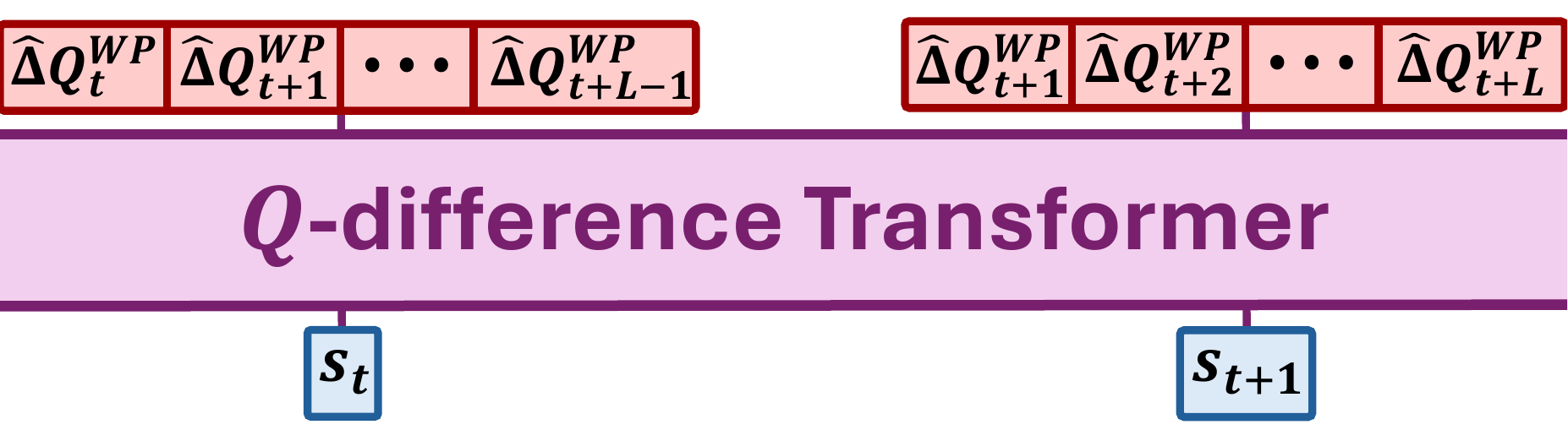} 
        \caption{$Q$-difference Transformer}  
    \end{subfigure}
    \vspace{-0.1in}
    \caption{Structure of Transformers}
    \label{fig:transformer_str}
\end{figure}
\label{appsubsec:trans}

Critical attacking step selection introduced in Section \ref{subsec:attackstep} requires planning to compute the reduction in $Q$-function values, $\Delta Q_l^{\mathrm{WP}}$ ($l=t, \cdots, t+L-1$), over multiple time steps. To facilitate this process, a Transformer model is employed. During training, the Transformer predicts $(\hat{s}_{t+1}, \hat{\mathbf{o}}_{t+1}, \cdots, \hat{s}_{t+t_{\mathrm{WP}}}, \hat{\mathbf{o}}_{t+t_{\mathrm{WP}}})$ at the current step $t$ to compute the target $\hat{\Delta}Q_t^{\mathrm{WP}}$. However, performing this planning process at each evaluation step to calculate the probability of an initial attack, $P_{t,\mathrm{attack}}$, is computationally expensive.

To address this, the single Transformer is split into two components: a planning Transformer and a $Q$-difference Transformer. The planning Transformer predicts $(\hat{s}_{t+1}, \hat{\mathbf{o}}_{t+1})$ and is used only during training, while the $Q$-difference Transformer predicts $\Delta Q_l^{\mathrm{WP}}$ ($l=t, \cdots, t+L-1$) and is employed exclusively during evaluation. This separation enables efficient computation of $P_{t,\mathrm{attack}}$ during evaluation, significantly reducing computational costs. Both Transformers adopt the Decision Transformer structure \cite{chen2021decision}, consisting of Transformer decoder layers. The planning Transformer is parameterized by $\phi_{\mathrm{planning}}$, and the $Q$-difference Transformer by $\phi_{\mathrm{qdiff}}$. Their respective loss functions are defined as:
\begin{align}
    \mathcal{L}_{\mathrm{planning}}(\phi_{\mathrm{planning}}) &= \mathbb{E} \left[ \| s_{t+1} - \hat{s}_{t+1}(\phi_{\mathrm{planning}}) \|^2 + \| \mathbf{o}_{t+1} - \hat{\mathbf{o}}_{t+1}(\phi_{\mathrm{planning}}) \|^2 \right], \\
    \mathcal{L}_{\mathrm{qdiff}}(\phi_{\mathrm{qdiff}}) &= \mathbb{E} \left[ \| \Delta Q_t^{\mathrm{WP}} - \hat{\Delta} Q_t^{\mathrm{WP}}(\phi_{\mathrm{qdiff}}) \|^2 \right],~\forall t.
\end{align}
As shown in Fig. \ref{fig:transformer_str}, the estimated state $\hat{s}_{t+1}$ and observations $\hat{\mathbf{o}}_{t+1}$ are generated by the planning Transformer, which takes previous trajectories as input. Similarly, the estimated $Q$-difference $\hat{\Delta} Q_t^{\mathrm{WP}}$ is produced by the $Q$-diff Transformer, using previous states as input. The loss function $\mathcal{L}_{\mathrm{planning}}$ minimizes the prediction error for the next state $s_{t+1}$ and observation $\mathbf{o}_{t+1}$, while $\mathcal{L}_{\mathrm{qdiff}}$ minimizes the prediction error for $\Delta Q_l^{\mathrm{WP}}$ resulting from the Wolfpack attack. Fig. \ref{fig:transformer_str} illustrates the architectures of the Transformers, where (a) represents the planning Transformer and (b) represents the $Q$-diff Transformer.

\subsection{Detailed Implementation of WALL}
\label{appsubsec:marl}

The WALL framework trains robust MARL policies to counter the Wolfpack adversarial attack by employing a $Q$-learning approach within the CTDE paradigm. Each agent computes its individual $Q$-values, $Q^i(\tau_t^i,a_t^i),~i=1,\cdots,n$, using separate $Q$-networks. These individual values, along with the global state, are combined through a mixing network to produce the total $Q$-value, $Q_\theta^{tot}$, parameterized by $\theta$. This joint value function ensures effective coordination among agents under adversarial scenarios.

To achieve robustness, the training process minimizes the temporal difference (TD) loss, which incorporates the observed rewards, state transitions, and target $Q$-values parameterized by a separate target network, $\theta^-$. By leveraging this target network, the CTDE frameworks stabilize learning and mitigate the impact of the Wolfpack attack. The TD loss is defined as:
\begin{equation}
\mathcal{L}_{\text{TD}}(\theta) = \mathbb{E}_{s, \mathbf{a}, r, s'} \left[ \left( r_t + \gamma \max_{\mathbf{a}'} Q_{\theta^-}^{tot}(s_{t+1}, \mathbf{a}') - Q_\theta^{tot}(s_t, \mathbf{a}_t) \right)^2 \right],
\label{eq:td_loss}
\end{equation}
where the target network parameter $\theta^-$ is updated by applying the exponential moving average (EMA) to $\theta$. This training mechanism allows agents to adapt and develop robust policies capable of resisting the coordinated disruptions caused by the Wolfpack adversarial attack, ensuring enhanced performance and resilience in MARL scenarios. 
In addition, we use 3 value-based CTDE algorithms as baselines for the WALL framework: QMIX, VDN, and QPLEX. Below, we provide an outline of the key details of these baseline algorithms:

\textbf{Value-Decomposition Networks (VDN)} \\
VDN \cite{sunehag2017value} is a $Q$-learning algorithm designed for cooperative MARL. It introduces an additive decomposition of the joint $Q$-value into individual agent $Q$-values, enabling centralized training and decentralized execution. The joint action-value function, $Q^{tot}$, is expressed as:
\begin{equation}
    Q^{tot}(s_t, \mathbf{a}_t) = \sum_{i=1}^{n} Q^i(\tau_t^i, a_t^i),
\end{equation}
allowing agents to act independently during execution by relying only on their local $Q^i$ values. \\

\textbf{QMIX} \\
QMIX \cite{rashid2020monotonic} extends VDN by introducing a more expressive, non-linear representation of the joint $Q$-value, while maintaining a monotonic relationship between $Q^{tot}$ and individual agent $Q$-values, $Q^i(\tau^i, a^i)$. This ensures individual-global-max (IGM) condition:
\begin{equation}
    \frac{\partial Q^{tot}}{\partial Q^i} \geq 0, \forall i,
\end{equation}
guaranteeing consistency between centralized and decentralized policies. Specifically:
\begin{equation}
\arg\max_{\mathbf{a}} Q^{\text{tot}}(s_t, \mathbf{a}_t) = 
\begin{pmatrix}
\arg\max_{a^1} Q^{1}(\tau_t^1, a_t^1), \\
\vdots \\
\arg\max_{a^n} Q^{n}(\tau_t^n, a_t^n)
\end{pmatrix}.  
\end{equation}
QMIX combines agent networks, a mixing network, and hypernetworks, where hypernetworks dynamically parameterize the mixing network based on the global state $s_t$. The weights generated by the hypernetworks are constrained to be non-negative to enforce the monotonicity constraint. \\

\textbf{QPLEX} \\
QPLEX \cite{wang2020qplex} introduces a duplex dueling architecture to enhance the representation of joint action-value functions while adhering to the IGM principle. QPLEX reformulates the IGM principle in an advantage-based form:
\begin{equation}
    \arg\max_{\mathbf{a}} A^{tot}(\tau, \mathbf{a}) = \begin{pmatrix} \arg\max_{a^1} A^1(\tau^1, a^1), \\
\vdots \\ \arg\max_{a^n} A^n(\tau^n, a^n) \end{pmatrix},
\end{equation}
where $A^{tot}$ and $A^i$ are the advantage functions for joint and individual action-value functions, respectively. The joint action-value function is expressed as:
\begin{equation}
    Q^{tot}(\tau, \mathbf{a}) = \sum_{i=1}^n Q^i(\tau, a^i) + \sum_{i=1}^n (\lambda_i(\tau, \mathbf{a}) - 1) A^i(\tau, a^i).
\end{equation}
where $\lambda_i(\tau, \mathbf{a}) > 0$ are importance weights generated using a multi-head attention mechanism to enhance expressiveness.

Here, VDN and QMIX are implemented using the PyMARL codebase \href{https://github.com/oxwhirl/pymarl}{https://github.com/oxwhirl/pymarl}, while QPLEX is implemented using its official codebase \href{https://github.com/wjh720/QPLEX}{https://github.com/wjh720/QPLEX}.

\newpage

\section{Experimental Details}
\label{appsec:expdetail}

All experiments in this paper are conducted on a GPU server equipped with an NVIDIA GeForce RTX 3090 GPU and AMD EPYC 7513 32-Core processors running Ubuntu 20.04 and PyTorch. We follow the implementations and loss scales provided by the CTDE algorithms and focus on parameter searches for hyperparameters related to the proposed Wolfpack adversarial attack. Comparisons are performed using the optimal hyperparameter setup, with an ablation study available in Appendix \ref{appsec:addabl}.

\subsection{Hyperparameter Setup} \label{appsubsec:hyper}

We conduct parameter search for the number of Wolfpack attacks $K_{\mathrm{WP}} \in [1,2,3,4]$, the attack duration $t_{\mathrm{WP}} \in [1,2,3,4]$, the number of follow-up agents $m$, and the temperature $T \in [0.1,0.2,0.5,1.0]$. The total number of attacks $K$ is then determined based on $K=K_{\mathrm{WP}}\times (t_{\mathrm{WP}}+1)$, separated into training and testing setups. During training, $K$ is selected through hyperparameter sweeping to ensure optimal performance. For testing, $K$ is unified across all adversarial attack setups, including Random Attack, EGA, and the Wolfpack Adversarial Attack, to ensure fair comparisons. Additionally, the attack period $L$ is chosen based on the average episode length of SMAC scenarios and the total number of attacks, with $L=20$ is fixed as appropriate. Transformer hyperparameters, shared between the Planning Transformer and $Q$-difference Transformer, such as the number of heads, decoder layers, embedding dimensions, and input sequence length, are selected to balance accuracy and computational efficiency.

The $Q$-learning hyperparameters (shared across all CTDE methods) and those specific to the CTDE algorithms are detailed in Table \ref{tab:common_hyperparameters} and Table \ref{tab:vdn_qmix_qplex_hyperparameters}, respectively. The Wolfpack adversarial attack-related hyperparameters for the WALL framework, shared across all SMAC scenarios and scenario-specific setups, are presented in Table \ref{tab:wolfpack_hyperparameters}.

\begin{table}[h]
\centering
\begin{tabular}{p{0.3\textwidth} p{0.2\textwidth}}
\hline
\textbf{Hyperparameter} & \textbf{Value} \\
\hline
Epsilon & 1.0 $\rightarrow$ 0.05 \\
Epsilon Anneal Time & 50000 timesteps \\
Train Interval & 1 episode \\
Gamma & 0.99 \\
Critic Loss & MSE Loss \\
Buffer Size & 5000 episodes \\
Batch Size & 32 episodes \\
Agent Learning Rate & 0.0005 \\
Critic Learning Rate & 0.0005 \\
Optimizer & RMSProp \\
Optimizer Alpha & 0.99 \\
Optimizer Eps & 1e-5 \\
Gradient Clip Norm & 10.0 \\
Num GRU Layers & 1 \\
RNN Hidden State Dim & 64 \\
Double Q & True \\
\hline
\end{tabular}
\caption{Common $Q$-learning hyperparameters}
\label{tab:common_hyperparameters}
\end{table}

\newpage
\begin{table}[h]
\centering
\begin{tabular}{p{0.3\textwidth} p{0.1\textwidth} p{0.1\textwidth} p{0.1\textwidth}}
\hline
\textbf{Hyperparameter}        & \textbf{VDN} & \textbf{QMIX} & \textbf{QPLEX} \\
\hline
Mixer                          & VDN          & QMIX          & QPLEX          \\
Mixing Embed Dim.               & -            & 32            & 32             \\
Hypernet Layers                & -            & 2             & 2              \\
Hypernet Embed Dim.             & -            & 64            & 64             \\
Adv Hypernet Layers            & -            & -             & 1              \\
Adv Hypernet Embed Dim.         & -            & -             & 64             \\
Num. Kernel                     & -            & -             & 2              \\
\hline
\end{tabular}
\caption{VDN, QMIX, QPLEX hyperparameters}
\label{tab:vdn_qmix_qplex_hyperparameters}
\end{table}

\begin{table}[h!]
\centering
\begin{tabular}{p{0.442\textwidth} p{0.442\textwidth}}
\hline
\textbf{Common Hyperparameters} & \textbf{Value} \\
\hline
Attack duration ($t_{\mathrm{WP}}$) & 3 \\
Temperature ($T$) & 0.5 \\
Attack Period ($L$) & 20 \\
Num. Transformer Head & 1 \\
Num. Transformer Decoder Layer & 1 \\
Transformer Embed Dim. & 64 \\
Input Sequence Length & 20 \\
\hline
\end{tabular} \\
\vspace{0.3cm}
\begin{tabular}{p{0.28\textwidth}  p{0.185\textwidth} p{0.185\textwidth} p{0.185\textwidth}}
\hline
\textbf{Scenario} & PP\_3/1 & PP\_6/2 & PP\_9/3  \\
\hline
Num. Total Attacks (Train) ($K$) & 4 & 4 & 4  \\
Num. Total Attacks (Test) ($K$) & 4 & 4 & 4 \\
Num. Wolfpack Attacks ($K_{\mathrm{WP}}$) & 1 & 1 & 1  \\
Num. Follow-up Agents ($m$) & 1 & 3 & 5  \\
\hline
\end{tabular} \\
\vspace{0.3cm}
\begin{tabular}{p{0.28\textwidth} p{0.08\textwidth} p{0.08\textwidth} p{0.08\textwidth} p{0.08\textwidth} p{0.08\textwidth} p{0.08\textwidth}}
\hline
\textbf{Scenario} & 3m & 3s\_vs\_3z & 2s3z & 8m & 1c3s5z & MMM \\
\hline
Num. Total Attacks (Train) ($K$) & 8 & 16 & 12 & 8 & 16 & 16 \\
Num. Total Attacks (Test) ($K$) & 8 & 4 & 8 & 4 & 8 & 8 \\
Num. Wolfpack Attacks ($K_{\mathrm{WP}}$) & 2 & 4 & 3 & 2 & 4 & 4 \\
Num. Follow-up Agents ($m$) & 1 & 1 & 2 & 3 & 4 & 4 \\
\hline
\end{tabular}
\caption{Wolfpack hyperparameters shared across scenarios and scenario-specific values}
\label{tab:wolfpack_hyperparameters}
\end{table} 

\onecolumn
\newpage

\section{Details of Other Robust MARL Methods}
\label{appsec:marlbase}
In this section, we detail various robust MARL methods compared against the proposed WALL framework, as below:

\textbf{Robust Adversarial Reinforcement Learning (RARL)} \\
RARL \cite{pinto2017robust} enhances policy robustness by training a protagonist agent and an adversary in a two-player zero-sum Markov game. At each timestep $t$, the agents observe state $s_t$ and take actions $a_t^1 \sim \mu(s_t)$ and $a_t^2 \sim \nu(s_t)$, where $\mu$ is the protagonist’s policy, and $\nu$ is the adversary’s policy. The state transitions follow:
\begin{equation}
    s_{t+1} = P(s_t, a_t^1, a_t^2).
\end{equation}
The protagonist maximizes its cumulative reward $R^1$, while the adversary minimizes it:
\begin{equation}
    R^1_* = \min_{\nu} \max_{\mu} R^1(\mu, \nu) = \max_{\mu} \min_{\nu} R^1(\mu, \nu).
\end{equation}

\textbf{Robustness via Adversary Populations (RAP)} \\
RAP \cite{vinitsky2020robust} improves robustness by training agents against a population of adversaries, reducing overfitting to specific attack patterns. During training, an adversary is sampled uniformly from the population ${\pi_{\phi_1}, \pi_{\phi_2}, \dots, \pi_{\phi_n}}$. The objective is:
\begin{equation}
    \max_{\theta} \min_{\phi_1, \dots, \phi_n} \mathbb{E}_{i \sim U(1, n)} \left[ \sum_{t=0}^T \gamma^t r(s_t, a_t, \alpha a_t^i) \middle| \pi_\theta, \pi_{\phi_i} \right],
\end{equation}
where $\pi_\theta$ is the agent’s policy, $\pi_{\phi_i}$ is the $i$-th adversary, and $\alpha$ controls adversary strength.

\textbf{Robust Multi-Agent Coordination via Evolutionary Generation of Auxiliary Adversarial Attackers (ROMANCE)} \\
ROMANCE \cite{yuan2023robust} generates diverse auxiliary adversarial attackers to improve robustness in CMARL. Its objective combines attack quality and diversity:
\begin{equation}
    L_{\text{adv}}(\phi) = \frac{1}{n_p} \sum_{j=1}^{n_p} L_{\text{opt}}(\phi_j) - \alpha L_{\text{div}}(\phi),
\end{equation}
where $L_{\text{opt}}$ minimizes the ego-system's return, $L_{\text{div}}$ promotes diversity using Jensen-Shannon Divergence, and $n_p$ is the number of adversarial policies. ROMANCE uses an evolutionary mechanism to explore diverse attacks.

We implement RARL and RAP for multi-agent systems, as well as ROMANCE with EGA, using the ROMANCE codebase available at \href{https://github.com/zzq-bot/ROMANCE}{https://github.com/zzq-bot/ROMANCE}.

\textbf{ERNIE} \\
ERNIE \cite{bukharin2024robust} improves robustness by promoting Lipschitz continuity through adversarial regularization. It minimizes discrepancies between policy outputs under perturbed and non-perturbed observations:
\begin{equation}
    R_\pi(o_k; \theta_k) = \max_{\|\delta\| \leq \epsilon} D(\pi_{\theta_k}(o_k + \delta), \pi_{\theta_k}(o_k)),
\end{equation}
where $o_k$ is the agent’s observation, $\delta$ is a bounded perturbation, and $D$ measures divergence (e.g., KL-divergence). ERNIE reformulates adversarial training as a Stackelberg game and extends its framework to mean-field MARL for scalability in large-agent settings. We evaluate ERNIE using its official codebase at \href{https://github.com/abukharin3/ERNIE}{https://github.com/abukharin3/ERNIE}.

\newpage 

\section{Additional Experiments Results}
This section presents additional experimental results. ~\ref{appsubsec:compotherctde} reports comparison results for other CTDE algorithms, ~\ref{appsubsec:learning_curves_smac} provides learning curves across additional SMAC scenarios, ~\ref{appsubsec:perf_ega} presents a performance comparison for EGA, ~\ref{appsubsec:computation_cost} discusses computational cost, and ~\ref{appsubsec:general_robustness} includes general robustness experiments.

\subsection{Comparison Results for Other CTDE Algorithms}
\label{appsubsec:compotherctde}

Our proposed Wolfpack attack is compatible with various value-based MARL algorithms. Experimental results in this section demonstrate its significant impact on robustness, not only in QMIX, as discussed in the main text, but also in VDN and QPLEX. The hyperparameters used in these experiments follow those outlined in Appendix \ref{appsubsec:hyper}, with WALL hyperparameters remaining consistent across all algorithms, including QMIX.

{\bf VDN Results:}
Table \ref{table:perf_vdn} shows the average win rates for various robust MARL methods with VDN against attacker baselines. Both models and attackers are trained using the VDN-based algorithm. Results indicate that the proposed Wolfpack attack is more detrimental than Random Attack or EGA for VDN. For example, while EGA reduces Vanilla VDN's performance by $95.6\%-73.8\%=21.8\%$, the Wolfpack attack causes a larger reduction of $95.6\%-44.1\%=51.5\%$, demonstrating its severity. Additionally, the WALL framework outperforms other baselines under both Natural conditions and adversarial attacks, showcasing its robustness.

\begin{table*}[h!]
\centering
\resizebox{0.95\textwidth}{!}{
\begin{tabular}{|c|c|c|c|c|c|c|c|c|}
\hline 
\multicolumn{2}{|c|}{\diagbox[innerwidth=5cm,dir=NW]{Method}{Scenario}}
& 2s3z & 3m & 3s\_vs\_3z & 8m & MMM & 1c3s5z & Mean \\
\hline
\multirow{7}{*}{Natural} 
& Vanilla VDN & $98.5 \pm 1.4$ & $96.0 \pm 3.0$ & $99.3 \pm 0.5$ & $97.8 \pm 2.0$ & $98.3 \pm 0.5$ & $83.5 \pm 9.5$ & $95.6 \pm 1.7$ \\
& RANDOM & $99.0 \pm 0.1$ & $\mathbf{99.8 \pm 0.1}$ & $99.4 \pm 0.5$ & $\mathbf{98.9 \pm 0.1}$ & $98.8 \pm 1.0$ & $97.6 \pm 0.5$ & $98.9 \pm 0.4$ \\
& RARL & $95.3 \pm 0.5$ & $92.5 \pm 3.4$ & $99.3 \pm 0.5$ & $96.3 \pm 1.5$ & $93.2 \pm 1.5$ & $91.8 \pm 0.3$ & $94.7 \pm 0.6$ \\
& RAP & $93.4 \pm 1.3$ & $96.8 \pm 1.8$ & $99.3 \pm 0.5$ & $98.7 \pm 1.0$ & $97.3 \pm 0.5$ & $97.3 \pm 0.5$ & $97.1 \pm 0.4$ \\
& ERNIE & $94.6\pm3.9$ & $99.4\pm0.5$ & $97.1\pm0.5$ & $98.4\pm1.4$ & $98.8\pm0.8$ & $96.5\pm1.7$ & $97.5\pm0.9$ \\
& ROMANCE & $98.2 \pm 0.1$ & $97.9 \pm 1.1$ & $99.4 \pm 0.5$ & $93.5 \pm 5.4$ & $97.9 \pm 1.0$ & $93.0 \pm 3.1$ & $96.6 \pm 0.8$ \\
& WALL (ours) & $\mathbf{99.8 \pm 0.1}$ & $99.4 \pm 0.5$ & $\mathbf{99.9 \pm 0.1}$ & $96.9 \pm 2.0$ & $\mathbf{99.4 \pm 0.5}$ & $\mathbf{100.0 \pm 0.1}$ & $\mathbf{99.2 \pm 0.2}$ \\
\hline
\multirow{7}{*}{Random Attack} 
& Vanilla VDN & $80.5 \pm 1.5$ & $65.5 \pm 6.5$ & $96.5 \pm 0.5$ & $52.5 \pm 10.5$ & $95.0 \pm 1.0$ & $76.5 \pm 7.5$ & $77.8 \pm 1.9$ \\
& RANDOM & $83.0 \pm 1.0$ & $\mathbf{94.5 \pm 1.5}$ & $96.0 \pm 0.7$ & $88.5 \pm 0.5$ & $95.5 \pm 2.5$ & $93.5 \pm 1.5$ & $91.8 \pm 0.5$ \\
& RARL & $81.0 \pm 4.2$ & $64.5 \pm 8.4$ & $97.3 \pm 0.5$ & $72.9 \pm 0.1$ & $76.3 \pm 7.5$ & $92.0 \pm 0.2$ & $80.7 \pm 2.1$ \\
& RAP & $89.8 \pm 3.1$ & $75.8 \pm 37.9$ & $97.7 \pm 1.9$ & $81.2 \pm 0.5$ & $95.2 \pm 2.6$ & $92.3 \pm 0.5$ & $88.7 \pm 1.6$ \\
& ERNIE & $80.2\pm2.0$ & $66.9\pm2.8$ & $93.1\pm1.7$ & $67.1\pm4.9$ & $89.1\pm2.1$ & $90.4\pm4.5$ & $81.1\pm1.2$ \\
& ROMANCE & $91.0 \pm 5.0$ & $79.0 \pm 6.0$ & $98.4 \pm 0.4$ & $54.0 \pm 5.0$ & $97.5 \pm 0.5$ & $92.0 \pm 3.0$ & $85.3 \pm 0.6$ \\
& WALL (ours) & $\mathbf{95.5 \pm 1.5}$ & $86.0 \pm 6.0$ & $\mathbf{99.4 \pm 0.4}$ & $\mathbf{90.0 \pm 3.0}$ & $\mathbf{98.5 \pm 0.5}$ & $\mathbf{98.1 \pm 0.1}$ & $\mathbf{94.6 \pm 1.4}$ \\
\hline
\multirow{7}{*}{EGA} 
& Vanilla VDN & $69.5 \pm 7.5$ & $54.5 \pm 12.5$ & $94.5 \pm 1.5$ & $59.5 \pm 12.5$ & $90.5 \pm 2.5$ & $74.5 \pm 9.5$ & $73.8 \pm 1.0$ \\
& RANDOM & $56.0 \pm 6.0$ & $73.0 \pm 6.0$ & $84.0 \pm 13.0$ & $84.5 \pm 6.5$ & $85.5 \pm 1.5$ & $86.5 \pm 3.5$ & $78.3 \pm 2.4$ \\
& RARL & $58.0 \pm 1.2$ & $79.0 \pm 2.9$ & $96.5 \pm 0.7$ & $76.7 \pm 4.0$ & $75.9 \pm 8.2$ & $81.3 \pm 4.5$ & $77.9 \pm 1.9$ \\
& RAP & $79.9 \pm 3.1$ & $93.0 \pm 4.6$ & $97.3 \pm 0.5$ & $85.7 \pm 1.3$ & $91.8 \pm 3.8$ & $87.2 \pm 1.5$ & $89.3 \pm 0.2$ \\
& ERNIE & $62.0\pm14.9$ & $62.5\pm8.2$ & $89.1\pm6.6$ & $74.7\pm1.6$ & $89.3\pm2.5$ & $82.2\pm1.9$ & $76.6\pm3.6$ \\
& ROMANCE & $86.5 \pm 2.5$ & $\mathbf{93.8 \pm 3.0}$ & $98.0 \pm 0.2$ & $76.5 \pm 0.5$ & $95.5 \pm 2.5$ & $92.0 \pm 0.1$ & $90.3 \pm 0.6$ \\
& WALL (ours) & $\mathbf{91.5 \pm 0.5}$ & $91.0 \pm 2.0$ & $\mathbf{99.0 \pm 1.0}$ & $\mathbf{90.0 \pm 4.0}$ & $\mathbf{97.5 \pm 0.5}$ & $\mathbf{95.5 \pm 0.5}$ & $\mathbf{94.1 \pm 0.1}$ \\
\hline
\multirow{7}{*}{\makecell[c]{Wolfpack\\ Adversarial \\ Attack (ours)}} 
& Vanilla VDN & $54.0 \pm 4.0$ & $20.5 \pm 5.5$ & $91.5 \pm 0.5$ & $24.5 \pm 12.5$ & $18.5 \pm 9.5$ & $55.5 \pm 2.5$ & $44.1 \pm 1.1$ \\
& RANDOM & $47.0 \pm 4.0$ & $89.0 \pm 1.0$ & $90.0 \pm 5.0$ & $41.0 \pm 14.0$ & $18.5 \pm 5.5$ & $83.5 \pm 0.5$ & $61.5 \pm 2.7$ \\
& RARL & $59.5 \pm 8.7$ & $41.3 \pm 16.4$ & $96.8 \pm 0.9$ & $13.1 \pm 2.0$ & $24.3 \pm 6.5$ & $61.6 \pm 11.7$ & $49.4 \pm 1.8$ \\
& RAP & $64.3 \pm 3.5$ & $67.7 \pm 33.8$ & $98.9 \pm 0.1$ & $23.9 \pm 6.1$ & $53.3 \pm 2.5$ & $82.1 \pm 5.4$ & $65.0 \pm 1.7$ \\
& ERNIE & $33.1\pm3.6$ & $25.8\pm6.5$ & $93.0\pm4.5$ & $17.2\pm9.5$ & $23.5\pm9.3$ & $66.4\pm13.5$ & $43.2\pm1.7$ \\
& ROMANCE & $63.0 \pm 10.0$ & $46.5 \pm 20.5$ & $97.5 \pm 0.5$ & $19.5 \pm 7.5$ & $38.0 \pm 5.0$ & $83.0 \pm 4.0$ & $57.9 \pm 3.1$ \\
& WALL (ours) & $\mathbf{91.5 \pm 2.5}$ & $\mathbf{91.5 \pm 2.5}$ & $\mathbf{100.0 \pm 0.0}$ & $\mathbf{71.5 \pm 1.5}$ & $\mathbf{98.0 \pm 1.0}$ & $\mathbf{93.5 \pm 4.5}$ & $\mathbf{91.0 \pm 0.8}$ \\
\hline
\end{tabular}}
\caption{Average test win rates of robust MARL policies under various attack settings (VDN)}
\label{table:perf_vdn}
\end{table*}

\newpage
{\bf QPLEX Results:}
Similarly, Table \ref{table:perf_qplex} reports the average win rates various robust MARL methods with QPLEX against attacker baselines. Both models and attackers are trained using the QPLEX-based algorithm. Results reveal that the Wolfpack attack is also more detrimental for QPLEX compared to Random Attack and EGA. For instance, EGA reduces Vanilla QPLEX's performance by $98.4\% - 57.2\% = 41.2\%$, whereas the Wolfpack attack results in a larger reduction of  $98.4\% - 33.1\% = 65.3\%$. The WALL framework again demonstrates superior robustness, performing well against all attacks, including the Wolfpack attack.

\begin{table*}[h!]
\centering
\resizebox{0.95\textwidth}{!}{
\begin{tabular}{|c|c|c|c|c|c|c|c|c|}
\hline 
\multicolumn{2}{|c|}{\diagbox[innerwidth=5cm,dir=NW]{Method}{Scenario}}
& 2s3z & 3m & 3s\_vs\_3z & 8m & MMM & 1c3s5z & Mean \\ 
\hline
\multirow{7}{*}{Natural} 
& Vanilla QPLEX & $97.1 \pm 1.3$ & $\mathbf{99.2 \pm 0.5}$ & $99.4 \pm 0.5$ & $97.4 \pm 0.3$ & $99.5 \pm 0.5$ & $97.1 \pm 0.9$ & $98.4 \pm 0.1$ \\
& RANDOM & $97.7 \pm 2.1$ & $98.7 \pm 0.8$ & $99.6 \pm 0.1$ & $99.1 \pm 0.9$ & $98.2 \pm 1.2$ & $98.5 \pm 1.3$ & $98.7 \pm 0.5$ \\
& RARL & $94.4 \pm 4.5$ & $89.7 \pm 2.0$ & $88.4 \pm 6.8$ & $97.8 \pm 1.6$ & $98.8 \pm 0.8$ & $94.6 \pm 1.6$ & $94.0 \pm 1.4$ \\
& RAP & $96.5 \pm 1.7$ & $96.6 \pm 2.5$ & $93.2 \pm 0.9$ & $99.1 \pm 0.5$ & $98.8 \pm 0.8$ & $97.8 \pm 0.8$ & $97.7 \pm 0.4$ \\
& ERNIE & $96.7 \pm 2.5$ & $98.8 \pm 0.8$ & $99.7 \pm 0.1$ & $98.8 \pm 1.4$ & $99.1 \pm 0.5$ & $98.4 \pm 1.0$ & $98.6 \pm 0.6$ \\
& ROMANCE & $97.1 \pm 2.9$ & $93.2 \pm 5.9$ & $98.8 \pm 0.8$ & $94.7 \pm 3.2$ & $\mathbf{99.7 \pm 0.1}$ & $99.0 \pm 1.1$ & $96.8 \pm 1.2$ \\
& WALL (ours) & $\mathbf{99.5 \pm 0.5}$ & $97.7 \pm 2.1$ & $\mathbf{99.9 \pm 0.1}$ & $\mathbf{99.8 \pm 0.1}$ & $99.0 \pm 0.7$ & $\mathbf{99.5 \pm 0.6}$ & $\mathbf{99.2 \pm 0.3}$ \\
\hline
\multirow{7}{*}{Random Attack} 
& Vanilla QPLEX & $75.2 \pm 2.5$ & $36.2 \pm 6.6$ & $81.9 \pm 16.0$ & $40.0 \pm 9.8$ & $65.3 \pm 6.2$ & $75.0 \pm 2.8$ & $62.3 \pm 5.6$ \\
& RANDOM & $85.9 \pm 0.9$ & $69.0 \pm 5.8$ & $96.2 \pm 1.6$ & $89.0 \pm 7.7$ & $91.2 \pm 2.4$ & $95.5 \pm 1.2$ & $87.8 \pm 0.7$ \\
& RARL & $81.2 \pm 13.6$ & $61.3 \pm 13.9$ & $74.7 \pm 12.6$ & $73.7 \pm 21.6$ & $92.0 \pm 0.8$ & $92.0 \pm 2.9$ & $79.1 \pm 4.3$ \\
& RAP & $90.8 \pm 2.8$ & $78.0 \pm 7.0$ & $92.1 \pm 10.3$ & $65.3 \pm 4.5$ & $95.0 \pm 1.4$ & $95.7 \pm 1.9$ & $85.0 \pm 1.9$ \\
& ERNIE & $82.0 \pm 2.0$ & $59.0 \pm 9.0$ & $93.0 \pm 1.6$ & $80.0 \pm 1.4$ & $91.3 \pm 0.9$ & $93.0 \pm 0.8$ & $83.1 \pm 1.9$ \\
& ROMANCE & $89.2 \pm 2.1$ & $64.9 \pm 13.4$ & $93.7 \pm 2.1$ & $51.6 \pm 5.3$ & $92.2 \pm 4.1$ & $95.2 \pm 1.0$ & $76.0 \pm 2.8$ \\
& WALL (ours) & $\mathbf{97.4 \pm 0.9}$ & $\mathbf{85.4 \pm 2.5}$ & $\mathbf{99.4 \pm 0.5}$ & $\mathbf{92.9 \pm 3.5}$ & $\mathbf{98.3 \pm 1.2}$ & $\mathbf{98.6 \pm 1.2}$ & $\mathbf{95.3 \pm 0.7}$ \\
\hline
\multirow{7}{*}{EGA} 
& Vanilla QPLEX & $48.1 \pm 3.4$ & $16.0 \pm 5.1$ & $72.5 \pm 15.1$ & $58.5 \pm 16.8$ & $71.0 \pm 4.2$ & $76.9 \pm 1.5$ & $57.2 \pm 6.2$ \\
& RANDOM & $60.5 \pm 6.8$ & $61.4 \pm 14.9$ & $82.0 \pm 6.6$ & $86.2 \pm 3.4$ & $85.3 \pm 2.5$ & $88.8 \pm 2.5$ & $77.4 \pm 3.1$ \\
& RARL & $63.4 \pm 0.5$ & $65.7 \pm 6.2$ & $71.0 \pm 8.2$ & $86.0 \pm 6.5$ & $89.7 \pm 1.9$ & $84.7 \pm 2.1$ & $76.7 \pm 1.0$ \\
& RAP & $78.5 \pm 4.0$ & $83.3 \pm 0.9$ & $85.4 \pm 5.3$ & $89.0 \pm 2.2$ & $92.7 \pm 1.7$ & $96.7 \pm 0.5$ & $88.0 \pm 0.7$ \\
& ERNIE & $64.4 \pm 7.0$ & $67.3 \pm 10.3$ & $51.7 \pm 9.5$ & $86.7 \pm 4.7$ & $86.0 \pm 5.0$ & $89.3 \pm 4.5$ & $74.2 \pm 3.0$ \\
& ROMANCE & $79.5 \pm 6.0$ & $80.3 \pm 1.7$ & $90.3 \pm 1.3$ & $80.7 \pm 8.3$ & $95.2 \pm 2.0$ & $92.5 \pm 1.3$ & $85.6 \pm 1.0$ \\
& WALL (ours) & $\mathbf{89.0 \pm 3.8}$ & $\mathbf{83.9 \pm 2.9}$ & $\mathbf{99.6 \pm 0.6}$ & $\mathbf{94.4 \pm 1.2}$ & $\mathbf{96.4 \pm 1.2}$ & $\mathbf{96.1 \pm 0.7}$ & $\mathbf{93.2 \pm 0.7}$ \\
\hline
\multirow{7}{*}{\makecell[c]{Wolfpack\\ Adversarial \\ Attack (ours)}} 
& Vanilla QPLEX & $30.8 \pm 4.3$ & $11.8 \pm 7.5$ & $63.7 \pm 24.3$ & $30.7 \pm 11.1$ & $20.0 \pm 12.5$ & $41.9 \pm 6.1$ & $33.1 \pm 7.1$ \\
& RANDOM & $50.5 \pm 2.9$ & $16.8 \pm 6.8$ & $89.5 \pm 5.7$ & $45.6 \pm 18.0$ & $34.0 \pm 14.1$ & $82.5 \pm 11.3$ & $53.2 \pm 4.9$ \\
& RARL & $55.5 \pm 2.1$ & $27.3 \pm 9.2$ & $78.0 \pm 4.3$ & $46.0 \pm 14.9$ & $21.3 \pm 14.4$ & $79.7 \pm 6.9$ & $51.3 \pm 2.8$ \\
& RAP & $59.0 \pm 4.7$ & $50.3 \pm 14.7$ & $86.7 \pm 8.2$ & $33.7 \pm 3.1$ & $45.0 \pm 7.0$ & $93.7 \pm 2.4$ & $56.3 \pm 3.2$ \\
& ERNIE & $52.9 \pm 6.3$ & $49.0 \pm 9.9$ & $76.0 \pm 9.9$ & $42.7 \pm 10.6$ & $20.7 \pm 10.4$ & $78.7 \pm 6.7$ & $53.3 \pm 3.6$ \\
& ROMANCE & $57.3 \pm 8.4$ & $38.5 \pm 13.9$ & $89.7 \pm 1.3$ & $28.6 \pm 1.1$ & $46.2 \pm 10.6$ & $83.5 \pm 5.5$ & $50.8 \pm 0.9$ \\
& WALL (ours) & $\mathbf{88.3 \pm 1.2}$ & $\mathbf{87.6 \pm 4.7}$ & $\mathbf{99.7 \pm 0.5}$ & $\mathbf{84.5 \pm 1.7}$ & $\mathbf{96.3 \pm 3.9}$ & $\mathbf{99.3 \pm 0.9}$ & $\mathbf{92.6 \pm 1.3}$ \\
\hline
\end{tabular}}
\caption{Average test win rates of robust MARL policies under various attack settings (QPLEX)}
\label{table:perf_qplex}
\end{table*}

\begin{figure}[!h]
    \centering
    \begin{subfigure}{0.45\columnwidth}
        \centering
        \includegraphics[width=\linewidth]{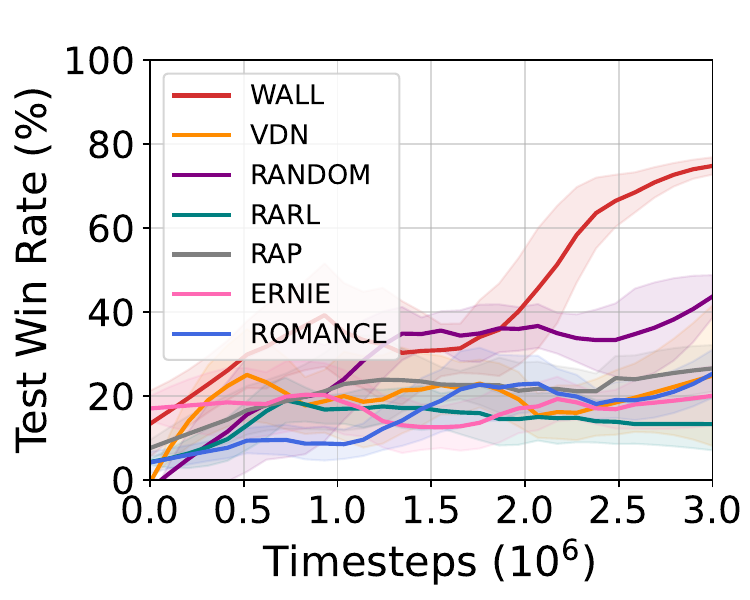}
        \caption{\texttt{8m}}  
    \end{subfigure}
    \begin{subfigure}{0.45\columnwidth}
        \centering
        \includegraphics[width=\linewidth]{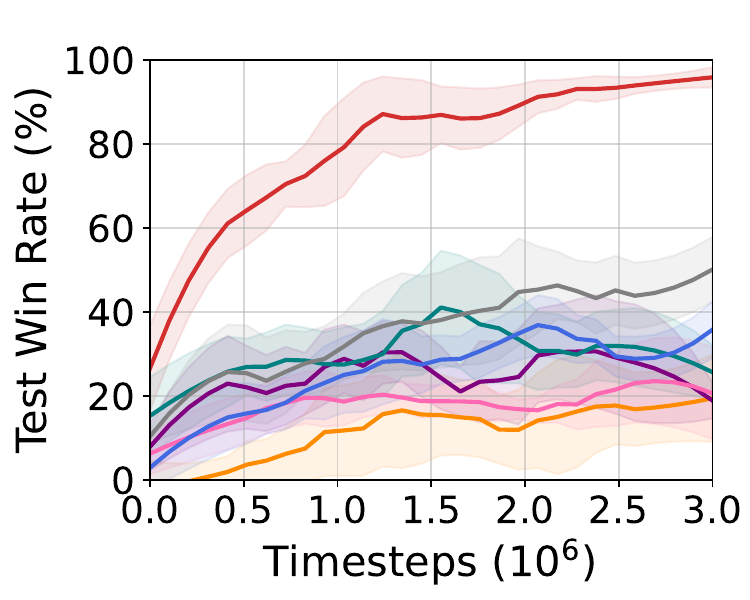}
        \caption{\texttt{MMM}}  
    \end{subfigure}\hfill
    \caption{Learning curves of MARL methods for Wolfpack attack (VDN)}
    \label{fig:vdn}
\end{figure}

\begin{figure}[!h]
    \centering
    \begin{subfigure}{0.45\columnwidth}
        \centering
        \includegraphics[width=\linewidth]{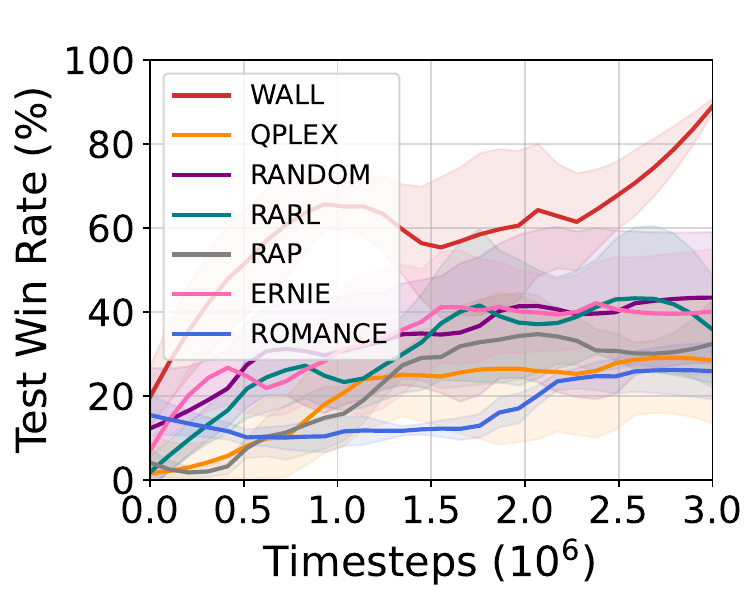}
        \caption{\texttt{8m}}  
    \end{subfigure}
    \begin{subfigure}{0.45\columnwidth}
        \centering
        \includegraphics[width=\linewidth]{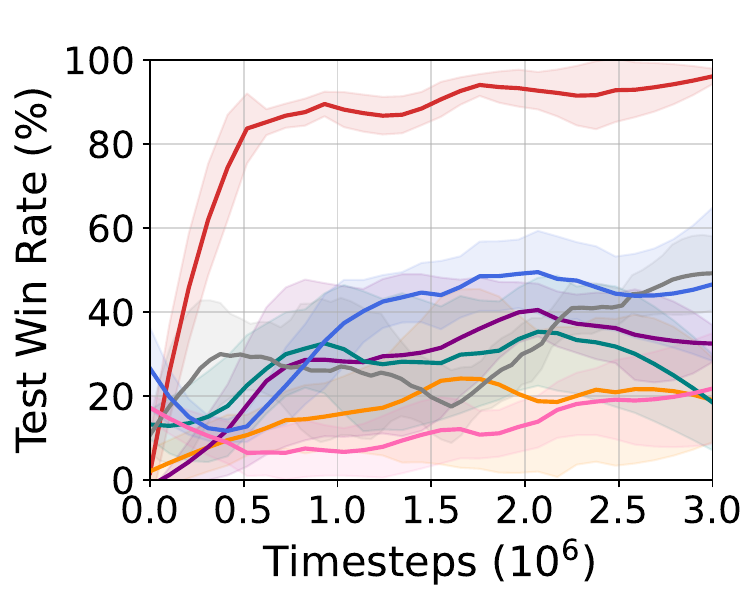}
        \caption{\texttt{MMM}}  
    \end{subfigure}
    \caption{Learning curves of MARL methods for Wolfpack attack (QPLEX)}
    \label{fig:qplex}
\end{figure}

\newpage
{\bf Learning Curves for VDN and QPLEX:}
We also analyze training curves and average test win rates across different CTDE algorithms. Graphs for the \texttt{8m} and \texttt{MMM} environments illustrate the average win rates of each policy over training steps under unseen Wolfpack adversarial attacks. Fig. \ref{fig:vdn} presents training curves for VDN, while Fig. \ref{fig:qplex} shows results for QPLEX. These curves highlight that WALL not only achieves greater robustness but also adapts more quickly to attacks across VDN and QPLEX, further confirming its effectiveness beyond QMIX.

\newpage

\subsection{Learning Curves Across Additional SMAC Scenarios}
\label{appsubsec:learning_curves_smac}
\begin{figure}[ht!]
    \centering
    \vspace{-0.1in}
    \begin{subfigure}{0.32\textwidth}
        \centering
        \includegraphics[width=\linewidth]{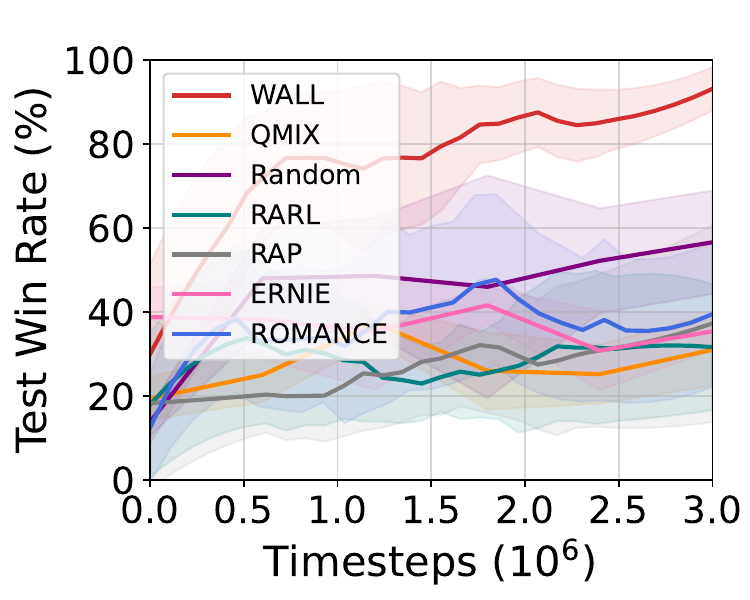}
        \caption{\texttt{3m}}
    \end{subfigure}
    \begin{subfigure}{0.32\textwidth}
        \centering
        \includegraphics[width=\linewidth]{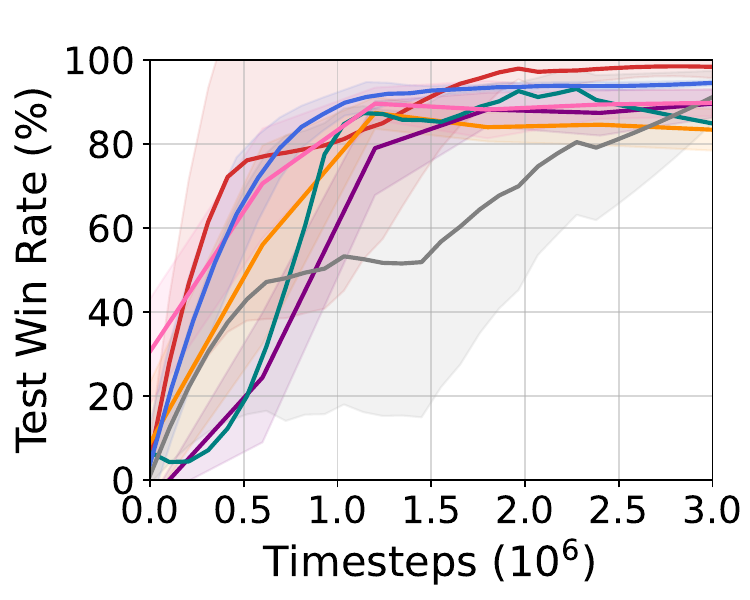}
        \caption{\texttt{3s\_vs\_3z}}
    \end{subfigure}
    \begin{subfigure}{0.32\textwidth}
        \centering
        \includegraphics[width=\linewidth]{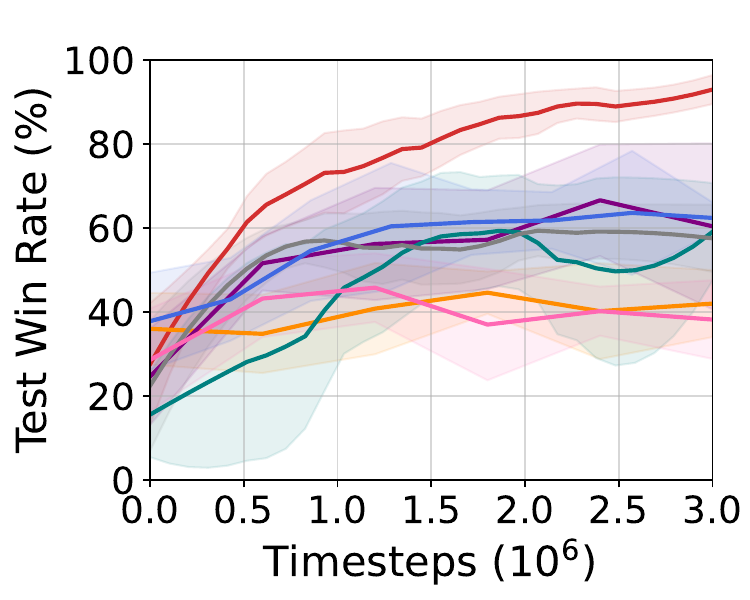}
        \caption{\texttt{2s3z}}
    \end{subfigure}

    \vspace{0.1in}
    \begin{subfigure}{0.32\textwidth}
        \centering
        \includegraphics[width=\linewidth]{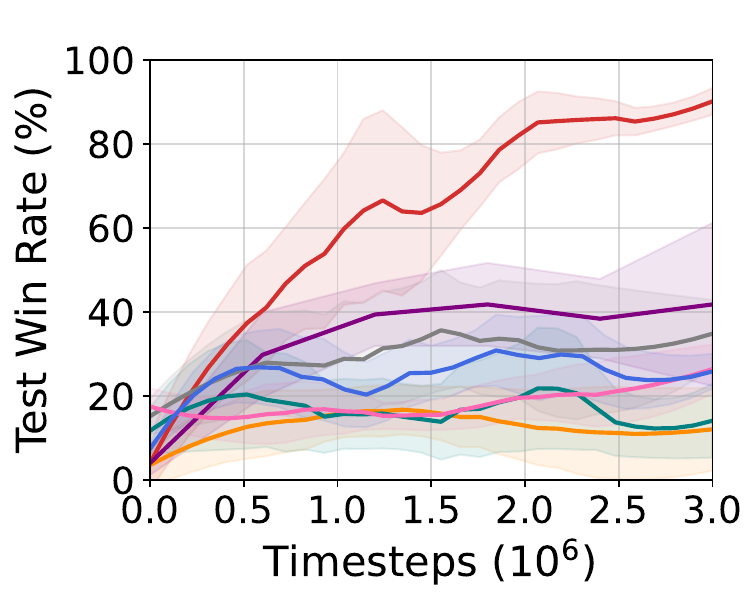}
        \caption{\texttt{8m}}
    \end{subfigure}
    \begin{subfigure}{0.32\textwidth}
        \centering
        \includegraphics[width=\linewidth]{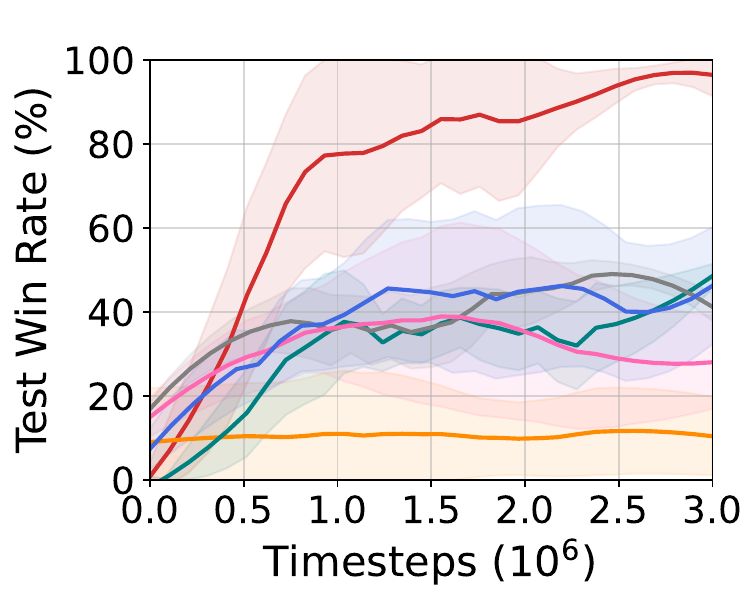}
        \caption{\texttt{MMM}}
    \end{subfigure}
    \begin{subfigure}{0.32\textwidth}
        \centering
        \includegraphics[width=\linewidth]{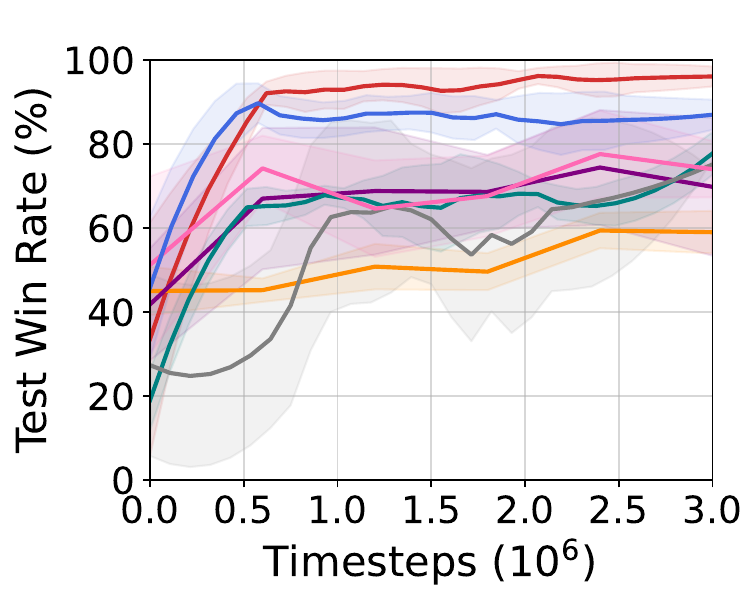}
        \caption{\texttt{1c3s5z}}
    \end{subfigure}
    \vspace{-0.1in}
    \caption{Learning curves of MARL methods for Wolfpack attack across 6 SMAC scenarios (QMIX)}
    \label{fig:wolfpack_learning_2}
\end{figure}

In this section, we provide the training performance of WALL under Wolfpack adversarial attack across additional SMAC scenarios beyond the \texttt{8m} and \texttt{MMM} environments, which are emphasized in the main text for their significant performance differences. Fig. \ref{fig:wolfpack_learning_2} illustrates the training curves for 6 scenarios: \texttt{3m}, \texttt{3s\_vs\_3z}, \texttt{2s3z}, \texttt{8m}, \texttt{MMM}, and \texttt{1c3s5z}. The results demonstrate that WALL consistently outperforms baseline methods, achieving superior win rates across all scenarios. Additionally, policies trained with WALL adapt more quickly to the challenges posed by Wolfpack attack, showing robust and efficient performance across environments of varying complexity. These findings highlight the effectiveness of WALL in enhancing the robustness of MARL policies against coordinated adversarial attacks.

\newpage
\subsection{Performance Comparison for EGA}
\label{appsubsec:perf_ega}
\begin{figure}[ht!]
    \centering
    \begin{subfigure}{0.32\textwidth}
        \centering
        \includegraphics[width=\linewidth]{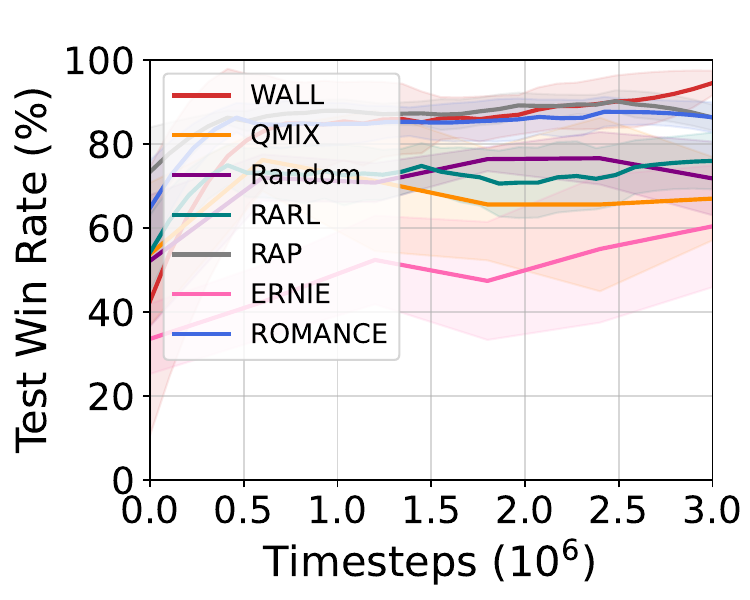}
        \caption{\texttt{3m}}
    \end{subfigure}
    \begin{subfigure}{0.32\textwidth}
        \centering
        \includegraphics[width=\linewidth]{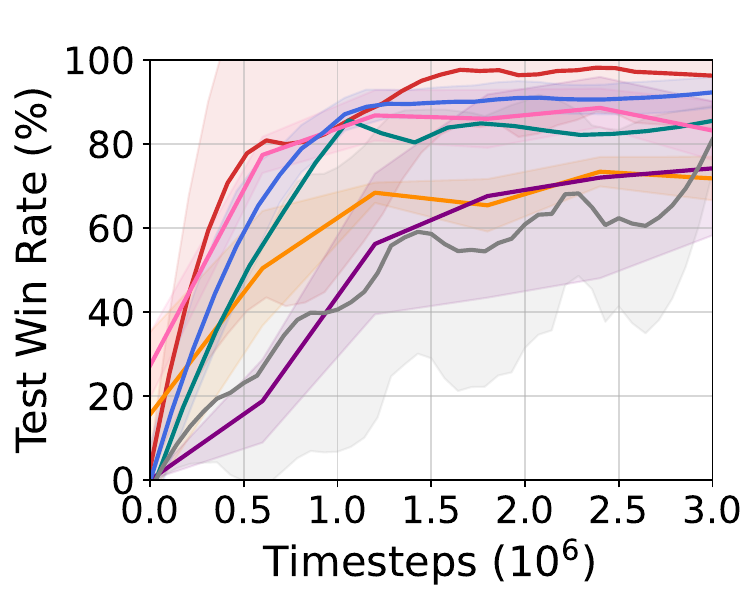}
        \caption{\texttt{3s\_vs\_3z}}
    \end{subfigure}
    \begin{subfigure}{0.32\textwidth}
        \centering
        \includegraphics[width=\linewidth]{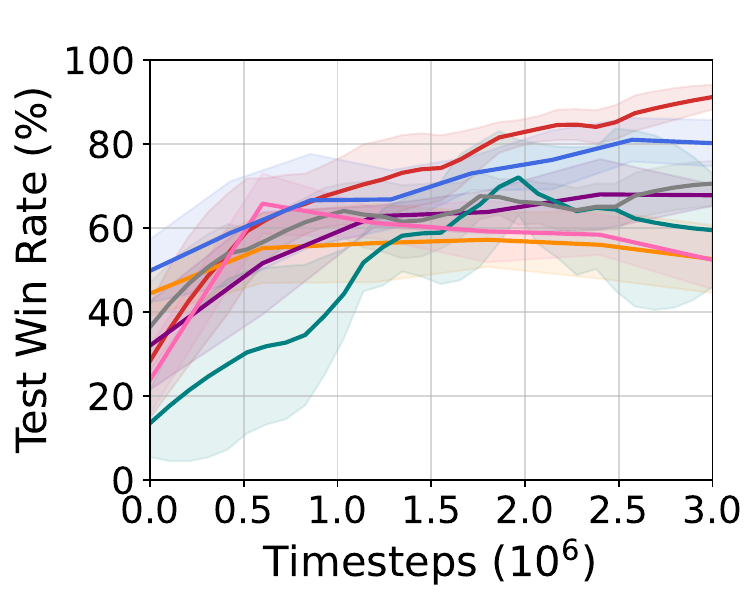}
        \caption{\texttt{2s3z}}
    \end{subfigure}

    \vspace{0.1in}
    \begin{subfigure}{0.32\textwidth}
        \centering
        \includegraphics[width=\linewidth]{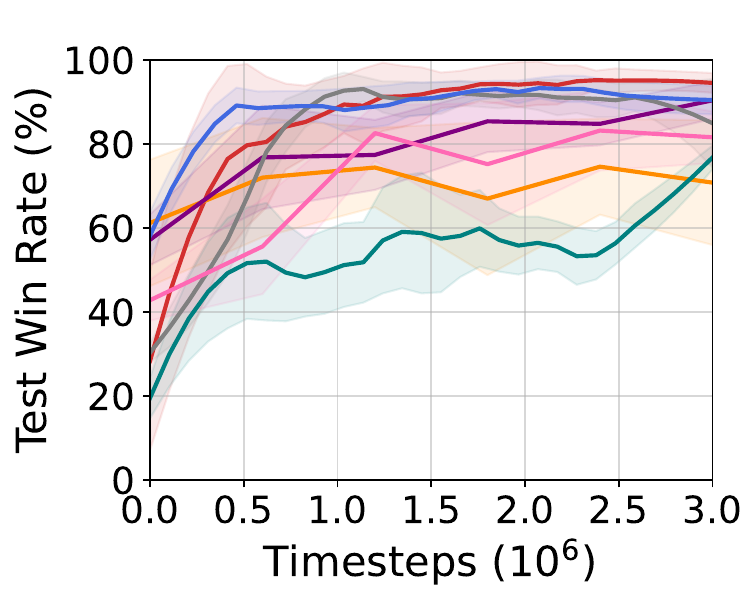}
        \caption{\texttt{8m}}
    \end{subfigure}
    \begin{subfigure}{0.32\textwidth}
        \centering
        \includegraphics[width=\linewidth]{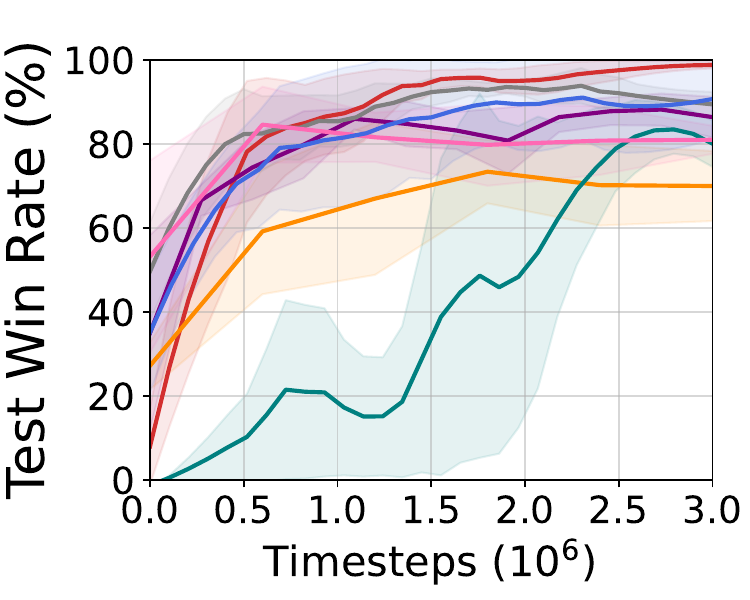}
        \caption{\texttt{MMM}}
    \end{subfigure}
    \begin{subfigure}{0.32\textwidth}
        \centering
        \includegraphics[width=\linewidth]{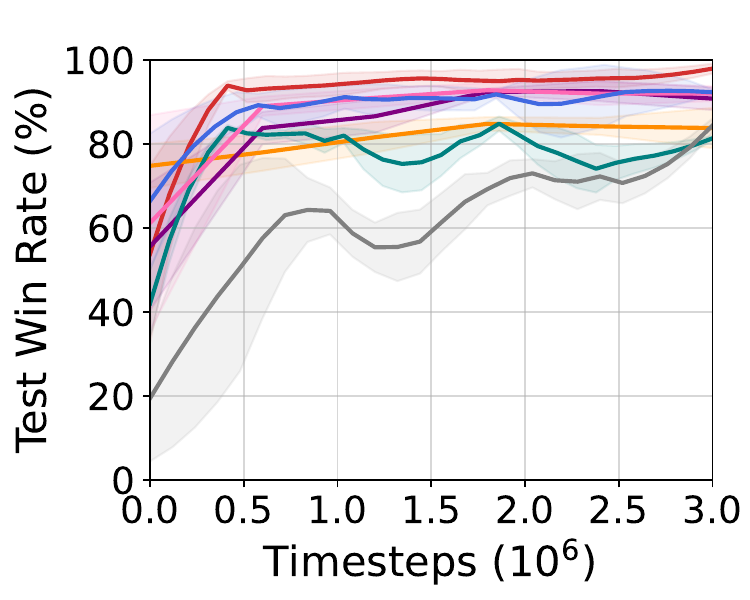}
        \caption{\texttt{1c3s5z}}
    \end{subfigure}

    \caption{Learning curves of MARL methods for EGA across 6 SMAC scenarios (QMIX)}
    \label{fig:EGA_learning}
\end{figure}

This section presents the training performance of WALL under the existing Evolutionary Generation-based Attackers (EGA) \cite{yuan2023robust} method across six SMAC scenarios, as shown in Fig. \ref{fig:EGA_learning}. The results demonstrate that WALL consistently achieves superior robustness across all scenarios, even against unseen EGA adversaries that are not included in its training process.

The EGA framework generates a diverse and high-quality population of adversarial attackers. Unlike single-adversary methods, EGA maintains an evolving archive of attackers optimized for both quality and diversity, ensuring robust evaluations against various attack strategies. During training, attackers are randomly selected from the archive to simulate diverse attack scenarios. The archive is iteratively updated by replacing low-quality or redundant attackers with newly generated ones.
For evaluation, an attacker policy is randomly selected from the archive. The chosen attacker identifies critical attack steps and targets specific victim agents, introducing action perturbations to reduce their individual $Q$-values. WALL demonstrates strong resilience in these challenging environments, effectively mitigating the impact of EGA adversaries and maintaining high performance across all evaluated scenarios.

\newpage

\subsection{Computational Cost}
\label{appsubsec:computation_cost}

 We measured the total training time (for 3M steps) required for each algorithm in the SMAC environments. The results are summarized in Table \ref{table:compute_cost}. WALL requires approximately 30\% more time than ROMANCE, primarily due to the use of the Transformer. However, despite this additional cost, the Transformer-based critical step selection plays an essential role in the effectiveness of our method, as it enables precise identification of attack timing where coordinated perturbations are most disruptive. This capability allows the robust policy to anticipate and defend against strategically timed adversarial threats, ultimately resulting in stronger robustness. In general, training on the \texttt{MMM} map takes longer than on the \texttt{8m} map due to the increased number of agents, which leads to higher computational costs as a result of scalability.
 
\begin{table}[h!]
\centering
\resizebox{0.6\textwidth}{!}{
\begin{tabular}{|c|c|c|c|}
\hline
\multicolumn{2}{|c|}{\diagbox[innerwidth=5.5cm,dir=NW]{Method}{Scenario}} & 8m & MMM \\
\hline
\multirow{7}{*}{Training Time}
& Vanilla QMIX & $6\text{h}\ 30\text{m}$ & $7\text{h}\ 30\text{m}$ \\
& RANDOM       & $6\text{h}\ 30\text{m}$ & $7\text{h}\ 35\text{m}$ \\
& RARL         & $11\text{h}\ 10\text{m}$ & $15\text{h}\ 50\text{m}$ \\
& RAP          & $14\text{h}\ 35\text{m}$ & $17\text{h}\ 25\text{m}$ \\
& ERNIE        & $16\text{h}\ 30\text{m}$ & $20\text{h}\ 55\text{m}$ \\
& ROMANCE      & $16\text{h}\ 35\text{m}$ & $18\text{h}\ 45\text{m}$ \\
& WALL (ours)  & $21\text{h}\ 05\text{m}$ & $23\text{h}\ 30\text{m}$ \\
\hline
\end{tabular}
}
\caption{Training time comparison on \texttt{8m} and \texttt{MMM}}
\label{table:compute_cost}
\end{table}

\subsection{General Robustness Experiments}
\label{appsubsec:general_robustness}

We conducted additional experiments to evaluate robustness under other commonly used criteria in the \texttt{8m} and \texttt{MMM} environments. Specifically, we considered: (1) Gaussian observation noise: Standard Gaussian noise (mean 0, standard deviation 1) is injected into agents’ local observations, where attack steps (total attack steps 
: 8 for \texttt{8m}, 16 for \texttt{MMM}) and the attack group are selected randomly; and (2) Different parameterization in the SMAC test environment: Robustness is evaluated under perturbed unit attributes by reducing the initial health of allied units by 10\%, 15\%, and 20\% in the test setup, compared to the training configuration. These types of noise and perturbations are standard in robustness evaluations and enable us to assess how well the proposed method generalizes beyond action-level perturbations. The Table \ref{table:general_robustness} below summarizes the results. In both settings, WALL demonstrates consistently stronger performance than existing baselines, suggesting that it generalizes well to broader forms of distributional shift.

\begin{table}[h!]
\centering
\resizebox{0.7\textwidth}{!}{
\begin{tabular}{|c|c|c|c|c|}
\hline
\multicolumn{2}{|c|}{\diagbox[innerwidth=6cm,dir=NW]{Method}{Scenario}} & 8m & MMM & Mean \\
\hline
\multirow{4}{*}{Gaussian Obs. Noise}
& Vanilla QMIX & $62.6 \pm 2.3$ & $75.3 \pm 3.1$ & $69.0 \pm 2.1$ \\
& RANDOM       & $72.3 \pm 4.2$ & $79.3 \pm 3.9$ & $75.8 \pm 7.2$ \\
& ROMANCE      & $69.6 \pm 11.2$ & $76.6 \pm 7.5$ & $73.1 \pm 6.6$ \\
& WALL (ours)  & $\mathbf{91.3 \pm 3.3}$ & $\mathbf{97.3 \pm 1.7}$ & $\mathbf{94.3 \pm 3.3}$ \\
\hline
\multirow{4}{*}{Ally HP ↓ 10\%}
& Vanilla QMIX & $52.8 \pm 4.8$ & $88.4 \pm 2.7$ & $70.6 \pm 1.8$ \\
& RANDOM       & $50.8 \pm 8.2$ & $95.4 \pm 2.5$ & $73.1 \pm 5.2$ \\
& ROMANCE      & $56.8 \pm 16.0$ & $92.4 \pm 4.5$ & $74.6 \pm 8.5$ \\
& WALL (ours)  & $\mathbf{73.2 \pm 7.5}$ & $\mathbf{98.6 \pm 1.1}$ & $\mathbf{85.9 \pm 3.7}$ \\
\hline
\multirow{4}{*}{Ally HP ↓ 15\%}
& Vanilla QMIX & $47.4 \pm 6.9$ & $69.0 \pm 5.4$ & $58.2 \pm 5.8$ \\
& RANDOM       & $49.2 \pm 4.5$ & $81.4 \pm 5.5$ & $65.3 \pm 4.6$ \\
& ROMANCE      & $57.6 \pm 15.7$ & $81.2 \pm 3.5$ & $69.4 \pm 7.8$ \\
& WALL (ours)  & $\mathbf{69.2 \pm 10.5}$ & $\mathbf{94.0 \pm 3.2}$ & $\mathbf{81.6 \pm 5.7}$ \\
\hline
\multirow{4}{*}{Ally HP ↓ 20\%}
& Vanilla QMIX & $0.3 \pm 0.4$  & $41.2 \pm 4.5$ & $20.8 \pm 2.1$ \\
& RANDOM       & $0.0 \pm 0.0$  & $65.0 \pm 2.7$ & $32.5 \pm 1.3$ \\
& ROMANCE      & $2.3 \pm 0.4$  & $57.0 \pm 8.9$ & $29.7 \pm 4.3$ \\
& WALL (ours)  & $\mathbf{4.3 \pm 1.9}$ & $\mathbf{89.6 \pm 5.6}$ & $\mathbf{47.0 \pm 3.3}$ \\
\hline
\end{tabular}
}
\caption{Average test win rates of robust MARL policies under various perturbation settings}
\vspace{-0.2in}
\label{table:general_robustness}
\end{table}

\newpage
\section{Additional Ablation studies}
\label{appsec:addabl}
In this section, we provide additional ablation studies on the number of Wolfpack adversarial attacks $K_{\mathrm{WP}}$ and the attack duration $t_{\mathrm{WP}}$ in the \texttt{8m} and \texttt{MMM} environments, where the performance differences between WALL and other robust MARL methods are most pronounced.

\begin{figure}[ht!]
    \centering
    \vspace{-0.1in}
    \begin{subfigure}{0.38\columnwidth}
        \centering
        \includegraphics[width=\linewidth]{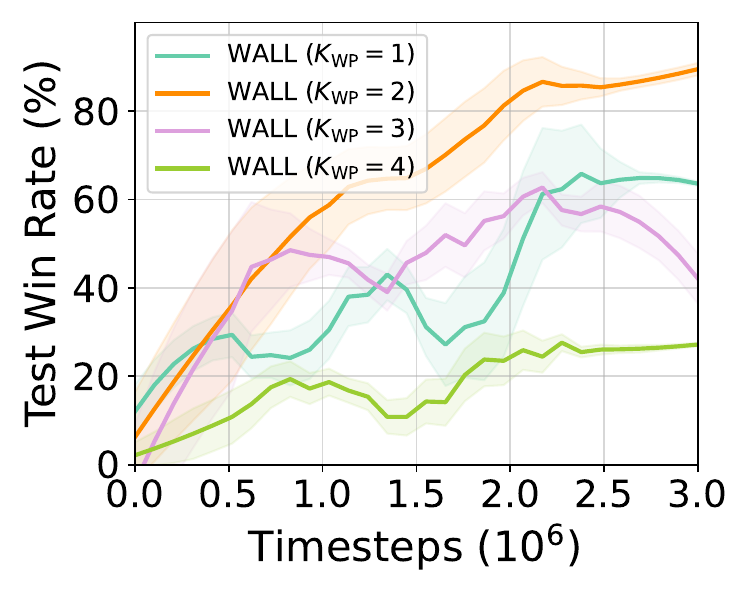}
        \caption{\texttt{8m}}  
    \end{subfigure}
    \begin{subfigure}{0.38\columnwidth}
        \centering
        \includegraphics[width=\linewidth]{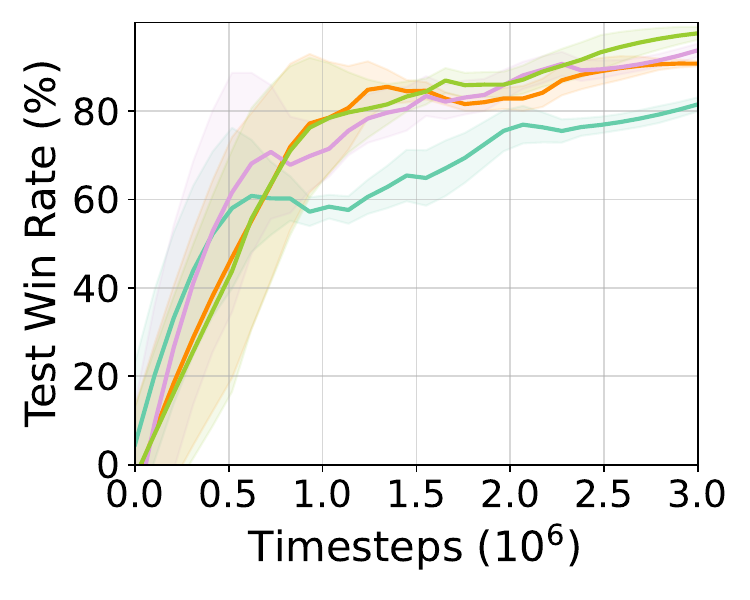}
        \caption{\texttt{MMM}}  
    \end{subfigure}
    \vspace{-0.1in}
    \caption{Number of Wolfpack attacks ($K_{\mathrm{WP}}$)}
    \label{fig:attack_num}
\end{figure}

{\bf Number of Wolfpack Attacks $K_{\mathrm{WP}}$:}
The hyperparameter $K_{\mathrm{WP}}$ determines the number of Wolfpack attacks, with each attack consisting of an initial attack and follow-up attacks over $t_{\mathrm{WP}}=3$ timesteps. The total number of attack steps $K$ for Wolfpack attack is then calculated as $K=4\times K_{\mathrm{WP}}$. In this section, we conduct a parameter search for $K_{\mathrm{WP}}\in[1,2,3,4]$. Fig. \ref{fig:attack_num} illustrates the robustness of WALL policies trained with different $K_{\mathrm{WP}}$  values under the default Wolfpack attack in \texttt{8m} and \texttt{MMM}. In both environments, having too small $K_{\mathrm{WP}}$ results in insufficiently severe attacks, which leads to reduced robustness of the WALL framework. Conversely, in the \texttt{8m} environment, excessively large $K_{\mathrm{WP}}$ values create overly devastating attacks, making it difficult for CTDE methods to learn strategies to counter the Wolfpack attack, which degrades learning performance. Therefore, an optimal $K_{\mathrm{WP}}$  exists in both environments: $K_{\mathrm{WP}}=2$ for \texttt{8m} and $K_{\mathrm{WP}}=4$ for \texttt{MMM}, which we choose as the default hyperparameters.

\begin{figure}[ht!]
    \centering
    \vspace{-0.1in}
    \begin{subfigure}{0.38\columnwidth}
        \centering
        \includegraphics[width=\linewidth]{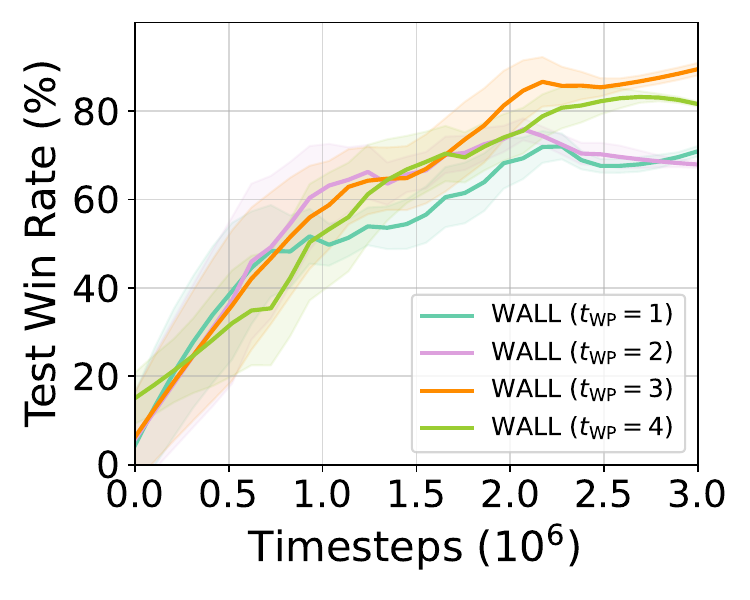}
        \caption{\texttt{8m}}  
    \end{subfigure}
    \begin{subfigure}{0.38\columnwidth}
        \centering
        \includegraphics[width=\linewidth]{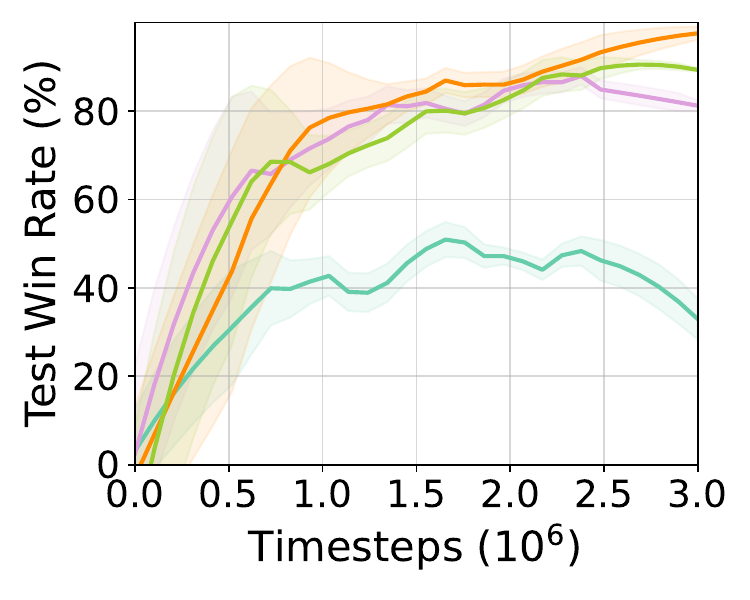}
        \caption{\texttt{MMM}}  
    \end{subfigure}
    \vspace{-0.1in}
    \caption{Attack duration ($t_{\mathrm{WP}}$)}
    \label{fig:Attack duration}
\end{figure}

{\bf Attack duration $t_{\mathrm{WP}}$:}
The hyperparameter $t_{\mathrm{WP}}$ determines the duration of follow-up attacks after the initial attack. To analyze its impact on robustness, Fig. \ref{fig:Attack duration} compares performance for $t_{\mathrm{WP}}\in[1,2,3,4]$ in the \texttt{8m} and \texttt{MMM} environments. As shown in the figure, similar to the case of $K_{\mathrm{WP}}$, setting $t_{\mathrm{WP}}$ too low results in insufficient follow-up attacks on assisting agents, reducing the severity of the attack and lowering the robustness of WALL. On the other hand, excessively high  $t_{\mathrm{WP}}$ values lead to overly severe attacks, making it challenging for WALL to learn effective defenses against the Wolfpack attack. Both environments demonstrate that $t_{\mathrm{WP}}=3$ yields optimal performance and is selected as the best hyperparameter.

\newpage
\section{Additional Visualizations of Wolfpack Adversarial Attack}
\label{appsec:vis}

\subsection{Visualization of Wolfpack Adversarial Attack Across Additional SMAC Scenarios}
\begin{figure*}[h!]
    \begin{center}
    \centerline{\includegraphics[width=0.98\textwidth]{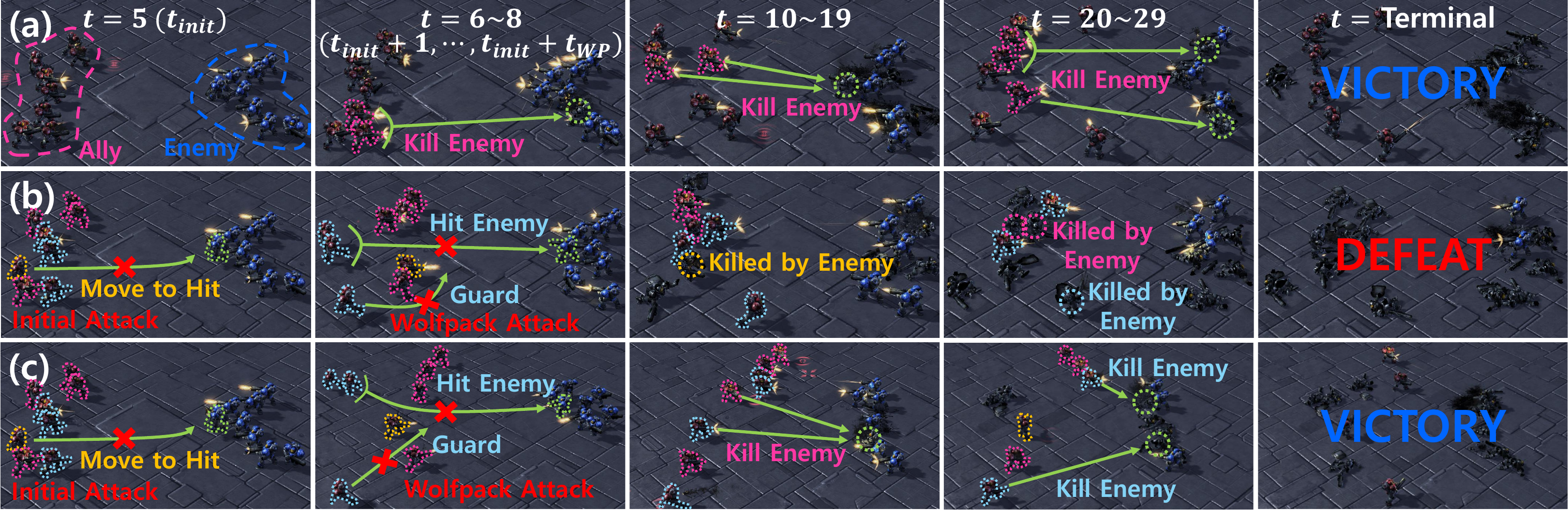}}
    \caption{Attack comparison on \texttt{8m} task in the SMAC: (a) QMIX/Natural, (b) QMIX/Wolfpack attack, and (c) WALL/Wolfpack attack}
    \label{fig:vis_8m}
    \vspace{-0.1in}
    \end{center}
\end{figure*}
\begin{figure*}[h!]
    \begin{center}
    \centerline{\includegraphics[width=0.98\textwidth]{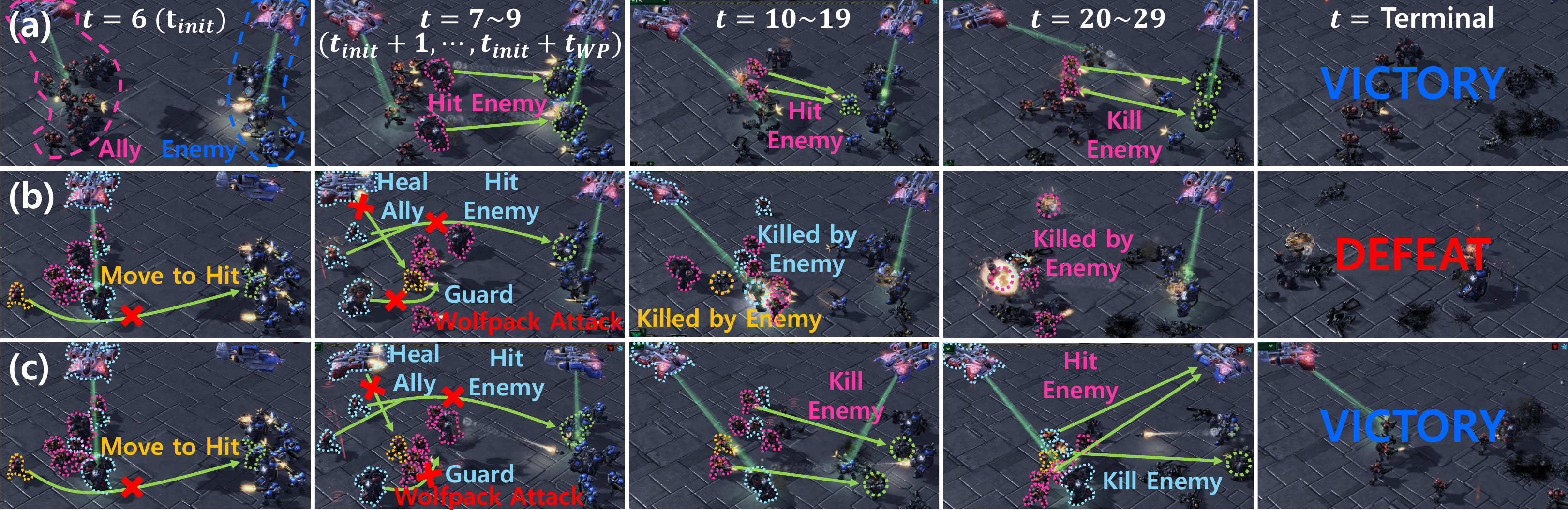}}
    \vspace{-0.1in}
    \caption{Attack comparison on \texttt{MMM} task in the SMAC: (a) QMIX/Natural, (b) QMIX/Wolfpack attack, and (c) WALL/Wolfpack attack}
    \label{fig:vis_MMM}
    \vspace{-0.1in}
    \end{center}
\end{figure*}

To analyze the superior performance of the Wolfpack attack, we provide a visualization of its execution in various SMAC environments. Fig. \ref{fig:visualization} illustrates the \texttt{2s3z} task, while Fig. \ref{fig:vis_8m} visualizes the \texttt{8m} task, and Fig. \ref{fig:vis_MMM} presents the \texttt{MMM} task.

Fig. \ref{fig:vis_8m}(a) illustrates Vanilla QMIX operating in a natural scenario without any attack, where the agents successfully defeat all enemy units and achieve victory. In this scenario, agents with low health continuously move to the backline to avoid enemy attacks, while agents with higher health position themselves at the frontline to absorb damage. This dynamic coordination enables the team to manage their resources effectively, withstand enemy attacks, and secure a successful outcome.

Fig. \ref{fig:vis_8m}(b) depicts Vanilla QMIX under the Wolfpack adversarial attack, where an initial attack is launched at $t=5$, and follow-up agents are targeted between $t=6$ and $t=8$. Agents with higher remaining health are selected as follow-up agents, preventing them from guarding the targeted ally or engaging the enemy effectively. Between $t=10$ and $t=19$, the initial agent continues to take focused enemy fire, eventually succumbing to the attacks and being eliminated. This disruption renders the remaining agents ineffective in defending against the adversarial attack, leading to a loss as all agents are defeated. 

Fig. \ref{fig:vis_8m}(c) shows the policy trained with the WALL framework. During $t=6$ to $t=8$, the same agents as in (b) are selected as follow-up agents and subjected to the Wolfpack attack, limiting their ability to guard or engage the enemy. Nevertheless, the non-attacked agents adjust by forming a wider formation vertically, effectively dispersing enemy firepower while delivering coordinated attacks. Additionally, agents with higher health move forward to guard the initial agents, ensuring that the initial agents do not die. This tactical adaptation enables the team to eliminate enemy units and secure victory.

\newpage
Fig. \ref{fig:vis_MMM}(a) showcases Vanilla QMIX operating in a natural scenario without any adversarial interference, where all enemy units are successfully eliminated, leading to a decisive victory. During this process, agents with lower health retreat to the back while the Medivac agent provides healing, and agents with higher health move forward to absorb enemy attacks. This coordinated strategy enables the team to secure victory efficiently.

Fig. \ref{fig:vis_MMM}(b) illustrates Vanilla QMIX under the Wolfpack adversarial attack. An initial attack is launched at $t=6$, followed by the targeting of four follow-up agents between $t=7$ and $t=9$. The follow-up agents selected include one healing agent attempting to heal the initial agent, one guarding agent positioned to protect the initial agent, and two agents actively engaging the enemy targeting the initial agent. Due to the Wolfpack attack, the initial agent failed to receive critical healing or guarding support at $t=10$ and $t=19$, leading to its elimination. This disruption severely hinders the remaining agents' ability to defend against the adversarial attack, ultimately resulting in a loss as all agents are defeated. 

Fig. \ref{fig:vis_MMM}(c) illustrates the policy trained with the WALL framework. During $t=7$ to $t=9$, the same agents as in (b) are selected as follow-up agents and subjected to the Wolfpack attack, restricting their actions such as healing, guarding, or targeting the enemy effectively. However, the non-attacked agents adapt by positioning themselves ahead of the initial agent to provide protection and focus their fire on the enemies targeting the initial agent. This strategic adaptation enables the team to successfully repel the adversarial attack, eliminate enemy units, and secure victory. Notably, in the terminal timesteps, more agents survive under the WALL framework compared to the natural scenario depicted in (a), highlighting the enhanced robustness and stability of the policy learned with WALL.

\newpage

\newpage
\subsection{Additional Analysis of Follow-up Agent Group Selection}
\label{appsec:followup}

\begin{figure}[h!]
    \centering
    \vspace{-0.1in}
    \begin{subfigure}{0.3\columnwidth}
        \centering
        \includegraphics[width=1.0\textwidth]{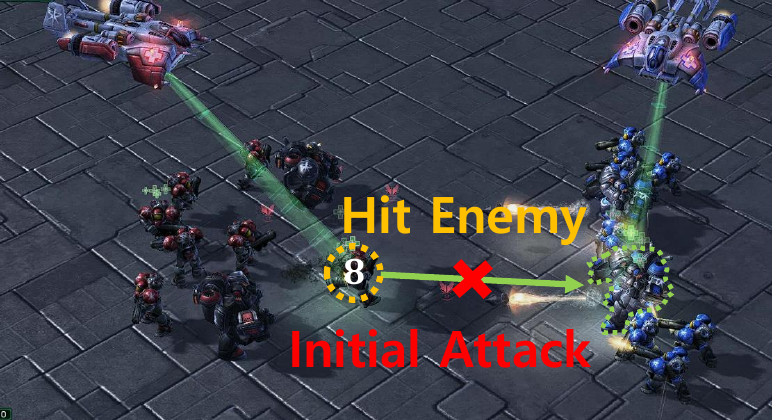}
        \caption{Initial attack}  
    \end{subfigure}
    \begin{subfigure}{0.3\columnwidth}
        \centering
        \includegraphics[width=1.0\textwidth]{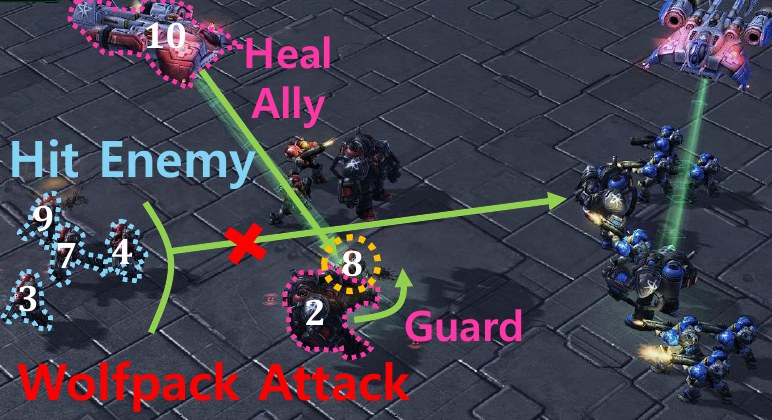}
        \caption{Follow-up (L2)}  
    \end{subfigure}
    \begin{subfigure}{0.3\columnwidth}
        \centering
        \includegraphics[width=1.0\textwidth]{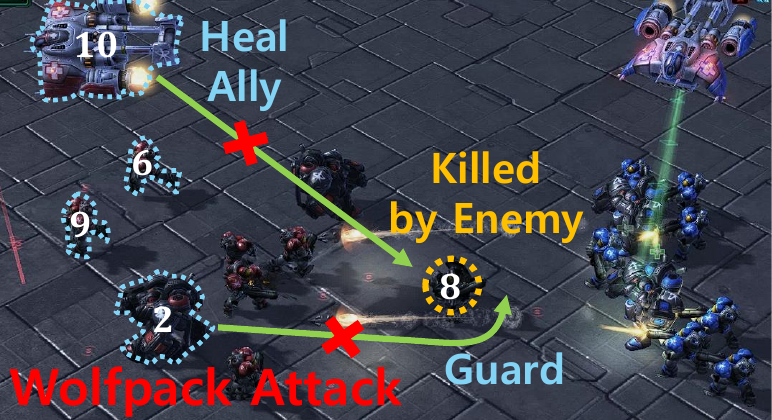}
        \caption{Follow-up agent group selection}  
    \end{subfigure}
    \vspace{-0.1in}
    \caption{Visualization of follow-up agent group selection comparison for the \texttt{MMM} task in SMAC}
    \label{fig:followup}
\end{figure}
In this section, we demonstrate that the proposed Follow-up Agent Group Selection method effectively identifies responding agents that protect the initially attacked agent. Fig. \ref{fig:followup} visualizes the process of selecting the follow-up agent group after an initial attack. By comparing the proposed method with a baseline method, Follow-up (L2), which selects $m$ agents closest to the initial agent based on observation L2 distance, we show that our method better identifies responding agents, enabling a more impactful Wolfpack attack.

Fig. \ref{fig:followup}(a) illustrates an initial attack on agent $8$, preventing it from performing its original action of hitting the enemy and forcing it to move forward, exposing it to enemy attacks. Fig. \ref{fig:followup}(b) shows the Follow-up (L2) method selecting agents $3, 4, 7, 9$ as the follow-up agents based on their proximity to the initial agent. Despite the follow-up attack, non-attacked agents, which is far from the initial agent in terms of observation L2 distance, such as agent $10$, heal the initial agent, while agent $2$ guards it, effectively protecting the initial agent from the attack.

Fig. \ref{fig:followup}(c) illustrates the follow-up agents selected using our proposed Follow-up Agent Group Selection method. The selected group includes agents $2$ and $10$, which are responsible for healing and guarding the initial agent, and agents $6$ and $9$, which are hitting enemies targeting the initial agent. These agents are subjected to the follow-up attack, preventing them from performing their protective actions. As a result, the initial agent is left vulnerable, succumbs to enemy attacks, and is ultimately eliminated.

\newpage

\begin{figure}[h]
    \centering
    \begin{subfigure}{0.19\columnwidth}
        \centering
        \includegraphics[width=\linewidth]{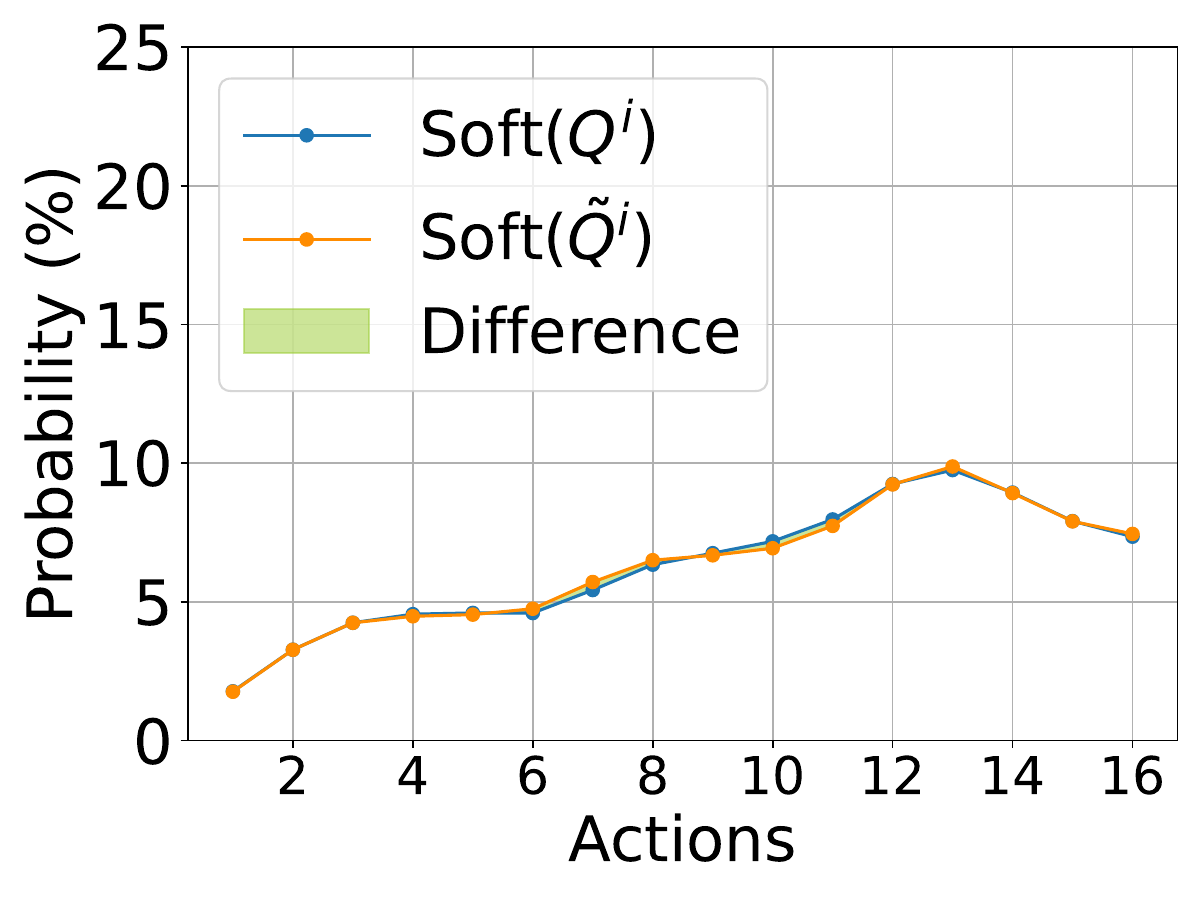}
        \caption{Agent=$1$}  
    \end{subfigure}
    \begin{subfigure}{0.19\columnwidth}
        \centering
        \includegraphics[width=\linewidth]{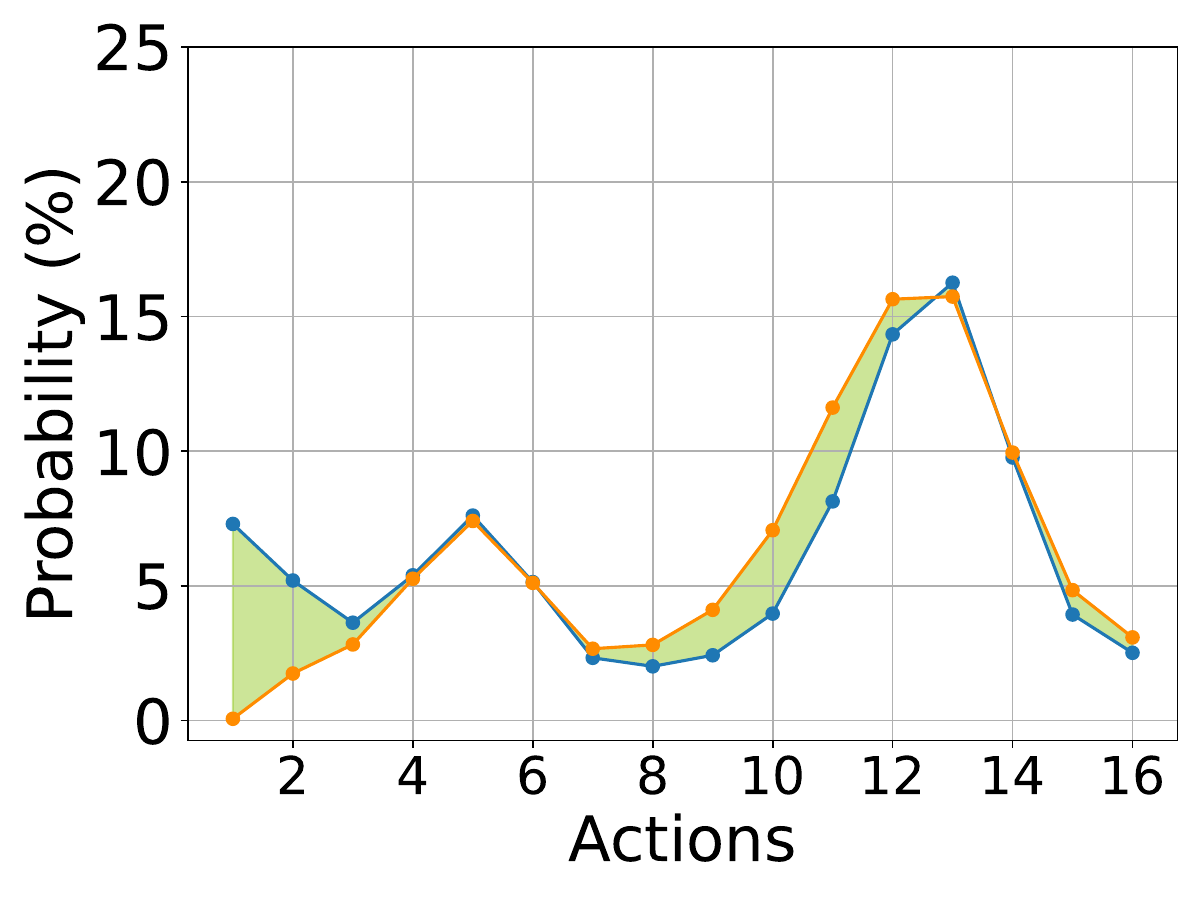}
        \caption{Agent=$2$ (selected)}  
    \end{subfigure}
    \begin{subfigure}{0.19\columnwidth}
        \centering
        \includegraphics[width=\linewidth]{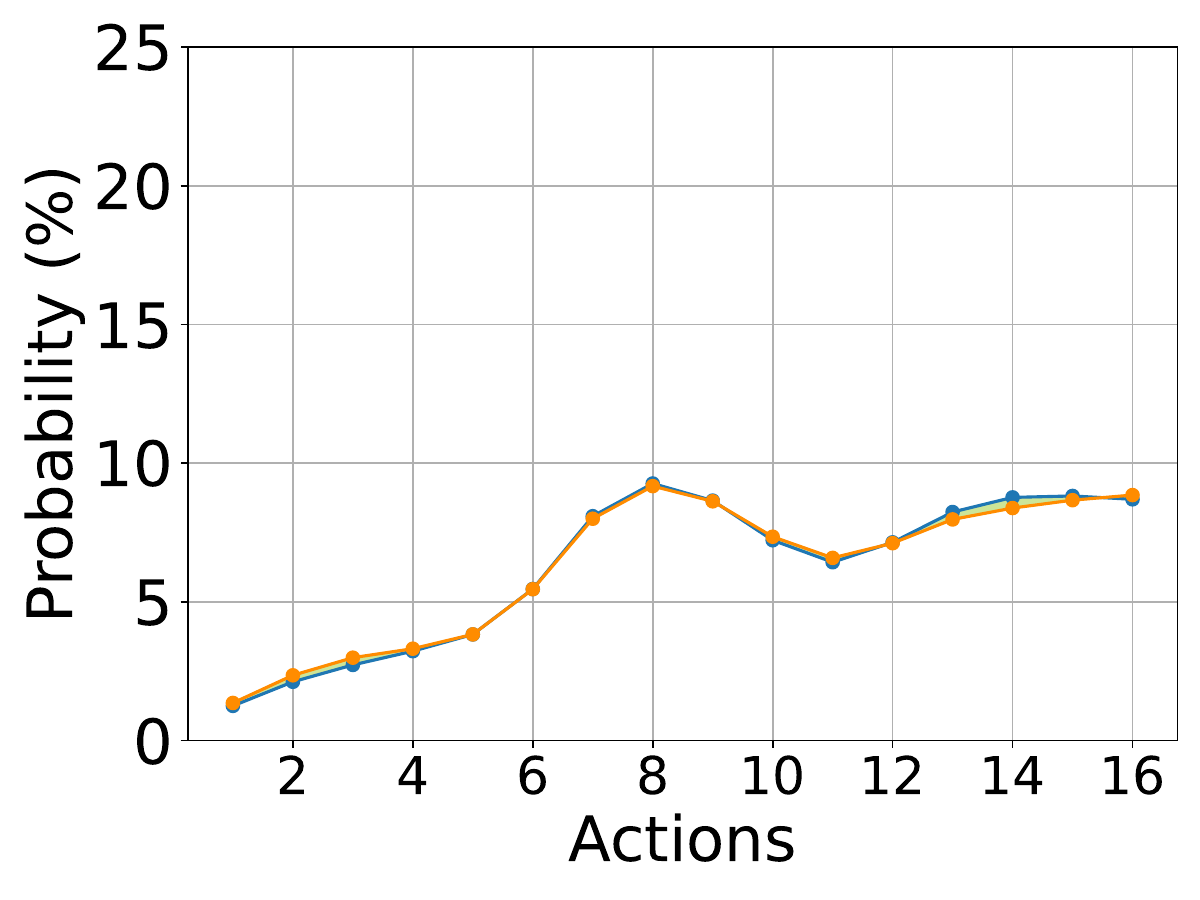}
        \caption{Agent=$3$}  
    \end{subfigure}
    \begin{subfigure}{0.19\columnwidth}
        \centering
        \includegraphics[width=\linewidth]{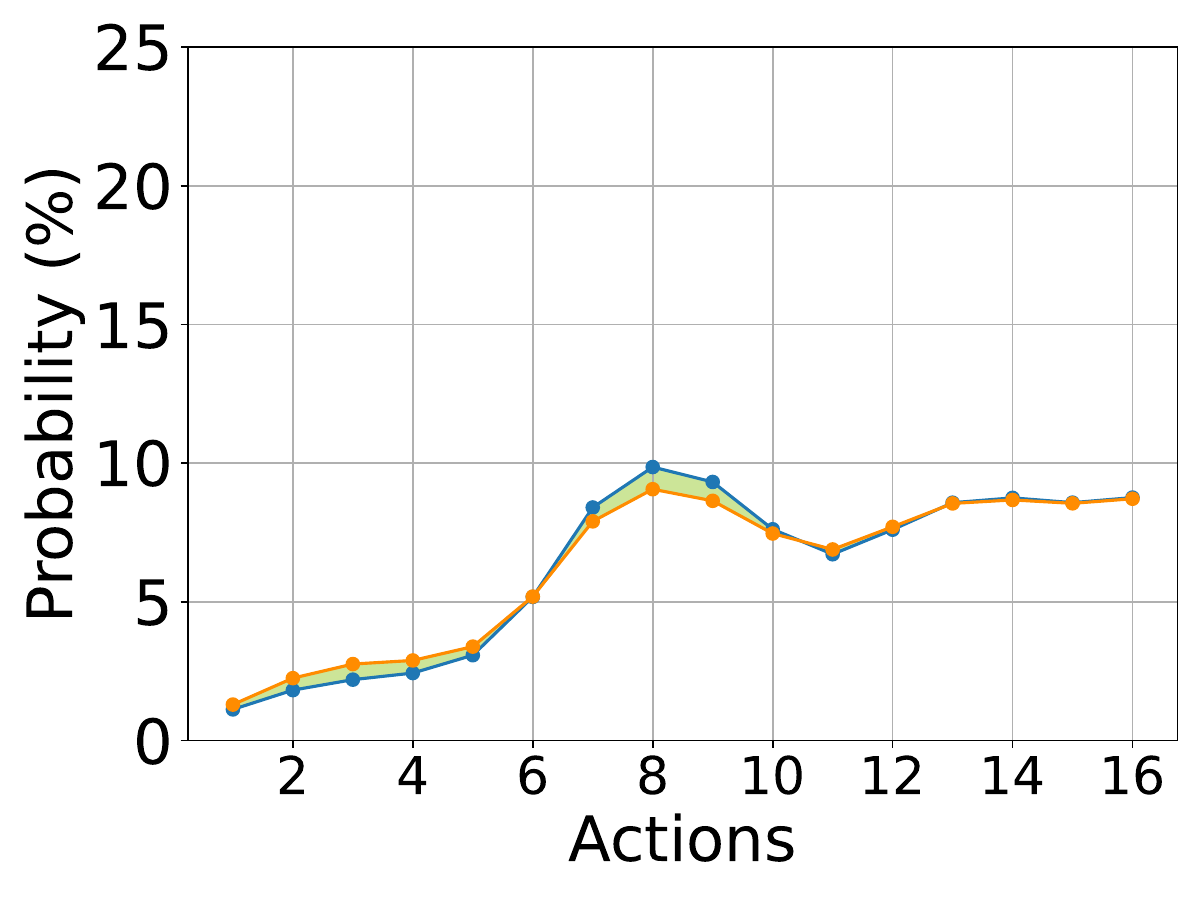}
        \caption{Agent=$4$}  
    \end{subfigure}
    \begin{subfigure}{0.19\columnwidth}
        \centering
        \includegraphics[width=\linewidth]{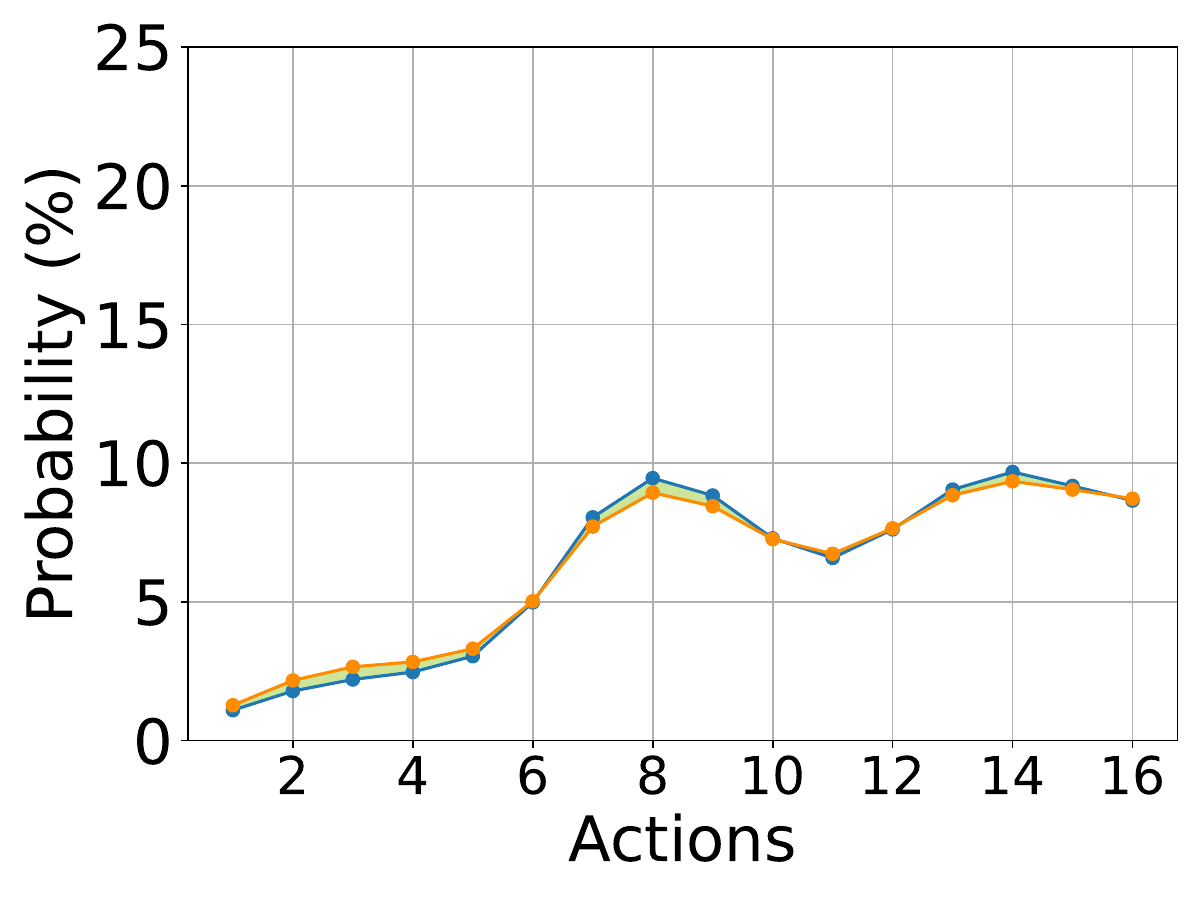}
        \caption{Agent=$5$}  
    \end{subfigure}
    \begin{subfigure}{0.19\columnwidth}
        \centering
        \includegraphics[width=\linewidth]{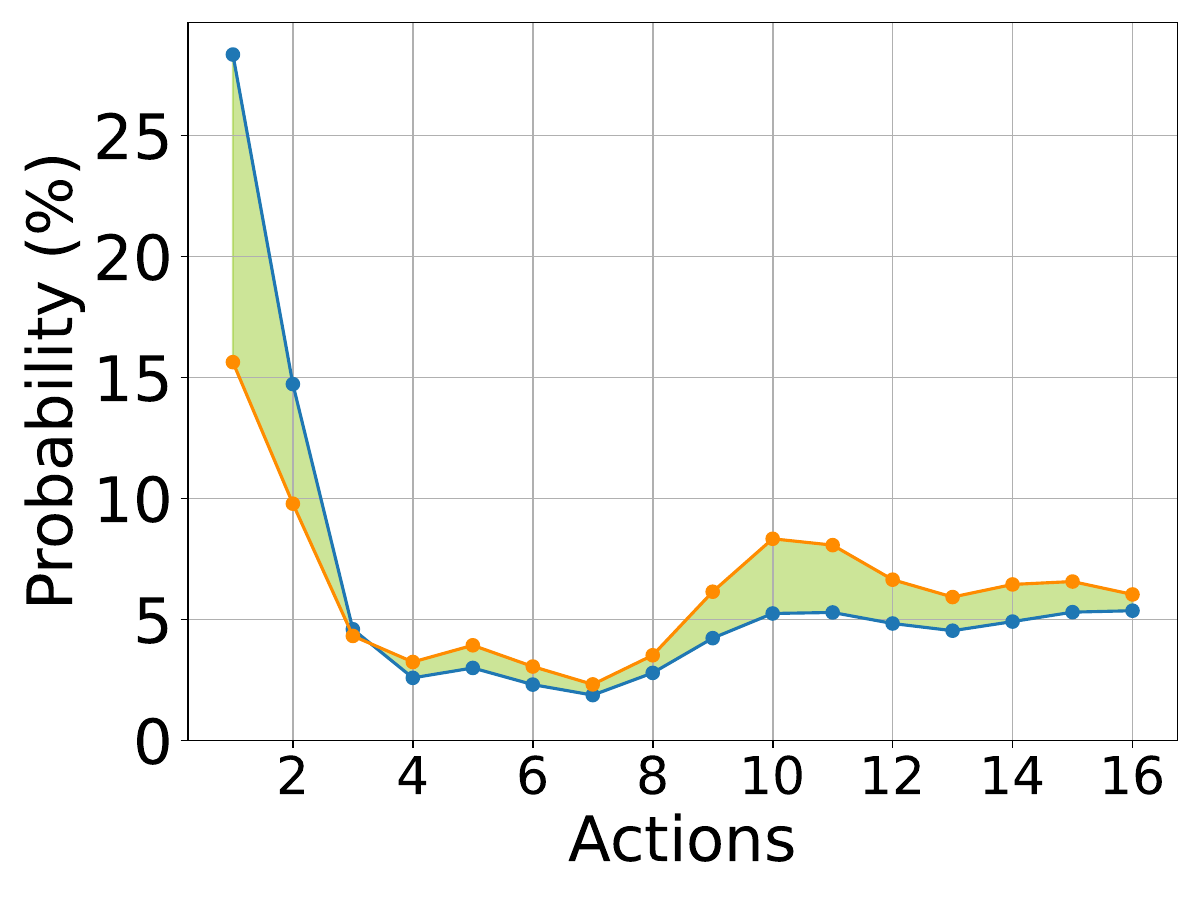}
        \caption{Agent=$6$ (selected)}  
    \end{subfigure}
    \begin{subfigure}{0.19\columnwidth}
        \centering
        \includegraphics[width=\linewidth]{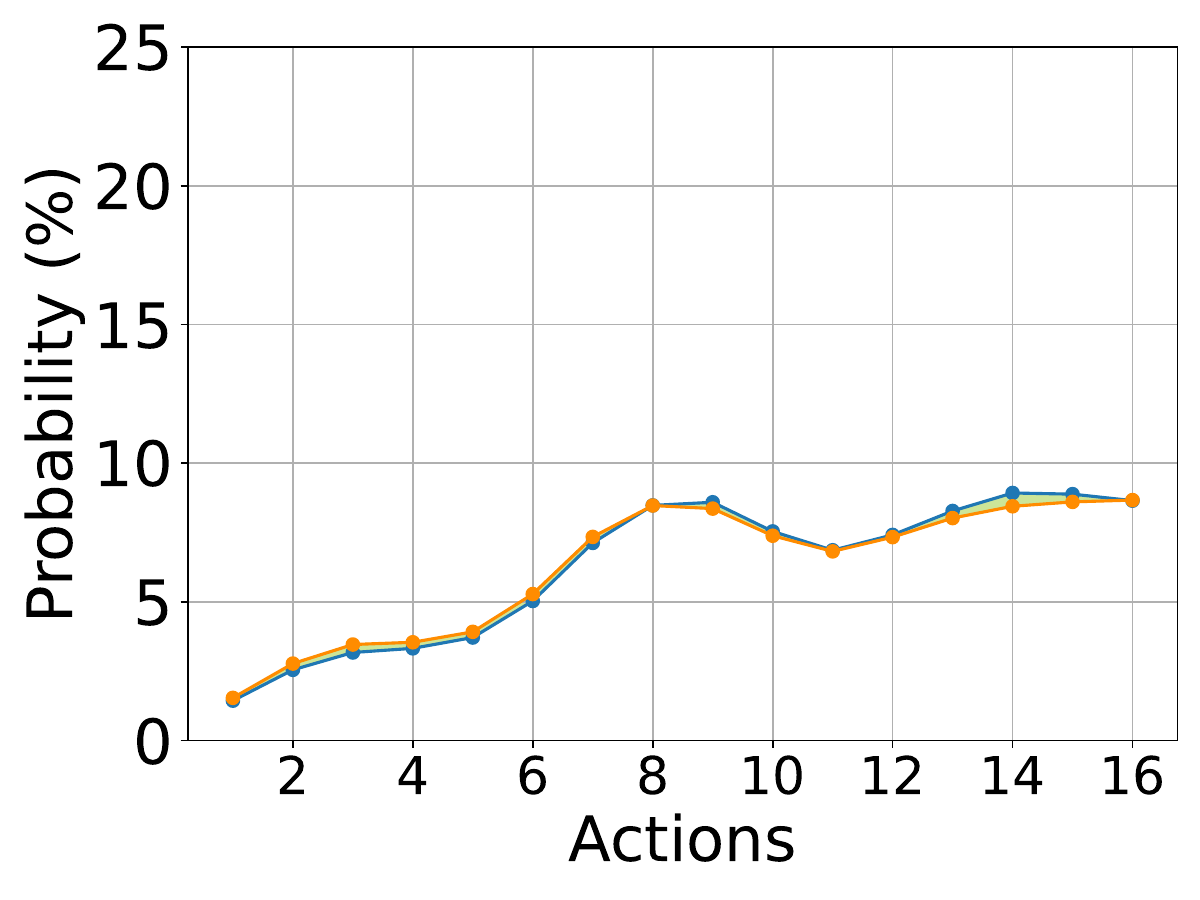}
        \caption{Agent=$7$}  
    \end{subfigure}
    \begin{subfigure}{0.19\columnwidth}
        \centering
        \includegraphics[width=\linewidth]{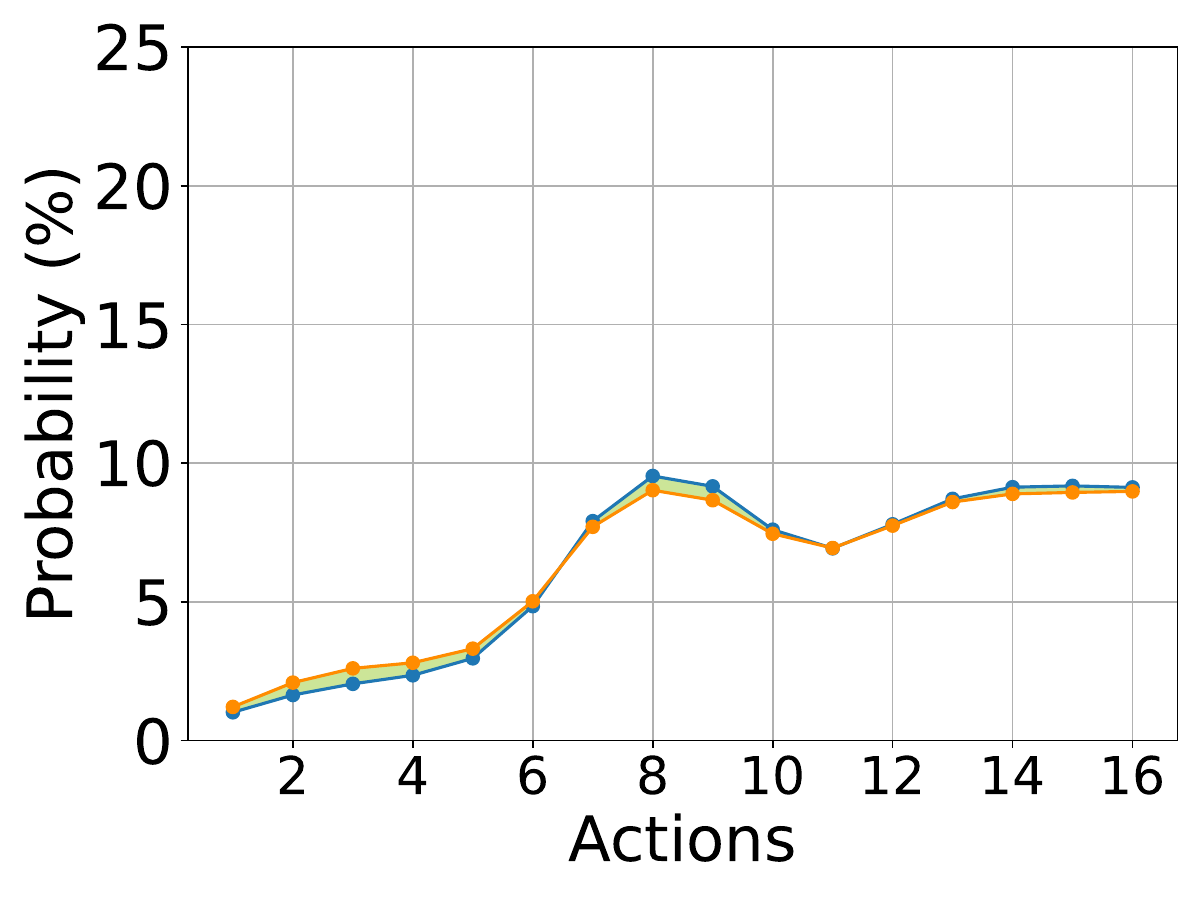}
        \caption{Agent=$8$}  
    \end{subfigure}
    \begin{subfigure}{0.19\columnwidth}
        \centering
        \includegraphics[width=\linewidth]{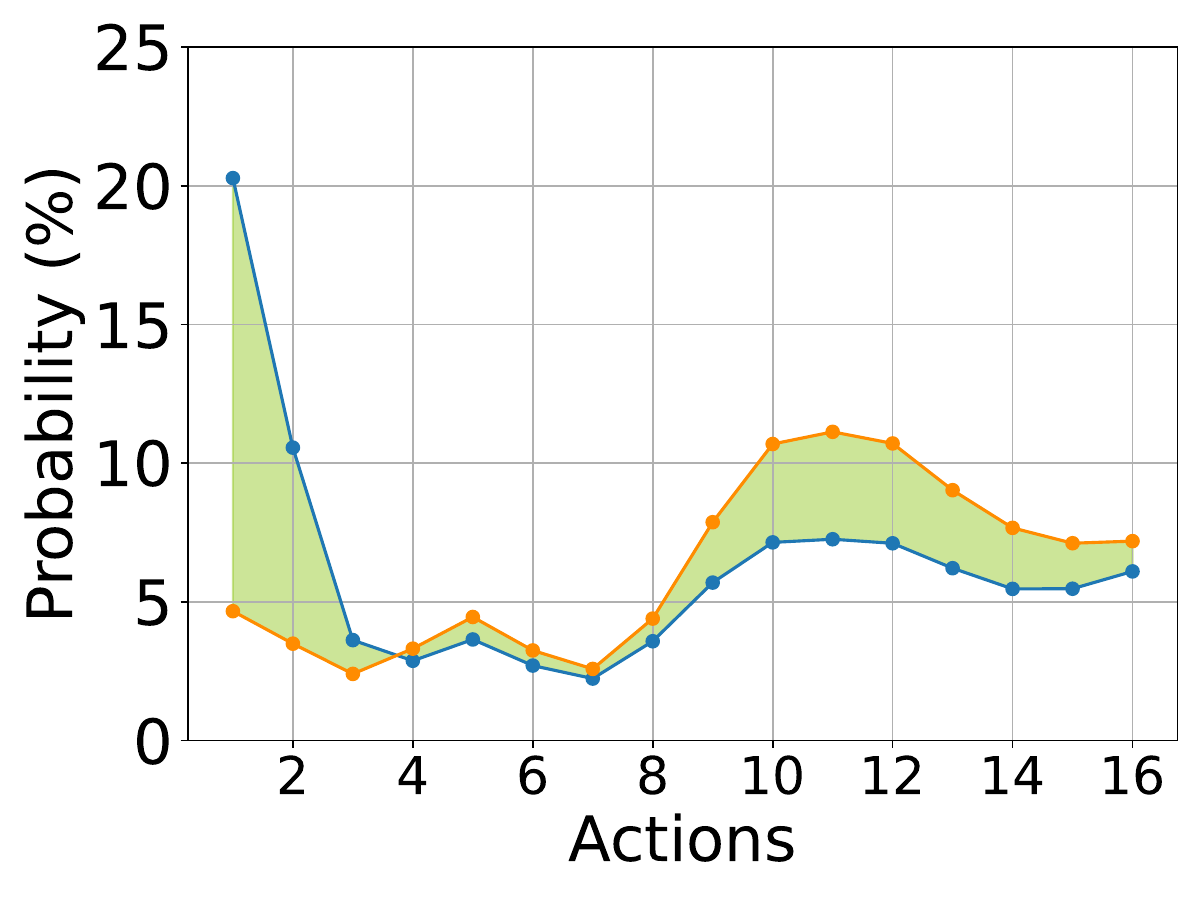}
        \caption{Agent=$9$ (selected)}  
    \end{subfigure}
    \begin{subfigure}{0.19\columnwidth}
        \centering
        \includegraphics[width=\linewidth]{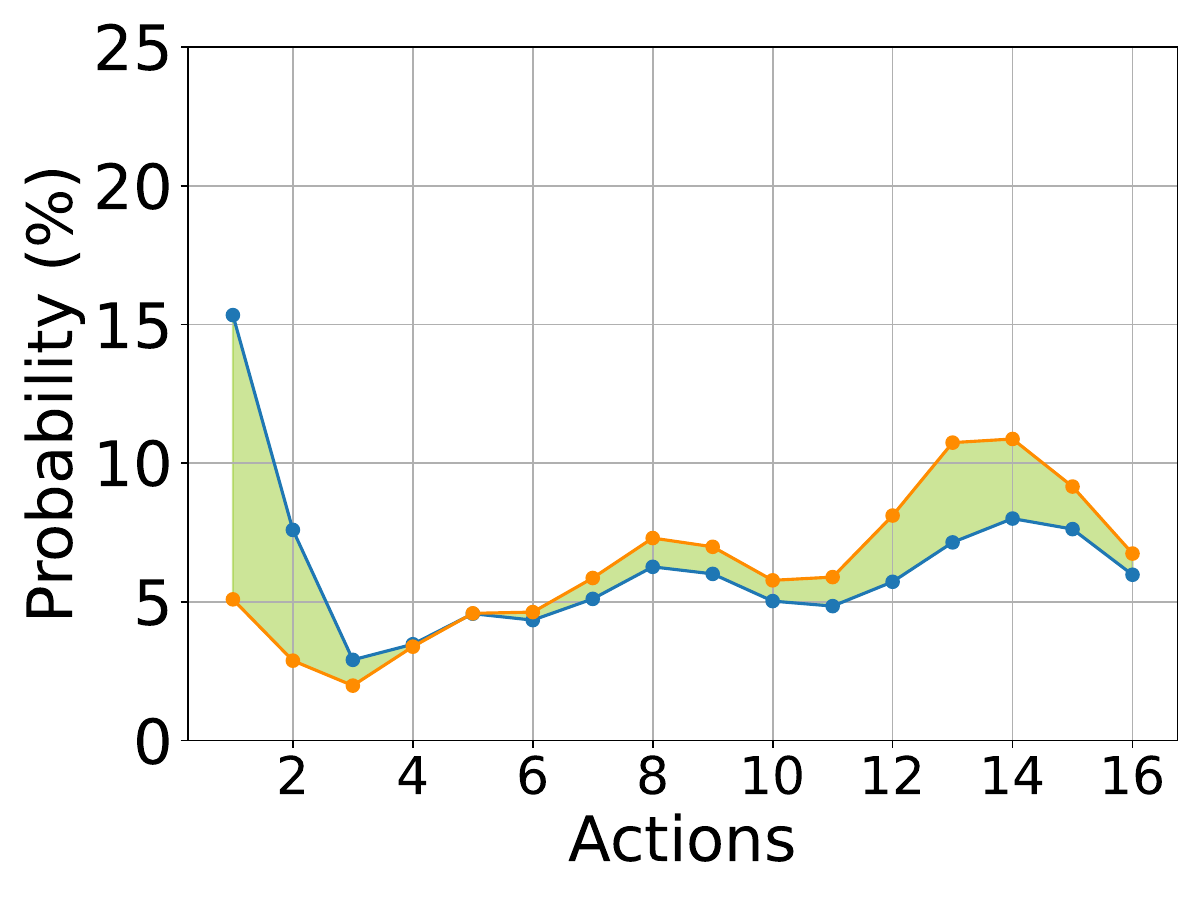}
        \caption{Agent=$10$ (selected)}  
    \end{subfigure}
    \caption{$Soft(Q^i)$ and $Soft(\tilde{Q}^i)$ for each agent, along with the difference between the two distributions.}
    \label{fig:difference}
\end{figure}

\begin{figure}[h]
    \centering
    \includegraphics[width=0.6\textwidth]{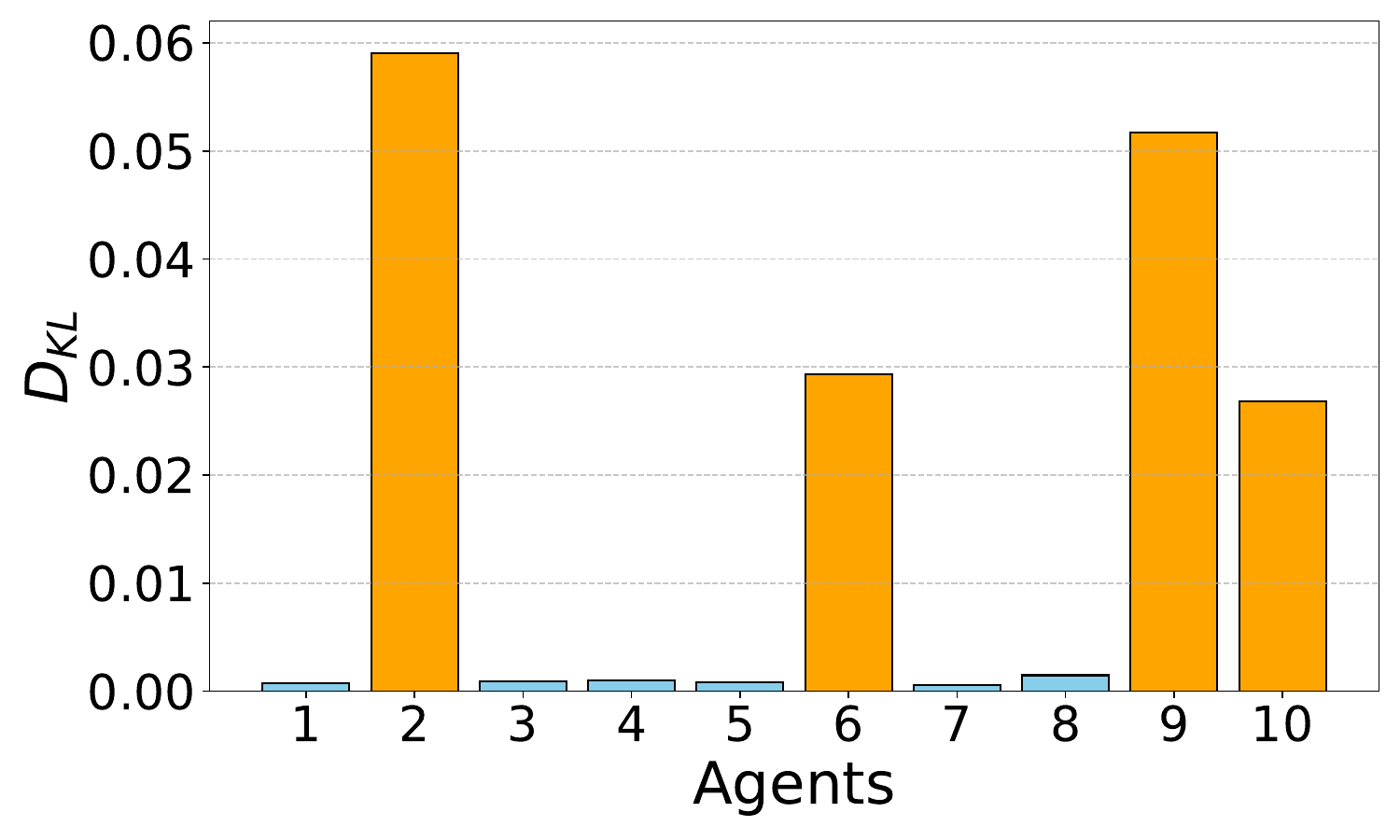} 
    \caption{KL divergence values for each agent, representing the difference between $Soft(Q^i)$ and $Soft(\tilde{Q}^i)$ distributions.}
    \label{fig:kl}
\end{figure}

Fig. \ref{fig:difference} illustrates the $Soft(Q^i)$ distribution and the updated $Soft(\tilde{Q}^i)$ distribution based on Equation \ref{eq:1}, highlighting the differences between the two distributions. It is evident that agents $2, 6, 9, 10$ exhibit the largest differences in their distributions. This suggests that, following the initial attack, these agents show noticeable policy changes to adapt and defend against it. Additionally, Fig. \ref{fig:kl} presents the KL divergence values between these two distributions, further confirming that agents $2, 6, 9, 10$ have the highest KL divergence values.

Consequently, based on Equation \ref{eq:2}, agents $2, 6, 9, 10$ are selected as the follow-up agent group. This aligns with the visualization results shown in Fig. \ref{fig:followup}, demonstrating consistency in the SMAC environment.

\newpage



\end{document}